\documentclass{article}

\usepackage[final, nonatbib]{neurips_2020}

\usepackage[utf8]{inputenc} 
\usepackage[T1]{fontenc}    
\usepackage{hyperref}       
\usepackage{url}            
\usepackage{booktabs}       
\usepackage{amsfonts}       
\usepackage{nicefrac}       
\usepackage{microtype}      
\usepackage{graphicx}
\usepackage{amsmath}
\usepackage{amsthm}
\usepackage{amssymb}
\usepackage{subfigure}
\usepackage{wrapfig}
\usepackage{algorithm}
\usepackage{algorithmic}
\usepackage{mathtools}

\usepackage{color}

\theoremstyle{plain}
\newtheorem{definition}{Definition}
\newtheorem{theorem}{Theorem}

\title{Generating Adjacency-Constrained Subgoals in Hierarchical Reinforcement Learning}

\author{%
  Tianren Zhang\thanks{Equal contribution.}\hspace{0.45em}$^{\hspace{-0.05em},1}$, Shangqi Guo$^{*,1}$, Tian Tan$^2$, Xiaolin Hu\thanks{Corresponding authors: Xiaolin Hu and Feng Chen.}$\hspace{0.35em}^{,3,4,5}$, Feng Chen$^{\dagger,1,6,7}$ \\
  $^1$ Department of Automation, Tsinghua University \\
  $^2$ Department of Civil and Environmental Engineering, Stanford University \\
  $^3$ Department of Computer Science and Technology, Tsinghua University \\
  $^4$ Beijing National Research Center for Information Science and Technology \\
  $^5$ State Key Laboratory of Intelligent Technology and Systems \\
  $^6$ Beijing Innovation Center for Future Chip \\
  $^7$ LSBDPA Beijing Key Laboratory \\
  \texttt{\{zhang-tr19,gsq15\}@mails.tsinghua.edu.cn; tiantan@stanford.edu;} \\
  \texttt{\{xlhu,chenfeng\}@mail.tsinghua.edu.cn}
}

\begin{document}

\maketitle

\addtocounter{footnote}{-2}

\begin{abstract}
Goal-conditioned hierarchical reinforcement learning (HRL) is a promising approach for scaling up reinforcement learning (RL) techniques. However, it often suffers from training inefficiency as the action space of the high-level, i.e., the goal space, is often large. Searching in a large goal space poses difficulties for both high-level subgoal generation and low-level policy learning. In this paper, we show that this problem can be effectively alleviated by restricting the high-level action space from the whole goal space to a $k$-step adjacent region of the current state using an adjacency constraint. We theoretically prove that the proposed adjacency constraint preserves the optimal hierarchical policy in deterministic MDPs, and show that this constraint can be practically implemented by training an adjacency network that can discriminate between adjacent and non-adjacent subgoals. Experimental results on discrete and continuous control tasks show that incorporating the adjacency constraint improves the performance of state-of-the-art HRL approaches in both deterministic and stochastic environments.\footnote{Code is available at~\url{https://github.com/trzhang0116/HRAC}.}
\end{abstract}

\section{Introduction}
\label{sec:intro}
Hierarchical reinforcement learning (HRL) has shown great potentials in scaling up reinforcement learning (RL) methods to tackle large, temporally extended problems with long-term credit assignment and sparse rewards~\cite{sutton_between_1999,precup_temporal_2000,barto_recent_2003}. As one of the prevailing HRL paradigms, goal-conditioned HRL framework~\cite{dayan_feudal_1993,schmidhuber_planning_1993,kulkarni_hierarchical_2016,vezhnevets_feudal_2017,nachum_data-efficient_2018,levy_learning_2019}, which comprises a high-level policy that breaks the original task into a series of subgoals and a low-level policy that aims to reach those subgoals, has recently achieved significant success.
However, the effectiveness of goal-conditioned HRL relies on the acquisition of effective and semantically meaningful subgoals, which still remains a key challenge.

As the subgoals can be interpreted as high-level actions, it is feasible to directly train the high-level policy to generate subgoals using external rewards as supervision, which has been widely adopted in previous research~\cite{nachum_data-efficient_2018,nachum_near-optimal_2019,levy_learning_2019,kulkarni_hierarchical_2016,vezhnevets_feudal_2017}. Although these methods require little task-specific design, they often suffer from training inefficiency. This is because the action space of the high-level, i.e., the goal space, is often as large as the state space. The high-level exploration in such a large action space results in inefficient learning. As a consequence, the low-level training also suffers as the agent tries to reach every possible subgoal produced by the high-level policy.    


One effective way for handling large action spaces is action space reduction or action elimination. However, it is difficult to perform action space reduction in general scenarios without additional information, since a restricted action set may not be expressive enough to form the optimal policy. There has been limited literature~\cite{zahavy_learn_nodate,van_de_wiele_q-learning_2020,khetarpal_what_2020} studying action space reduction in RL, and to our knowledge, there is no prior work studying action space reduction in HRL, since the information loss in the goal space can lead to severe performance degradation~\cite{nachum_near-optimal_2019}.

\begin{figure}[t]
\centering
\begin{minipage}[t]{0.38\textwidth}
\centering
\includegraphics[width=0.9\linewidth]{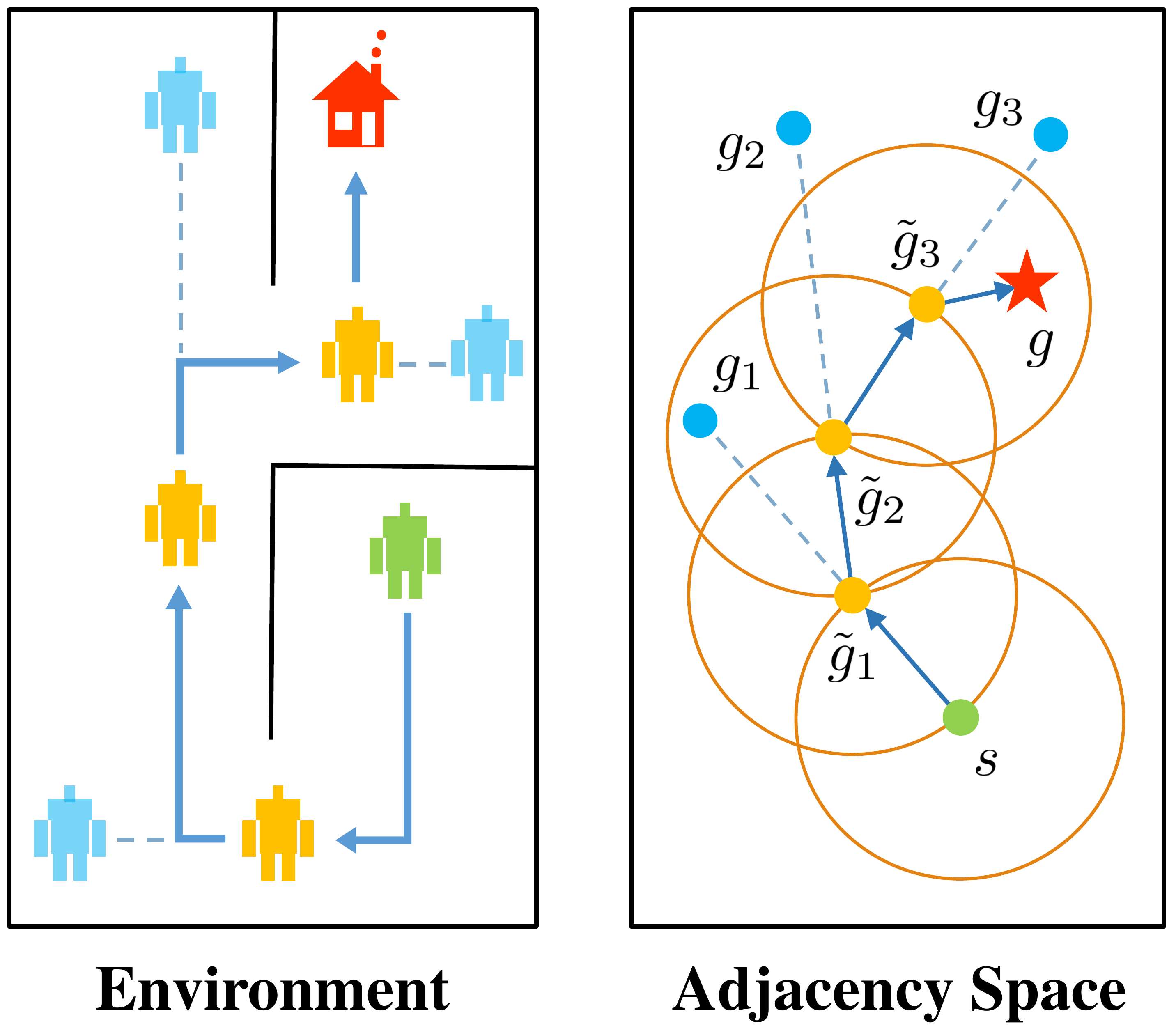}
\caption{High-level illustration of our method: distant subgoals $g_1,\,g_2,\,g_3$ (blue) can be surrogated by closer subgoals $\tilde{g}_1,\,\tilde{g}_2,\,\tilde{g}_3$ (yellow) that fall into the $k$-step adjacent regions.
}
\label{fig:motivation}
\end{minipage}
\hspace{0.4em}
\begin{minipage}[t]{0.22\textwidth}
\centering
\includegraphics[width=0.9\linewidth]{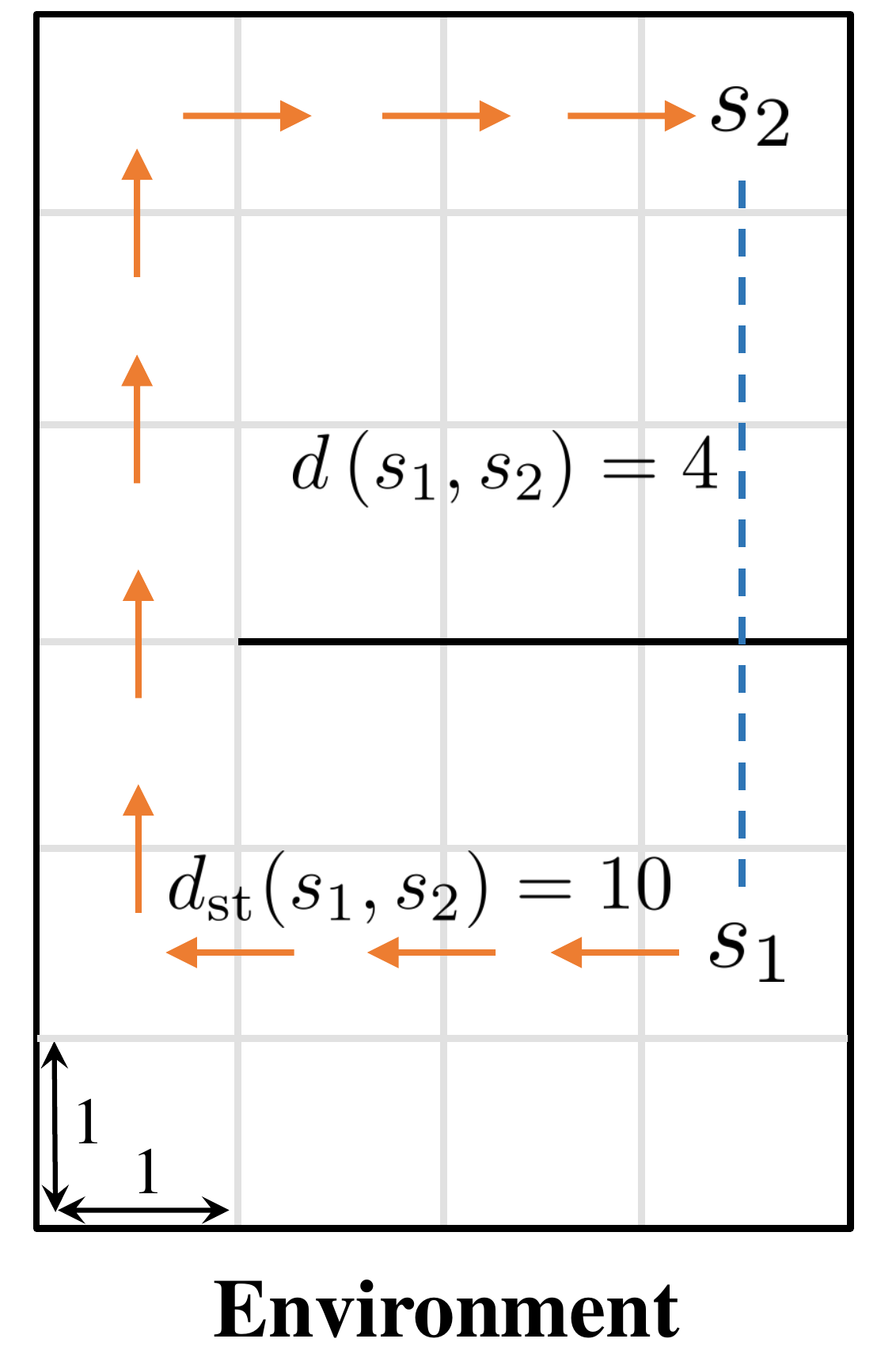}
\caption{Comparison between shortest transition distance $d_{\mathrm{st}}$ and Euclidean distance $d$ in a toy environment.}
\label{fig:dist}
\end{minipage}
\hspace{0.4em}
\begin{minipage}[t]{0.33\textwidth}
\centering
\includegraphics[width=0.95\linewidth]{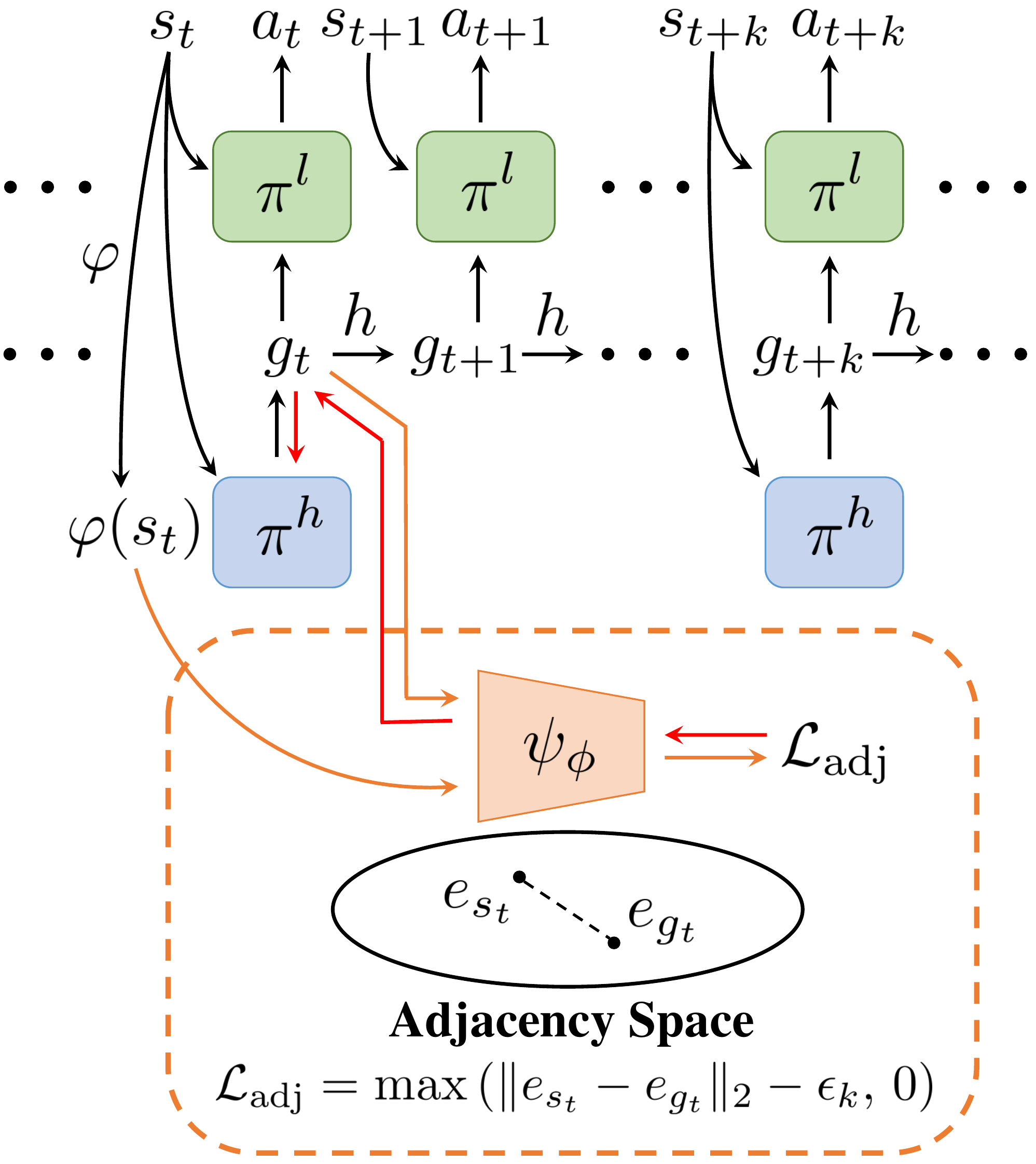}
\caption{The goal-conditioned HRL framework and the $k$-step adjacency constraint implemented by the adjacency network $\psi_\phi$ (dashed orange box).}
\label{fig:method}
\end{minipage}
\end{figure}

In this paper, we present an optimality-preserving high-level action space reduction method for goal-conditioned HRL. Concretely, we show that the high-level action space can be restricted from the whole goal space to a $k$-step adjacent region centered at the current state. Our main intuition is depicted in Figure~\ref{fig:motivation}: distant subgoals can be substituted by closer subgoals, as long as they drive the low-level to move towards the same ``direction''. Therefore, given the current state $s$ and the subgoal generation frequency $k$,
the high-level only needs to explore in a subset of subgoals covering states that the low-level can possibly reach within $k$ steps.
By reducing the action space of the high-level, the learning efficiency of both the high-level and the low-level can be improved: for the high-level, a considerably smaller action space relieves the burden of exploration and value function approximation; for the low-level, adjacent subgoals provide a stronger learning signal as the agent can be intrinsically rewarded with a higher frequency for reaching these subgoals. Formally, we introduce a $k$-step adjacency constraint
for high-level action space reduction,
and theoretically prove that the proposed constraint preserves the optimal hierarchical policy in deterministic MDPs. Also, to practically implement the constraint,
we propose to train an adjacency network so that the $k$-step adjacency between all states and subgoals can be succinctly derived.

We benchmark our method on various tasks, including discrete control and planning tasks on grid worlds and challenging continuous control tasks based on the MuJoCo simulator~\cite{todorov_mujoco_2012}, which have been widely used in HRL literature~\cite{nachum_data-efficient_2018,levy_learning_2019,nachum_near-optimal_2019,florensa_stochastic_2017}. Experimental results exhibit the superiority of our method on both sample efficiency and asymptotic performance compared with the state-of-the-art HRL approach HIRO~\cite{nachum_data-efficient_2018}, demonstrating the effectiveness of the proposed adjacency constraint.


\section{Preliminaries}

We consider a finite-horizon, goal-conditioned Markov Decision Process (MDP) defined as a tuple $\langle\mathcal{S},\mathcal{G},\mathcal{A},\mathcal{P},\mathcal{R},\gamma\rangle$, where $\mathcal{S}$ is a state set, $\mathcal{G}$ is a goal set, $\mathcal{A}$ is an action set, $\mathcal{P}:\mathcal{S}\times\mathcal{A}\times\mathcal{S}\rightarrow \mathbb{R}$ is a state transition function, $\mathcal{R}:\mathcal{S}\times\mathcal{A}\rightarrow\mathbb{R}$ is a reward function, and $\gamma \in [0,1)$ is a discount factor.
Following prior work~\cite{kulkarni_hierarchical_2016,vezhnevets_feudal_2017,nachum_data-efficient_2018}, we consider a framework comprising two hierarchies: a high-level controller with policy $\pi^h_{\theta_h}(g|s)$ and a low-level controller with policy $\pi^l_{\theta_l}(a|s,g)$ parameterized by two function approximators, e.g., neural networks with parameters $\theta_h$ and $\theta_l$ respectively, as shown in Figure~\ref{fig:method}.
The high-level controller aims to maximize the external reward and generates a high-level action, i.e., a subgoal $g_t\sim\pi^h_{\theta_h}(g|s_t)\in\mathcal{G}$ every $k$ time steps when $t\equiv 0\,(\mathrm{mod}\ k)$, where $k>1$ is a pre-determined hyper-parameter. It modulates the behavior of the low-level policy by intrinsically rewarding the low-level for reaching these subgoals.
The low-level aims to maximize the intrinsic reward provided by the high-level, and performs a primary action $a_t\sim\pi^l_{\theta_l}(a|s_t,g_t)\in\mathcal{A}$ at every time step.
Following prior methods~\cite{nachum_data-efficient_2018,andrychowicz_hindsight_2017}, we consider a goal space $\mathcal{G}$ which is a sub-space of $\mathcal{S}$ with a known mapping function $\varphi:\mathcal{S}\rightarrow\mathcal{G}$.
When $t\not\equiv 0\,(\mathrm{mod}\ k)$, a pre-defined goal transition process $g_t = h(g_{t-1},s_{t-1},s_t)$ is utilized.
We adopt directional subgoals that represent the differences between desired states and current states~\cite{vezhnevets_feudal_2017,nachum_data-efficient_2018}, where the goal transition function is set to $h(g_{t-1},s_{t-1},s_t) = g_{t-1} + s_{t-1} - s_t$. The reward function of the high-level policy is defined as:
\begin{equation}
r_{kt}^h = \sum_{i=kt}^{kt+k-1}\mathcal{R}(s_i,a_i),\quad t = 0,\,1,\,2,\,\cdots,
\label{equ:r_high}
\end{equation}
which is the accumulation of the external reward in the time interval $[kt,kt+k-1]$. 

While the high-level controller is motivated by the environmental reward, the low-level controller has no direct access to this external reward. Instead, the low-level is supervised by the intrinsic reward that describes subgoal-reaching performance, defined as $r_t^l = -D\left(g_t, \varphi(s_{t+1})\right)$, where $D$ is a binary or continuous distance function. In practice, we employ Euclidean distance as $D$.

The goal-conditioned HRL framework above enables us to train high-level and low-level policies concurrently in an end-to-end fashion. However, it often suffers from training inefficiency due to the unconstrained subgoal generation process, as we have mentioned in Section~\ref{sec:intro}. In the following section, we will introduce the $k$-step adjacency constraint to mitigate this issue.

\section{Theoretical Analysis}
\label{sec:theory}

In this section, we provide our theoretical results and show that the optimality can be preserved when learning a high-level policy with $k$-step adjacency constraint. We begin by introducing a distance measure that can decide whether a state is ``close'' to another state. In this regard, common distance functions such as the Euclidean distance are not suitable, as they often cannot reveal the real structure of the MDP.        
Therefore, we introduce \emph{shortest transition distance},
which equals to the minimum number of steps required to reach a target state from a start state, as shown in Figure~\ref{fig:dist}. In stochastic MDPs, the number of steps required is not a fixed number, but a distribution conditioned on a specific policy. In this case, we resort to the notion of \emph{first hit time} from stochastic processes, and 
define the shortest transition distance by minimizing the expected first hit time over all possible policies. 

\begin{definition}
Let $s_1,\,s_2\in\mathcal{S}$. Then, the shortest transition distance from $s_1$ to $s_2$ is defined as:
\begin{equation}
d_{\mathrm{st}}(s_1,s_2) \vcentcolon= \min_{\pi\in\Pi}\mathbb{E}[\mathcal{T}_{s_1s_2} |\pi] = \min_{\pi\in\Pi} \sum_{t=0}^\infty t P(\mathcal{T}_{s_1s_2}=t|\pi),
\label{equ:dist}
\end{equation}
where $\Pi$ is the complete policy set and $\mathcal{T}_{s_1s_2}$ denotes the first hit time from $s_1$ to $s_2$.  
\label{def:distance}
\end{definition}

The shortest transition distance is determined by a policy that connects states $s_1$ and $s_2$ in the most efficient way, which has also been studied by several prior work~\cite{florensa_self-supervised_2019,eysenbach_search_2019}. This policy is optimal in the sense that it requires the minimum number of steps to reach state $s_2$ from state $s_1$.
Compared with the dynamical distance~\cite{hartikainen_dynamical_2020}, our definition here does not rely on a specific non-optimal policy. Also, we do not assume that the environment is reversible, i.e., $d_{\mathrm{st}}(s_1, s_2) = d_{\mathrm{st}}(s_2, s_1)$ does not hold for all pairs of states. Therefore, the shortest transition distance is a quasi-metric as it does not satisfy the symmetry condition. However, this limitation does not affect the following analysis as we only need to consider the transition from the start state to the goal state without the reversed transition. 

Given the definition of the shortest transition distance, we now formulate the property of an optimal (deterministic) goal-conditioned policy $\pi^*:\mathcal{S}\times\mathcal{G}\rightarrow\mathcal{A}$~\cite{schaul_universal_2015}. We have:


\begin{equation}
\pi^*(s,g) \in \underset{a\in\mathcal{A}}{\arg\min} \sum_{s'\in\mathcal{S}}P(s'|s,a)\,d_{\mathrm{st}}\left(s',\varphi^{-1}(g)\right),\,\forall s \in \mathcal{S},\, g\in \mathcal{G},
\label{equ:goal_conditioned}
\end{equation}

where $\varphi^{-1}: \mathcal{G}\rightarrow\mathcal{S}$ is the known inverse mapping of $\varphi$. We then consider the goal-conditioned HRL framework with high-level action frequency $k$. Different from a flat goal-conditioned policy, in this setting the low-level policy is required to reach the subgoals with $k$ limited steps. As a result, only a subset of the original states can be reliably reached even with an optimal goal-conditioned policy. We introduce the notion of \emph{$k$-step adjacent region} to describe the set of subgoals mapped from this reachable subset of states. 

\begin{definition}
Let $s\in\mathcal{S}$. Then, the $k$-step adjacent region of $s$ is defined as:
\begin{equation}
\mathcal{G}_A(s,k) \vcentcolon= \{ g\in \mathcal{G} \,|\, d_{\mathrm{st}}\left(s, \varphi^{-1}(g)\right)\le k \}.
\label{equ:adjacent_region}
\end{equation}
\end{definition}

Harnessing the property of $\pi^*$, we can show that in deterministic MDPs, given an optimal low-level policy $\pi^{l*}=\pi^*$, subgoals that fall in the $k$-step adjacent region of the current state can represent all optimal subgoals in the whole goal space in terms of the induced $k$-step low-level action sequence. We summarize this finding in the following theorem.

\begin{theorem}
Let $s\in\mathcal{S},\,g\in\mathcal{G}$ and let $\pi^*$ be an optimal goal-conditioned policy. Under the assumptions that the MDP is deterministic and that the MDP states are strongly connected, for all $k\in\mathbb{N}_+$ satisfying $k \le d_{\mathrm{st}}(s,\varphi^{-1}(g))$, there exists a surrogate goal $\tilde{g}$ such that:
\label{theo:low}
\begin{equation}
\begin{aligned}
&\tilde{g} \in \mathcal{G}_A(s,k), \\
&\pi^*(s_i, \tilde{g}) = \pi^*(s_i, g),\, \forall s_i\in \tau\, (i\ne k), 
\end{aligned}
\end{equation}
where $\tau \vcentcolon= (s_0,s_1,\cdots,s_k)$ is the $k$-step state trajectory starting from state $s_0 = s$ under $\pi^*$ and $g$. 
\end{theorem}

Theorem~\ref{theo:low} suggests that the $k$-step low-level action sequence generated by an optimal low-level policy conditioned on a distant subgoal can be induced using a subgoal that is closer. Naturally, we can generalize this result to a two-level goal-conditioned HRL framework, where the low-level is actuated not by a single subgoal, but by a subgoal sequence produced by the high-level policy.

\begin{theorem}
Given the high-level action frequency $k$ and the high-level planning horizon $T$, for $s\in\mathcal{S}$, let $\rho^* = (g_0,g_k,\cdots,g_{(T-1)k})$ be the high-level subgoal trajectory starting from state $s_0=s$ under an optimal high-level policy $\pi^{h*}$. Also, let $\tau^* = (s_0,s_k,s_{2k},\cdots,s_{Tk})$ be the high-level state trajectory under $\rho^*$ and an optimal low-level policy $\pi^{l*}$. Then, there exists a surrogate subgoal trajectory $\tilde{\rho}^* = (\tilde{g}_0,\tilde{g}_k,\cdots,\tilde{g}_{(T-1)k})$ such that:
\begin{equation}
\begin{aligned}
&\tilde{g}_{kt} \in \mathcal{G}_A({s_{kt},k}), \\
&Q^*(s_{kt},\tilde{g}_{kt}) = Q^*(s_{kt},g_{kt}),\quad t=0,\,1,\,\cdots,\,T-1,
\end{aligned}
\end{equation}
where $Q^*$ is the optimal high-level $Q$-function under policy $\pi^{h*}$. 
\label{theo:high}
\end{theorem}

Theorem~\ref{theo:low} and~\ref{theo:high} show that we can constrain the high-level action space to state-wise $k$-step adjacent regions without the loss of optimality. We formulate the high-level objective incorporating this $k$-step adjacency constraint as:
\begin{equation}
  {\begin{aligned}
  \underset{\theta_h}{\mathrm{max}}\quad &\mathbb{E}_{\pi^h_{\theta_h}}\sum_{t=0}^{T-1}\gamma^{t} r_{kt}^h  \\
  \mathrm{subject\ to}\quad & d_{\mathrm{st}}\left(s_{kt}, \varphi^{-1}(g_{kt})\right) \le k,\quad t = 0,\,1,\,\cdots,\,T-1
  \end{aligned}
  },
\label{equ:formulation}
\end{equation}
where $r_{kt}^h$ is the high-level reward defined by Equation~\eqref{equ:r_high} and $g_{kt}\sim\pi^h_{\theta_h}(g|s_{kt})$.

In practice, Equation~\eqref{equ:formulation} is hard to optimize due to the strict constraint. Therefore, we employ relaxation methods and derive the following un-constrained optimizing objective:
\begin{equation}
  \underset{\theta_h}{\mathrm{max}}\quad \mathbb{E}_{\pi^h_{\theta_h}}\sum_{t=0}^{T-1}\Bigg[\gamma^{t} r_{kt}^h - \eta\cdot H\Big(d_{\mathrm{st}}\left(s_{kt},\varphi^{-1}(g_{kt})\right),k\Big)\Bigg] ,
  \label{equ:formulation_unconstrained}
\end{equation}
where $H(x,k) = \max(x/k-1,0)$ is a hinge loss function and $\eta$ is a balancing coefficient.

One limitation of our theoretical results is that the theorems are derived in the context of deterministic MDPs. However, these theorems are instructive for practical algorithm design in general cases, and the deterministic assumption has also been exploited by some prior works that investigate distance metrics in MDPs~\cite{hartikainen_dynamical_2020,castro_scalable_2019}. Also, we note that many real-world applications can be approximated as environments with deterministic dynamics where the stochasticity is mainly induced by noise. Hence, we may infer that the adjacency constraint could preserve a near-optimal policy when the magnitude of noise is small. Empirically, we show that our method is robust to certain types of stochasticity (see Section~\ref{sec:experiment} for details), and we leave rigorous theoretical analysis for future work.

\section{HRL with Adjacency Constraint}
\label{sec:method}

Although we have formulated the adjacency constraint in Section~\ref{sec:theory}, the exact calculation of the shortest transition distance $d_{\mathrm{st}}(s_1,s_2)$ between two arbitrary states $s_1,s_2\in\mathcal{S}$ remains complex and non-differentiable.
In this section, we introduce a simple method to collect and aggregate the adjacency information from the environment interactions. We then train an adjacency network using the aggregated adjacency information to approximate the shortest transition distance $d_{\mathrm{st}}(s_1,s_2)$ in a parameterized form, which enables a practical optimization of Equation~\eqref{equ:formulation_unconstrained}.

\subsection{Parameterized Approximation of Shortest Transition Distances}
\label{subsec:adjacency}

As shown in prior research~\cite{pong_temporal_2018,florensa_self-supervised_2019,eysenbach_search_2019,hartikainen_dynamical_2020}, accurately computing the shortest transition distance is not easy and often has the same complexity as learning an optimal low-level goal-conditioned policy. However, from the perspective of goal-conditioned HRL, we do not need a perfect shortest transition distance measure or a low-level policy that can reach any distant subgoals. Instead, only a discriminator of $k$-step adjacency is needed, and it is enough to learn a low-level policy that can reliably reach nearby subgoals (more accurately, subgoals that fall into the $k$-step adjacent region of the current state) rather than all potential subgoals in the goal space. 

Given the above, here we introduce a simple approach to determine whether a subgoal satisfies the $k$-step adjacency constraint. We first note that Equation~\eqref{equ:dist} can be approximated as follows:
\begin{equation}
d_{\mathrm{st}}(s_1,s_2) \approx \min_{\pi\in\{\pi_1,\pi_2,\cdots,\pi_n\}} \sum_{t=0}^\infty t P(\mathcal{T}_{s_1s_2}=t|\pi),
\label{equ:approximation}
\end{equation}
\begin{wrapfigure}[21]{r}{0pt}
\centering
\includegraphics[width=0.27\linewidth]{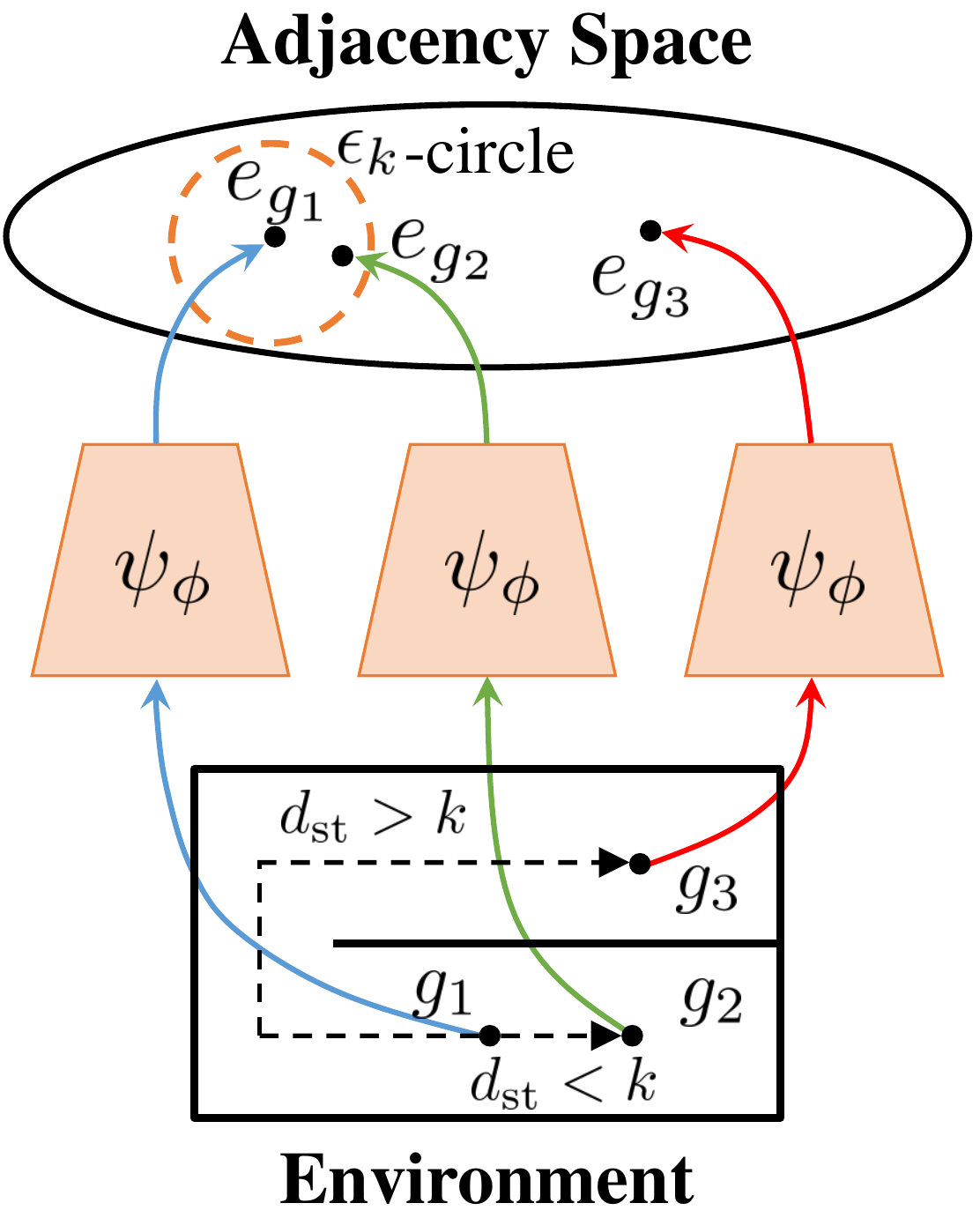}
\caption{The functionality of the adjacency network. The $k$-step adjacent region is mapped to an $\epsilon_k$-circle in the adjacency space, where $e_{g_i} = \psi_\theta(g_i),\,i=1,2,3$.}
\label{fig:distance_learning}
\end{wrapfigure}
where $\{\pi_1,\pi_2,\cdots,\pi_n\}$ is a finite policy set containing $n$ different deterministic policies. Obviously, if these policies are diverse enough, we can effectively approximate the shortest transition distance with a sufficiently large $n$. However,
training a set of diverse policies separately is costly,
and using one single policy to approximate the policy set ($n=1$)~\cite{savinov_semi-parametric_2018,savinov_episodic_2019} often leads to non-optimality. To handle this difficulty, we exploit the fact that the low-level policy itself changes over time during the training procedure.
We can thus build a policy set by sampling policies that emerge in different training stages.
To aggregate the adjacency information gathered by multiple policies, we propose to explicitly memorize the adjacency information by constructing a binary \emph{$k$-step adjacency matrix} of the explored states. The adjacency matrix has the same size as the number of explored states, and each element represents whether two states are $k$-step adjacent. In practice, we use the agent's trajectories, where the temporal distances between states can indicate their adjacency, to construct and update the adjacency matrix online. More details are in the supplementary material.
In practice, using an adjacency matrix is not enough as this procedure is non-differentiable and cannot generalize to newly-visited states.
To this end, we further distill the adjacency information stored in a constructed adjacency matrix into an adjacency network $\psi_\phi$ parameterized by $\phi$.
The adjacency network learns a mapping from the goal space to an adjacency space, where the Euclidean distance between the state and the goal is consistent with their shortest transition distance:
\begin{equation}
\tilde{d}_{\mathrm{st}}(s_1,s_2|\phi) \vcentcolon= \frac{k}{\epsilon_k}\lVert \psi_\phi(g_1) - \psi_\phi(g_2) \rVert_2\approx d_{\mathrm{st}}(s_1,s_2),
\label{equ:adjacency_net}
\end{equation}
where $g_1=\varphi(s_1),\,g_2=\varphi(s_2)$ and $\epsilon_k$ is a scaling factor. As we have mentioned above, it is hard to regress the Euclidean distance in the adjacency space to the shortest transition distance accurately, and we only need to ensure a binary relation for implementing the adjacency constraint, i.e., $\lVert \psi_\phi(g_1) - \psi_\phi(g_2) \rVert_2 > \epsilon_k$ for $d_{\mathrm{st}}(s_1,s_2)>k$, and $\lVert \psi_\phi(g_1) - \psi_\phi(g_2) \rVert_2 < \epsilon_k$ for $d_{\mathrm{st}}(s_1,s_2)<k$, as shown in Figure~\ref{fig:distance_learning}.
Inspired by modern metric learning approaches~\cite{hadsell_dimensionality_2006}, we adopt a contrastive-like loss function for this distillation process:
\begin{equation}
\begin{aligned}
    \mathcal{L}_{\mathrm{dis}}(\phi) &= \mathbb{E}_{s_i,s_j\in\mathcal{S}} \left[\,l \cdot \max\left(\lVert \psi_\phi(g_i) - \psi_\phi(g_j) \rVert_2 - \epsilon_k,\,0\right)\right. \\
    +& \left.(1 - l) \cdot \max\left(\epsilon_k + \delta - \lVert \psi_\phi(g_i) - \psi_\phi(g_j) \lVert_2,\,0\right)\right],
\end{aligned}
\label{equ:contrastive}
\end{equation}
where $g_i = \varphi(s_i)$, $g_j = \varphi(s_j)$, and a hyper-parameter $\delta>0$ is used to create a gap between the embeddings. $l\in\{0,1\}$ represents the label indicating $k$-step adjacency derived from the $k$-step adjacency matrix. Equation~\eqref{equ:contrastive} penalizes adjacent state embeddings ($l=1$) with large Euclidean distances in the adjacency space and non-adjacent state embeddings ($l=0$) with small Euclidean distances.
In practice, we use states evenly-sampled from the adjacency matrix to approximate the expectation, and train the adjacency network each time after the adjacency matrix is updated with newly-sampled trajectories.

Although the construction of an adjacency matrix limits our method to tasks with tabular state spaces, our method can also handle continuous state spaces using goal space discretization (see our continuous control experiments in Section~\ref{sec:experiment}). For applications with vast state spaces, constructing a complete adjacency matrix will be problematic, but it is still possible to scale our method to these scenarios using specific feature construction or dimension reduction methods~\cite{nair_visual_2018,nasiriany_planning_2019,ecoffet_go-explore:_2019}, or substituting the distance learning procedure with more accurate distance learning algorithms~\cite{florensa_self-supervised_2019,eysenbach_search_2019} at the cost of some learning efficiency. We consider possible extensions in this direction as our future work.


\subsection{Combining HRL and Adjacency Constraint}

With a learned adjacency network $\psi_\phi$, we can now incorporate the adjacency constraint into the goal-conditioned HRL framework.
According to Equation~\eqref{equ:formulation_unconstrained}, we introduce an adjacency loss $\mathcal{L_\mathrm{adj}}$ to replace the original strict adjacency constraint and minimize the following high-level objective:
\begin{equation}
\mathcal{L}_\mathrm{high}(\theta_h) = -\mathbb{E}_{\pi^h_{\theta_h}}\sum_{t=0}^{T-1}\left(\gamma^{t} r_{kt}^h - \eta\cdot\mathcal{L}_\mathrm{adj}\right),
\label{equ:total_loss}
\end{equation}
where $\eta$ is the balancing coefficient, and $\mathcal{L_\mathrm{adj}}$ is derived by replacing $d_{\mathrm{st}}$ with $\tilde{d}_{\mathrm{st}}$ defined by Equation~\eqref{equ:adjacency_net} in the second term of Equation~\eqref{equ:formulation_unconstrained}:
\begin{equation}
\mathcal{L}_\mathrm{adj}(\theta_h) = H\left(\tilde{d}_{\mathrm{st}}\left(s_{kt},\varphi^{-1}(g_{kt})|\phi\right),k\right) \propto \max\left(\lVert \psi_\phi(\varphi(s_{kt})) - \psi_\phi(g_{kt}) \rVert_2 - \epsilon_k,\,0\right),
\label{equ:goal_loss}
\end{equation}
where $g_{kt}\sim \pi^h_{\theta_h}(g|s_{kt})$. 
Equation~\eqref{equ:goal_loss} will output a non-zero value when the generated subgoal and the current state have an Euclidean distance larger than $\epsilon_k$ in the adjacency space, indicating non-adjacency. It is thus consistent with the $k$-step adjacency constraint. In practice, we plug $\mathcal{L}_{\mathrm{adj}}$ as an extra loss term into the original policy loss term of a specific high-level RL algorithm, e.g., TD error for temporal-difference learning methods.

\begin{figure}
    \centering
    \subfigure[]{\includegraphics[width=0.16\linewidth]{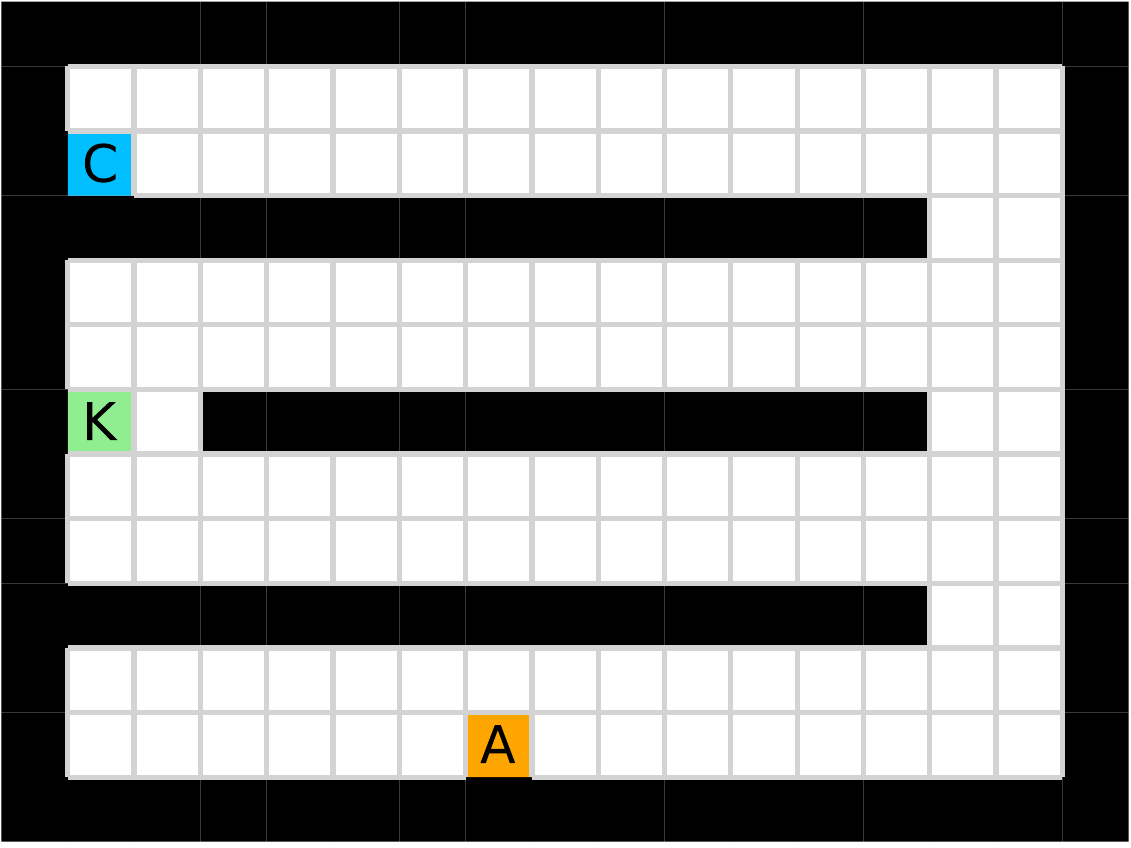}}
    \hspace{0.5em}
    \subfigure[]{\includegraphics[width=0.16\linewidth]{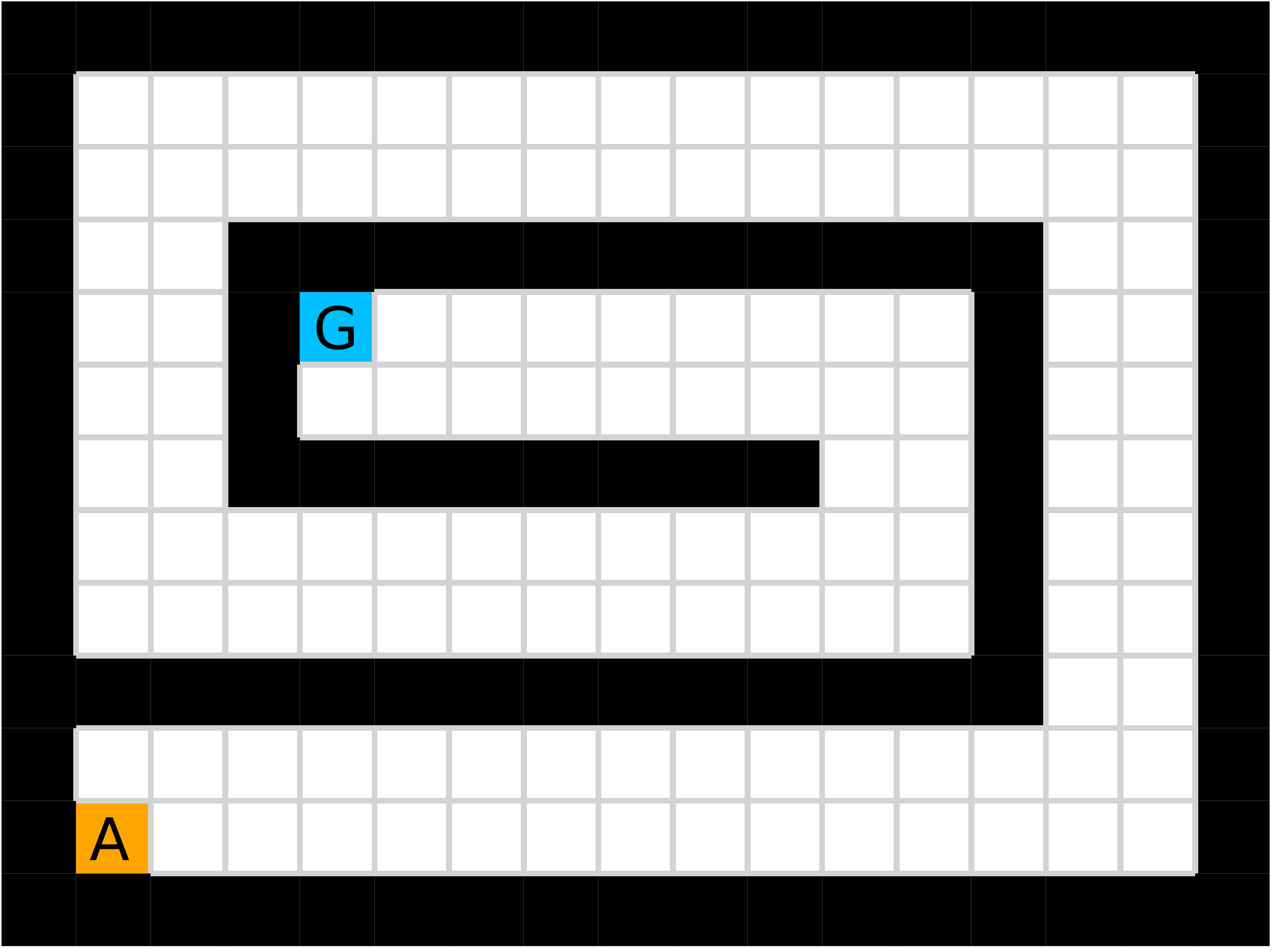}}
    \hspace{0.5em}
    \subfigure[]{\includegraphics[width=0.15\linewidth]{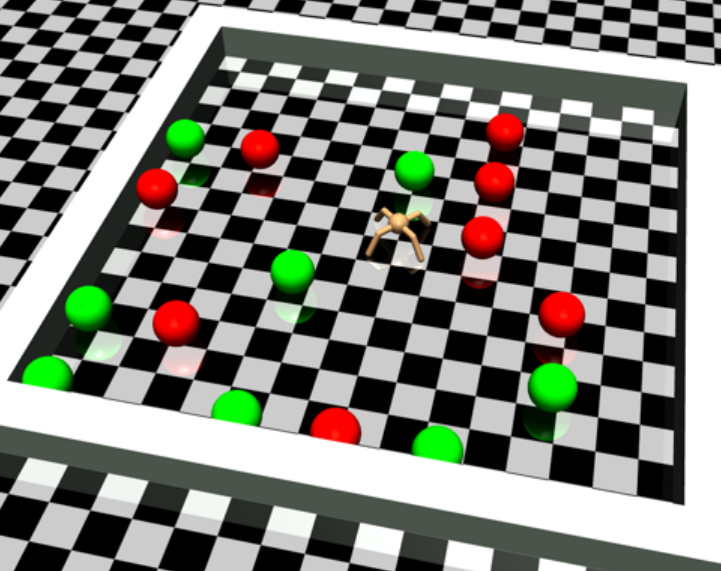}}
    \hspace{0.5em}
    \subfigure[]{\includegraphics[width=0.185\linewidth]{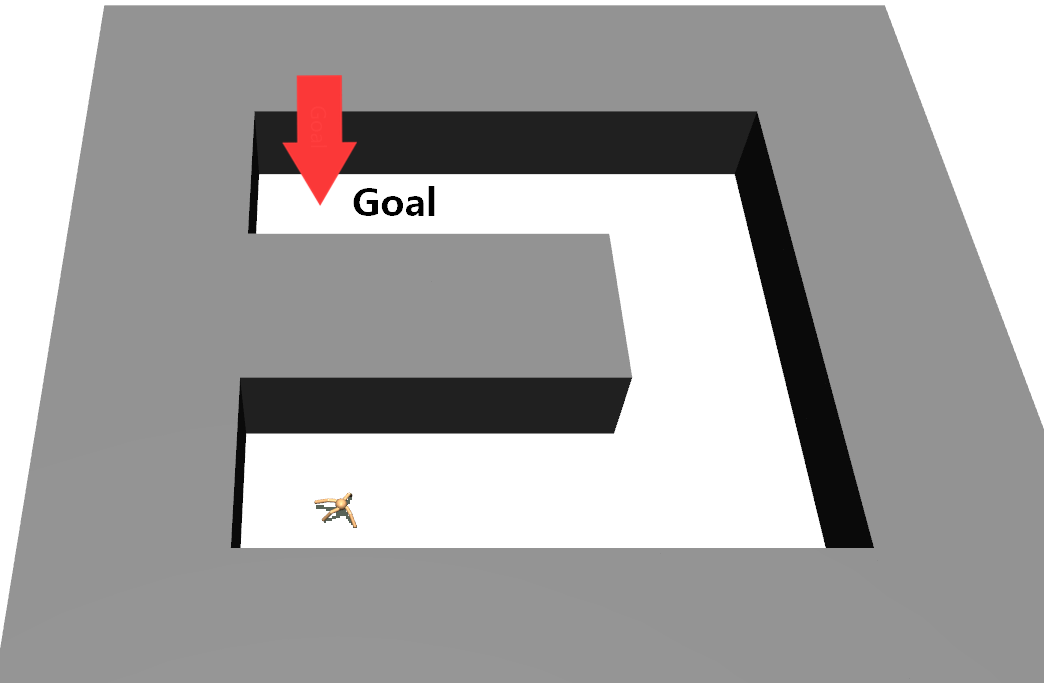}}
    \hspace{0.5em}
    \subfigure[]{\includegraphics[width=0.2\linewidth]{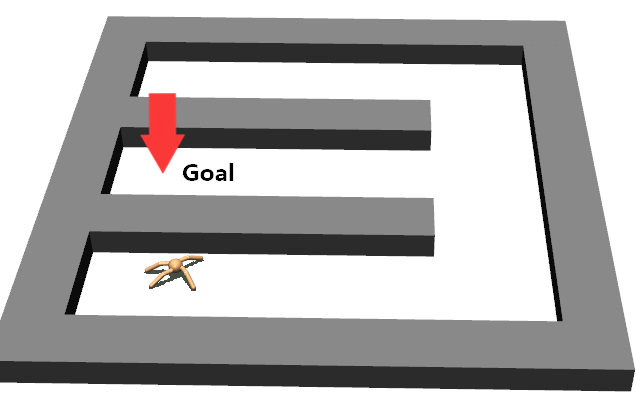}}
    \caption{Environments used in our experiments. \textbf{(a)}~Key-Chest. The agent (A) starts from a random position and needs to pick up the key (K) first, then uses the key to open the chest (C). \textbf{(b)}~Maze. The agent (A) starts from a fixed position and needs to reach the final goal (G) with dense rewards. \textbf{(c)}~Ant Gather. The ant robot starts from a fixed position and needs to collect apples (green) and avoid bombs (red) (the figure is adapted from Duan et al.~\cite{duan_benchmarking_2016}). \textbf{(d)}~Ant Maze. The ant robot starts from a fixed position and needs to reach a target position in a maze with dense rewards. \textbf{(e)}~Ant Maze Sparse. The ant robot starts from a random position and needs to reach a target position in a maze with sparse rewards.}
    \label{fig:env}
\end{figure}

\section{Experimental Evaluation}
\label{sec:experiment}
We have presented our method of Hierarchical Reinforcement learning with $k$-step Adjacency Constraint (HRAC).
Our experiments are designed to answer the following questions: (1)~Can HRAC promote the generation of adjacent subgoals? (2)~Can HRAC improve the sample efficiency and overall performance of goal-conditioned HRL? (3)~Can HRAC outperform other strategies that may also improve the learning efficiency of hierarchical agents, e.g., hindsight experience replay~\cite{andrychowicz_hindsight_2017}?

\subsection{Environment Setup}
We employed two types of tasks with discrete and continuous state and action spaces to evaluate the effectiveness of our method, as shown in Figure~\ref{fig:env}. Discrete tasks include Key-Chest and Maze, where the agents are spawned in grid worlds with injected stochasticity and need to accomplish tasks that require both low-level control and high-level planning. Continuous tasks include Ant Gather, Ant Maze and Ant Maze Sparse, where the first two tasks are widely-used benchmarks in HRL community~\cite{duan_benchmarking_2016,florensa_stochastic_2017,nachum_data-efficient_2018,nachum_near-optimal_2019,levy_learning_2019}, and the third task is a more challenging navigation task with sparse rewards. In all tasks, we used a pre-defined 2-dimensional goal space that represents the $(x,y)$ position of the agent. More details of the environments are in the supplementary material.

\subsection{Comparative Experiments}
\label{subsec:comp_exp}

To comprehensively evaluate the performance of HRAC with different HRL implementations, we employed two different HRL instances for different tasks. On discrete tasks, we used off-policy TD3~\cite{fujimoto_addressing_2018} for high-level training and on-policy A2C, the syncrhonous version of A3C~\cite{mnih_asynchronous_2016}, for the low-level. On continuous tasks, we used TD3 for both the high-level and the low-level training, following prior work~\cite{nachum_data-efficient_2018}, and discretized the goal space to $1\times1$ grids for adjacency learning.

We compared HRAC with the following baselines. (1)~\emph{HIRO}~\cite{nachum_data-efficient_2018}: one of the state-of-the-art goal-conditioned HRL approaches.
(2)~\emph{HIRO-B}: a baseline analagous to HIRO, using binary intrinsic reward for subgoal reaching instead of the shaped reward used by HIRO. (3)~\emph{HRL-HER}: a baseline that employs hindsight experience replay (HER)~\cite{andrychowicz_hindsight_2017} to produce alternative successful subgoal-reaching experiences as complementary low-level learning signals~\cite{levy_learning_2019}. (4)~\emph{Vanilla}: Kulkarni et al.~\cite{kulkarni_hierarchical_2016} used absolute subgoals instead of directional subgoals and adopted a binary intrinsic reward setting. More details of the baselines are in the supplementary material.

\begin{figure*}
    \centering
    \includegraphics[width=0.168\linewidth]{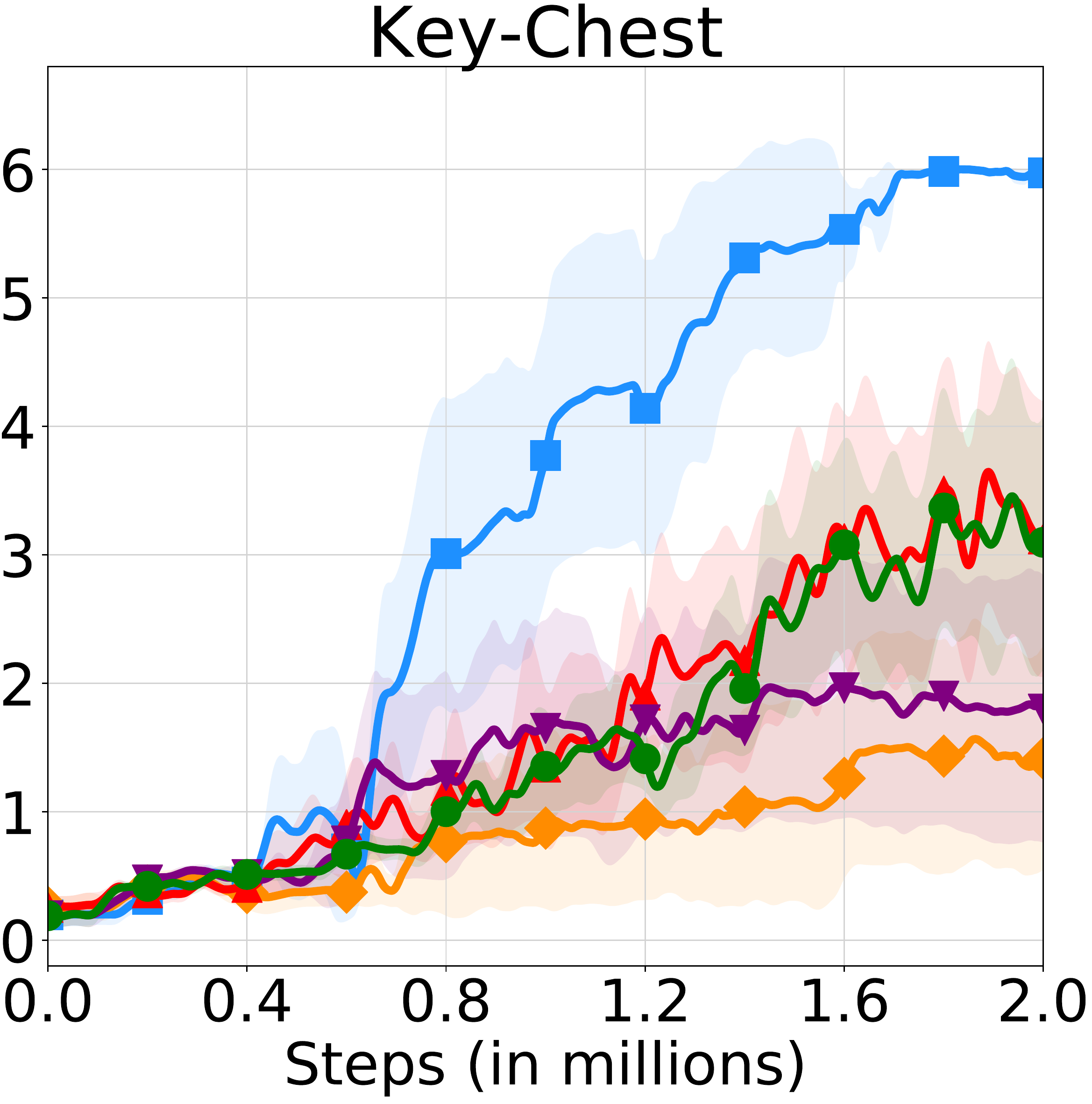}    
    \hspace{0.5em}
    \includegraphics[width=0.168\linewidth]{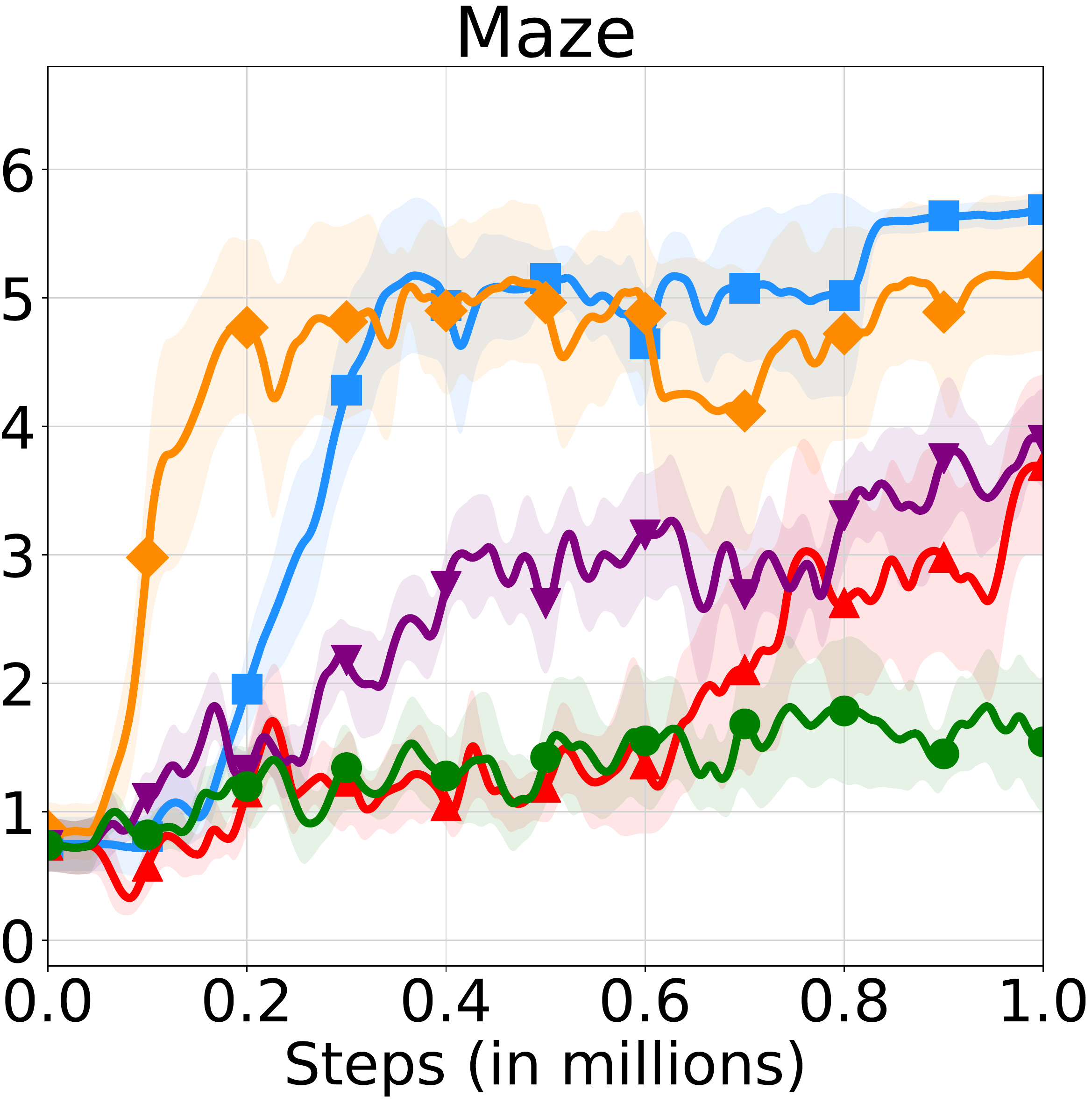}
    \hspace{0.5em}
    \includegraphics[width=0.175\linewidth]{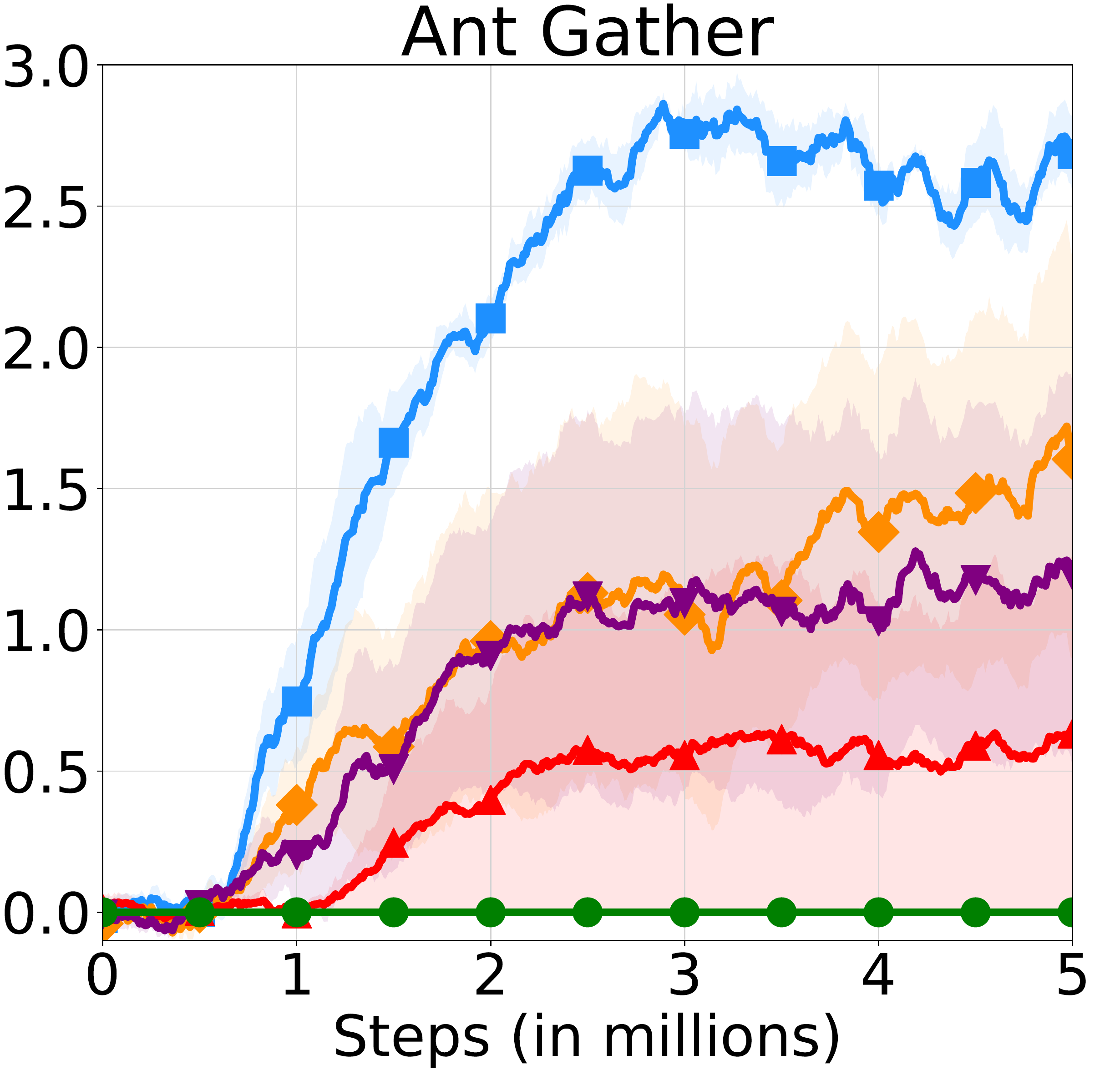}
    \hspace{0.5em}
    \includegraphics[width=0.175\linewidth]{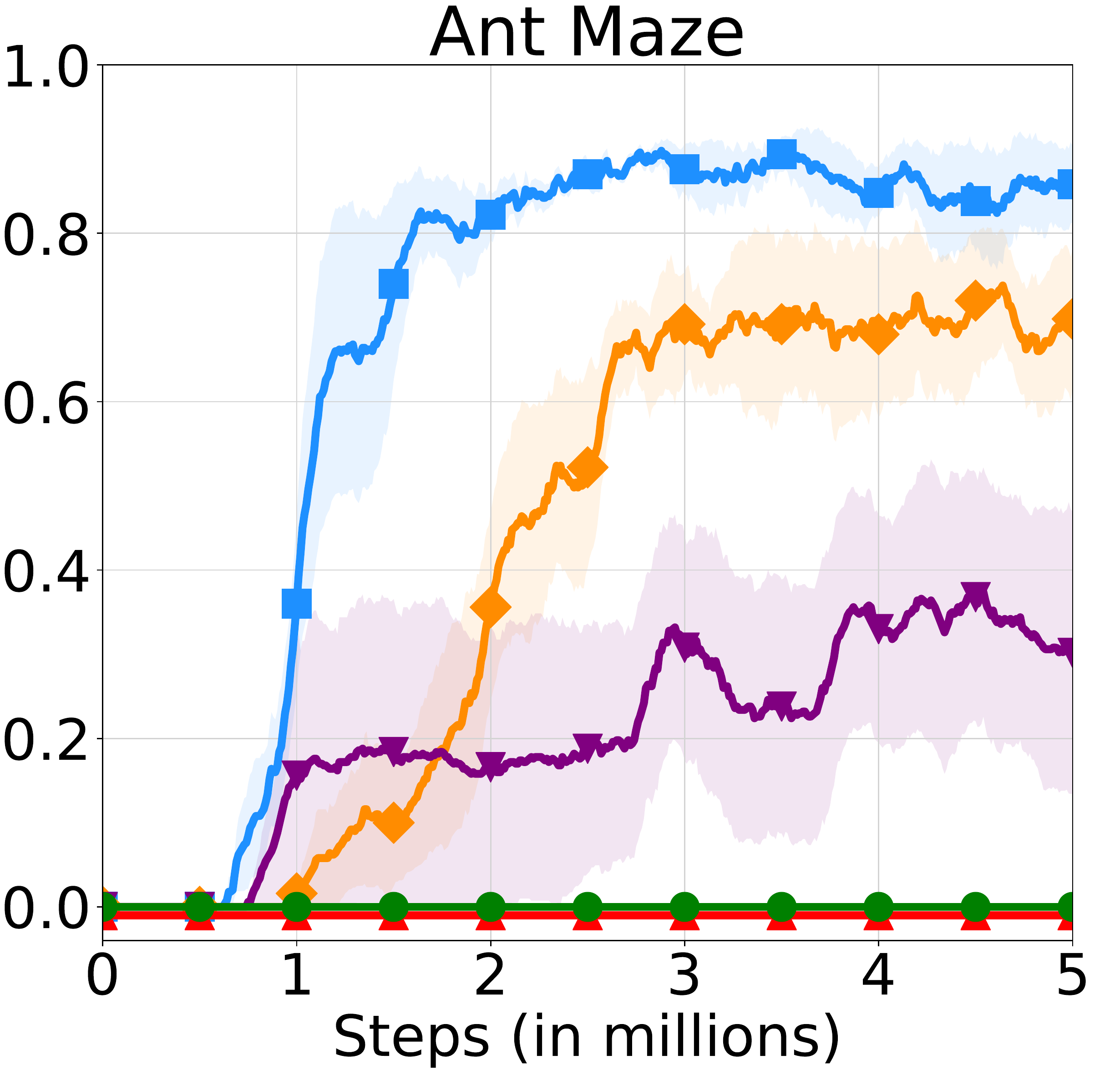}
    \hspace{0.5em}
    \includegraphics[width=0.175\linewidth]{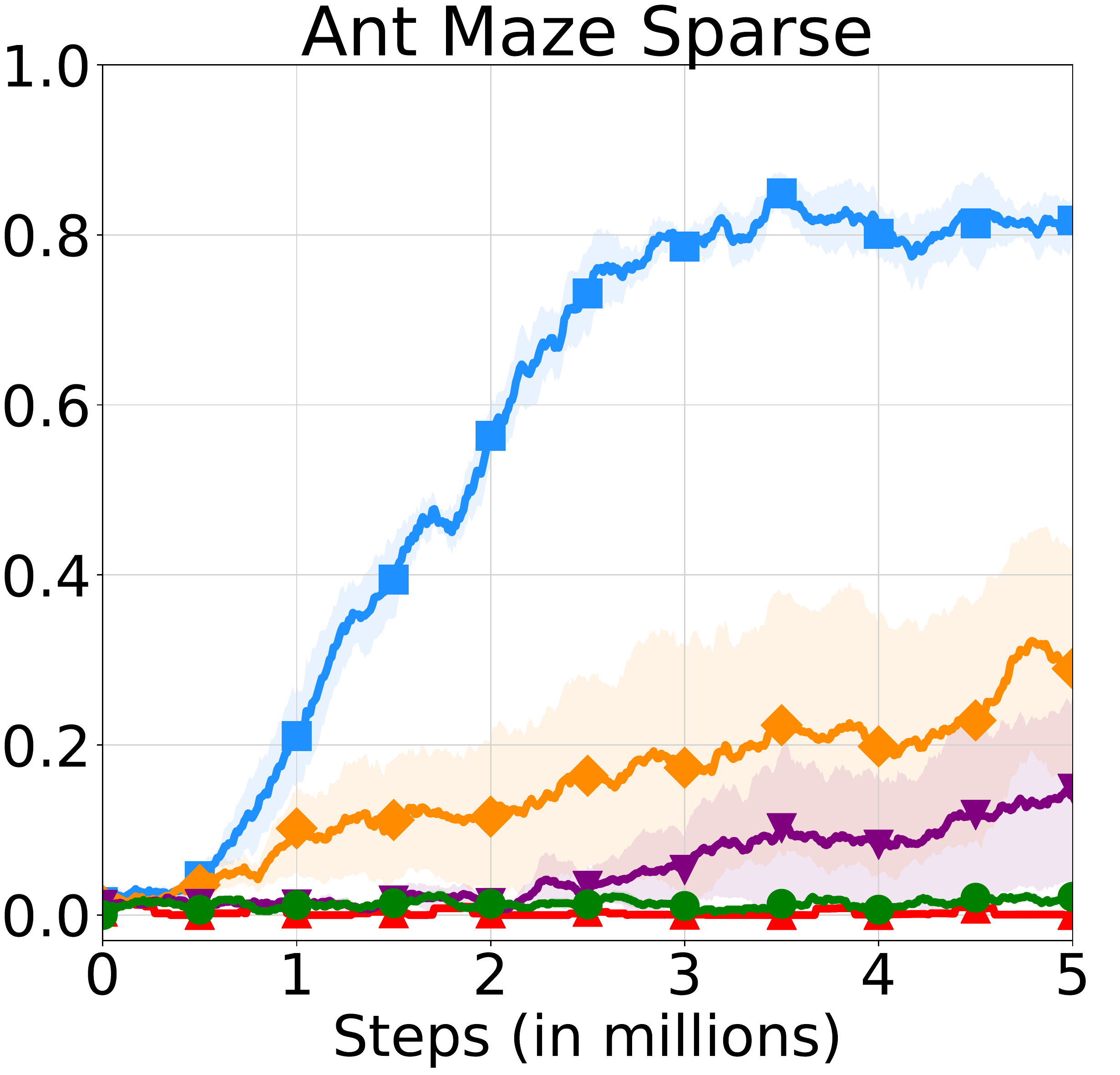} \\
    \vspace{0.5em}
    \includegraphics[width=0.6\linewidth]{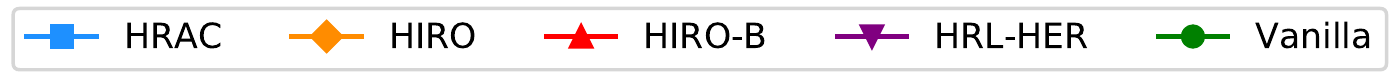}
    \caption{Learning curves of HRAC and baselines on all tasks. Each curve and its shaded region represent mean episode reward and standard error of the mean respectively, averaged over 5 independent trials. All curves have been smoothed equally for visual clarity.}
    \label{fig:results}
\end{figure*}

The learning curves of HRAC and baselines across all tasks are plotted in Figure~\ref{fig:results}. In the Maze task with dense rewards, HRAC achieves comparable performance with HIRO and outperforms other baselines, while in other tasks HRAC consistently surpasses all baselines both in sample efficiency and asymptotic performance. We note that the performance of the baseline HRL-HER matches the results in the previous study~\cite{nachum_data-efficient_2018} where introducing hindsight techniques often degrades the performance of HRL, potentially due to the additional burden introduced on low-level training.

\begin{figure}[t]
\centering
\begin{minipage}{0.53\linewidth}
{
\centering
\includegraphics[width=0.3\linewidth]{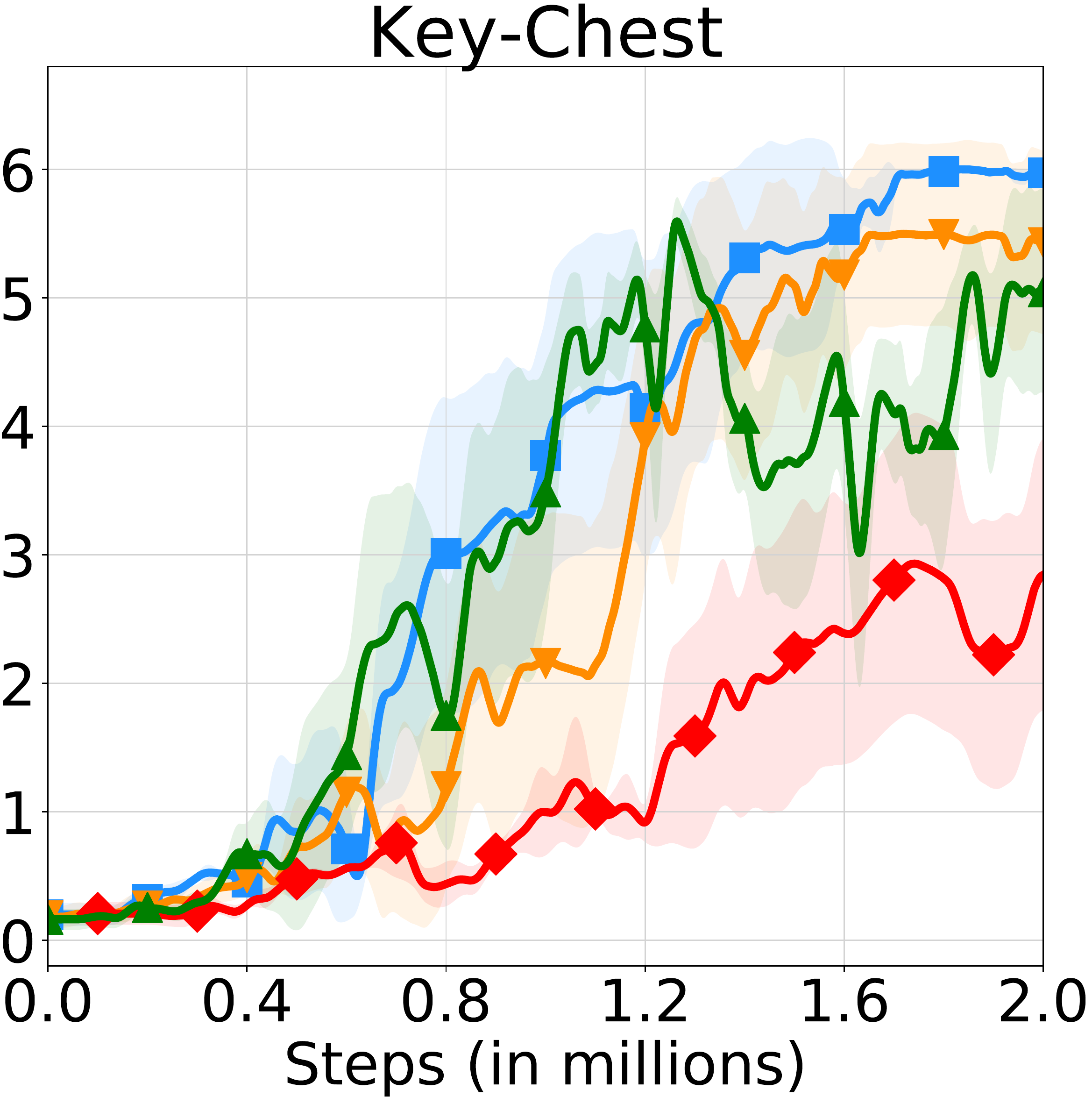}
\hspace{0.1em}
\includegraphics[width=0.3\linewidth]{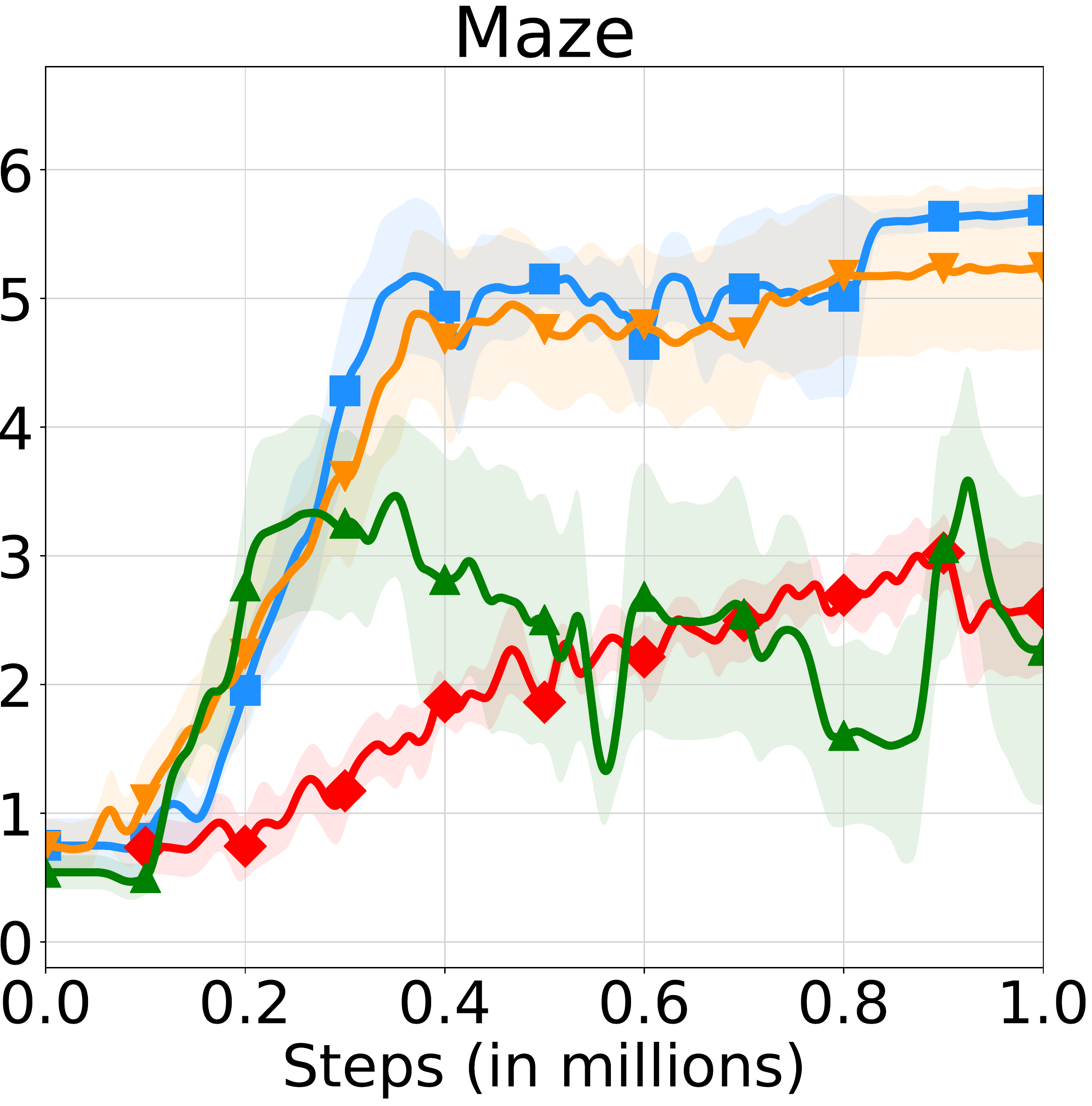}
\hspace{0.1em}
\includegraphics[width=0.31\linewidth]{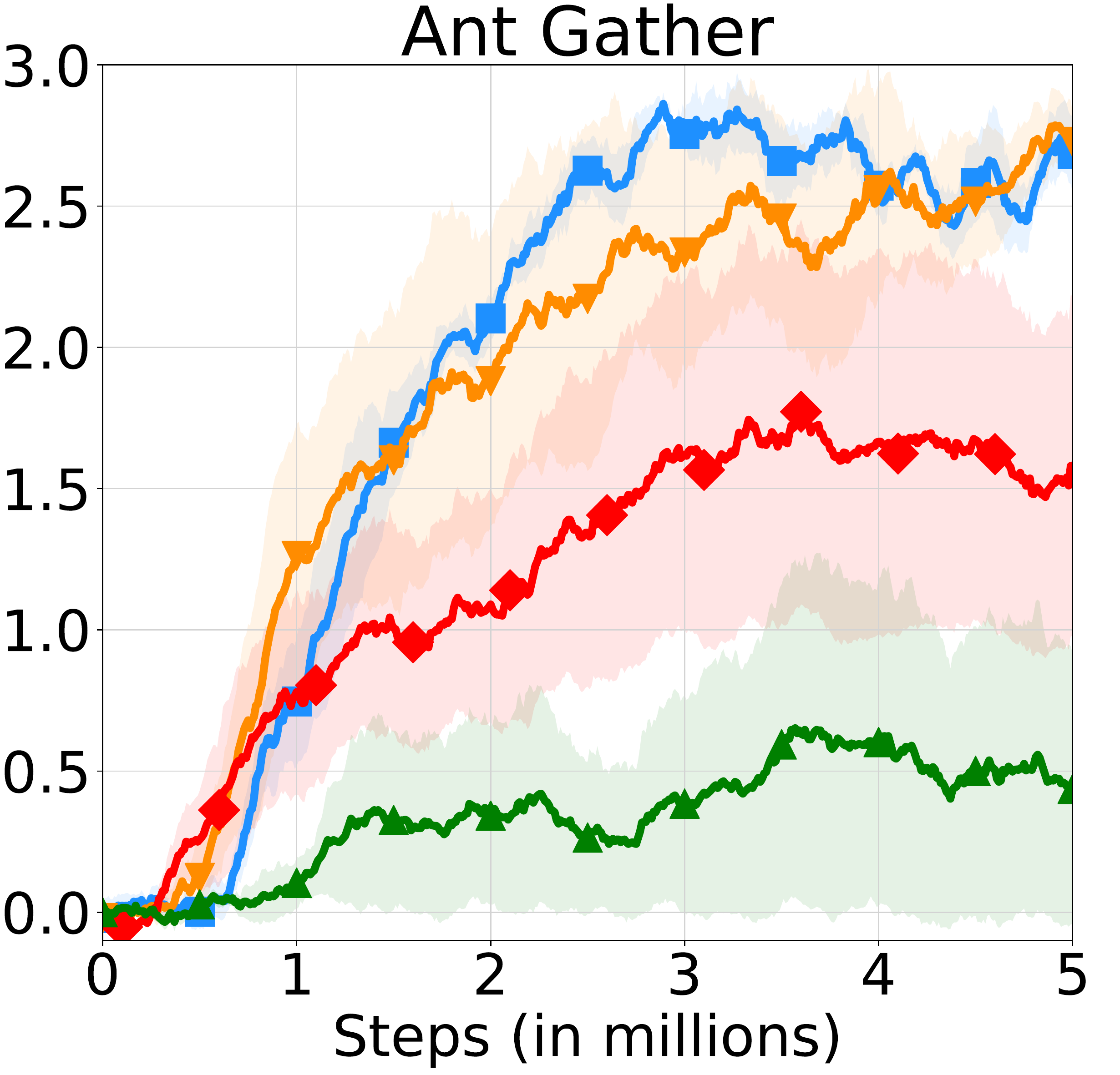} \\
\vspace{0.5em}
\includegraphics[width=0.94\linewidth]{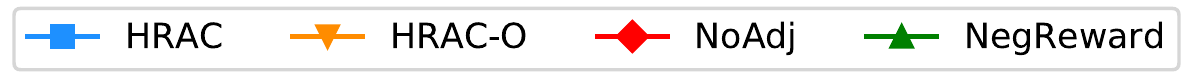}
\caption{Learning curves in the ablation study, averaged over 5 independent trials.}
\label{fig:ablation}
}
\end{minipage}
\hspace{0.5em}
\begin{minipage}{0.439\linewidth}{
\centering
\includegraphics[width=0.365\linewidth]{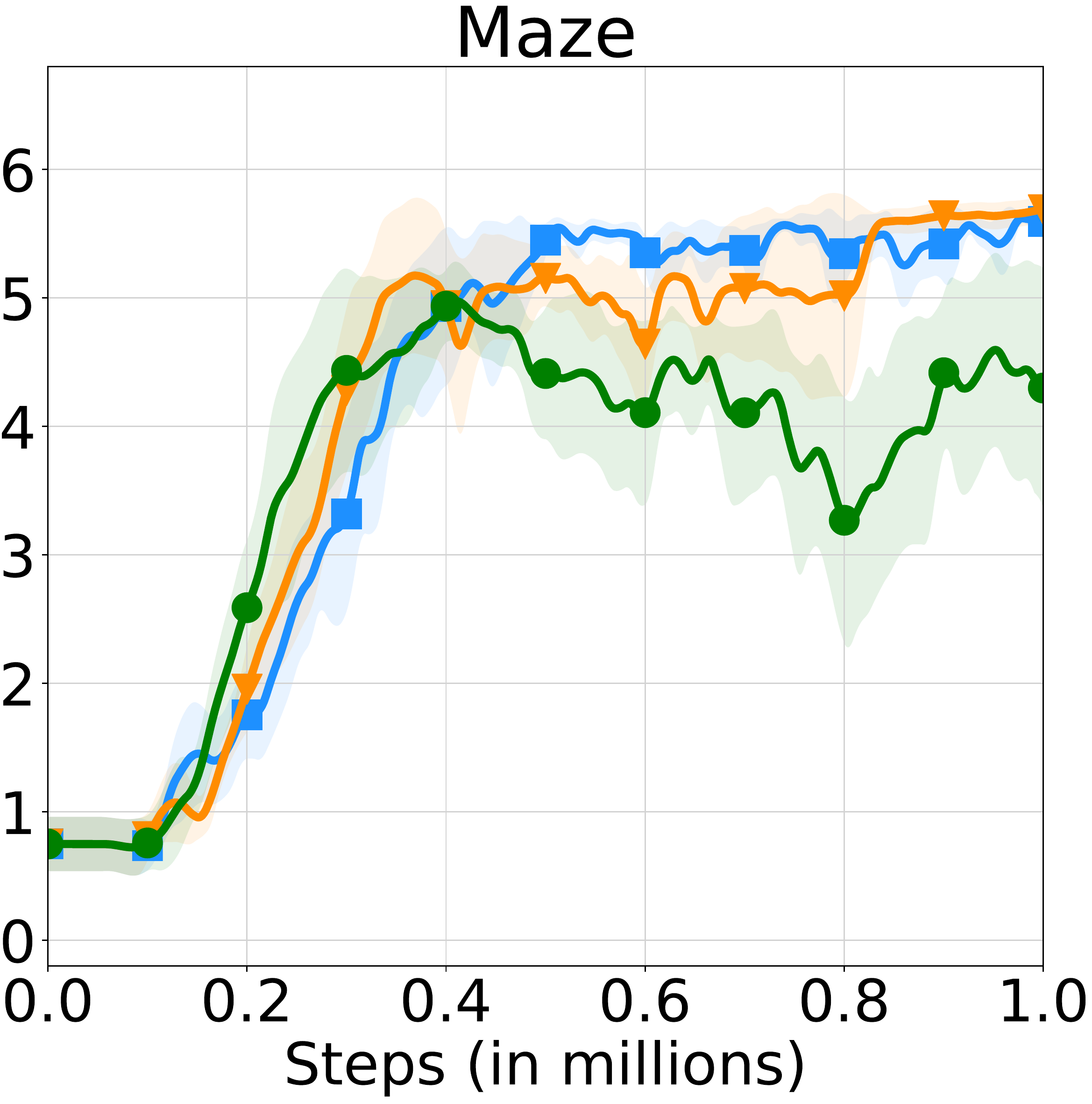}
\hspace{0.1em}
\includegraphics[width=0.385\linewidth]{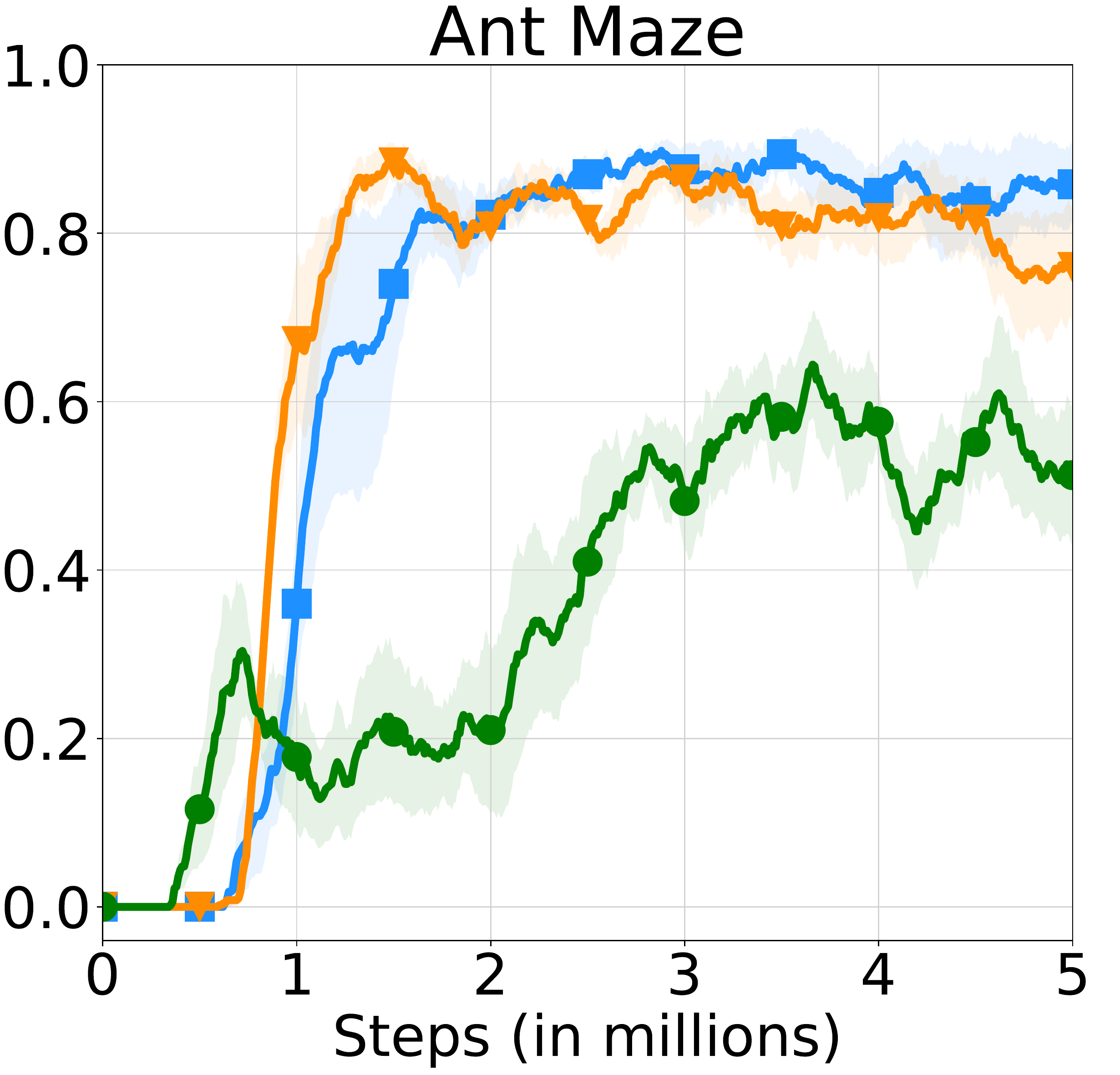} \\
\vspace{0.5em}
\includegraphics[width=0.7\linewidth]{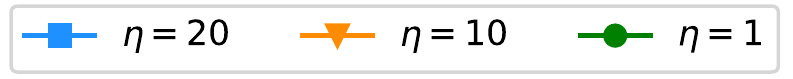}
\caption{Learning curves with different balancing coefficients.}
\label{fig:param}
}
\end{minipage}
\end{figure}

\subsection{Ablation Study and Visualizations}
\label{subsec:ablation}
We also compared HRAC with several variants to investigate the effectiveness of each component. (1)~\emph{HRAC-O}: An oracle variant that uses a perfect adjacency matrix directly obtained from the environment. We note that compared to other methods, this variant uses the additional information that is not available in many applications. (2)~\emph{NoAdj}: A variant that uses an adjacency training method analagous to the work of Savinov et al.~\cite{savinov_semi-parametric_2018,savinov_episodic_2019}, where no adjacency matrix is maintained. The adjacency network is trained using state-pairs directly sampled from stored trajectories, under the same training budget as HRAC. (3)~\emph{NegReward}: This variant implements the $k$-step adjacency constraint by penalizing the high-level with a negative reward when it generates non-adjacent subgoals, which is used by HAC~\cite{levy_learning_2019}.





We provide learning curves of HRAC and these variants in Figure~\ref{fig:ablation}. In all tasks, HRAC yields similar performance with the oracle variant HRAC-O while surpassing the NoAdj variant by a large margin, exhibiting the effectiveness of our adjacency learning method. Meanwhile, HRAC achieves better performance than the NegReward variant, suggesting the superiority of implementing the adjacency constraint using a differentiable adjacency loss, which provides a stronger supervision than a penalty. We also empirically studied the effect of different balancing coefficients $\eta$. Results are shown in Figure~\ref{fig:param}, which suggests that generally a large $\eta$ can lead to better and more stable performance.

\begin{figure}
    \centering
    \begin{minipage}{0.99\linewidth}
    {\centering
    \includegraphics[width=0.17\linewidth]{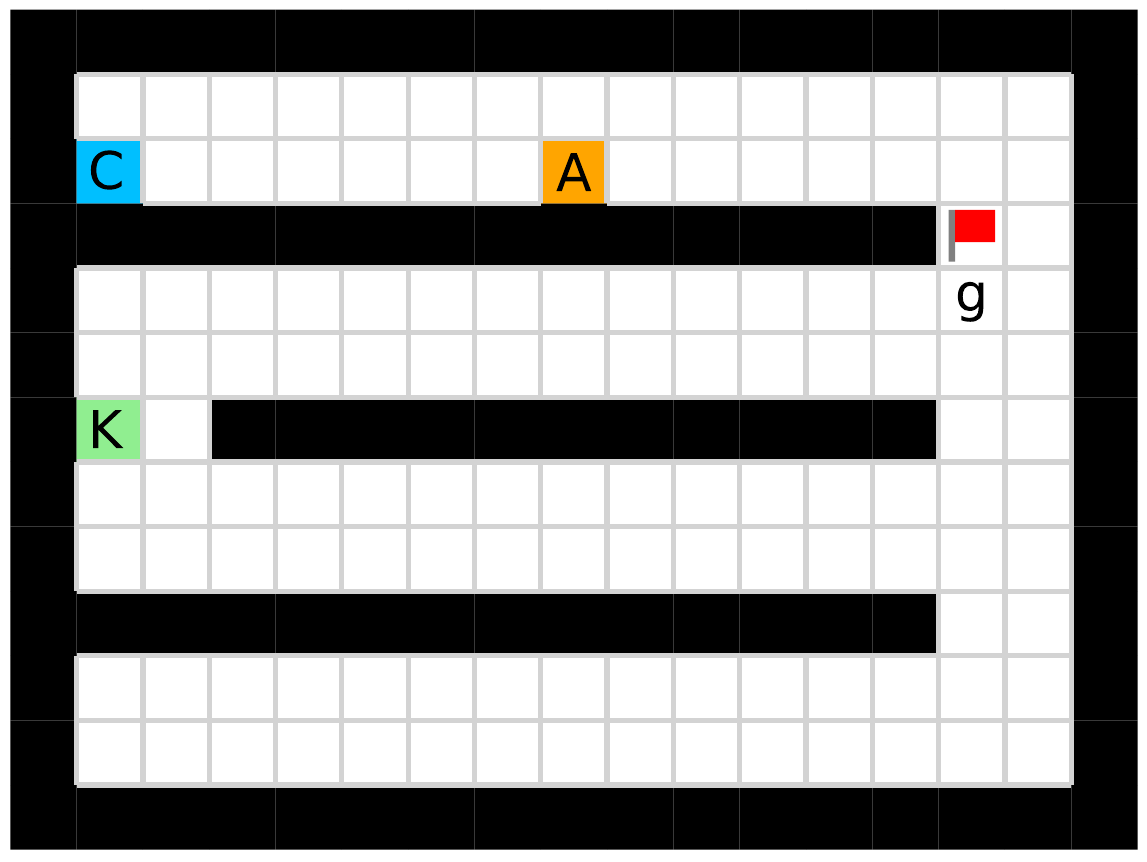}
    \hspace{0.4em}
    \includegraphics[width=0.17\linewidth]{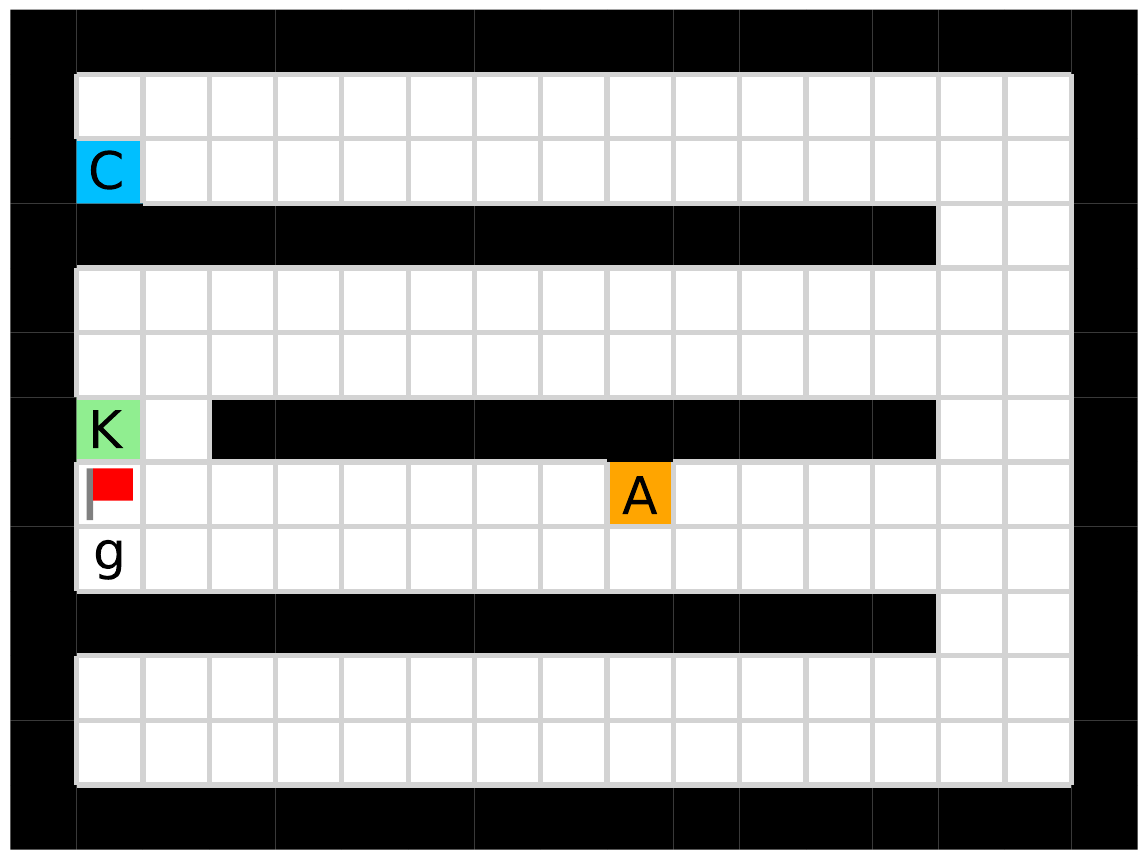}
    \hspace{0.4em}
    \includegraphics[width=0.17\linewidth]{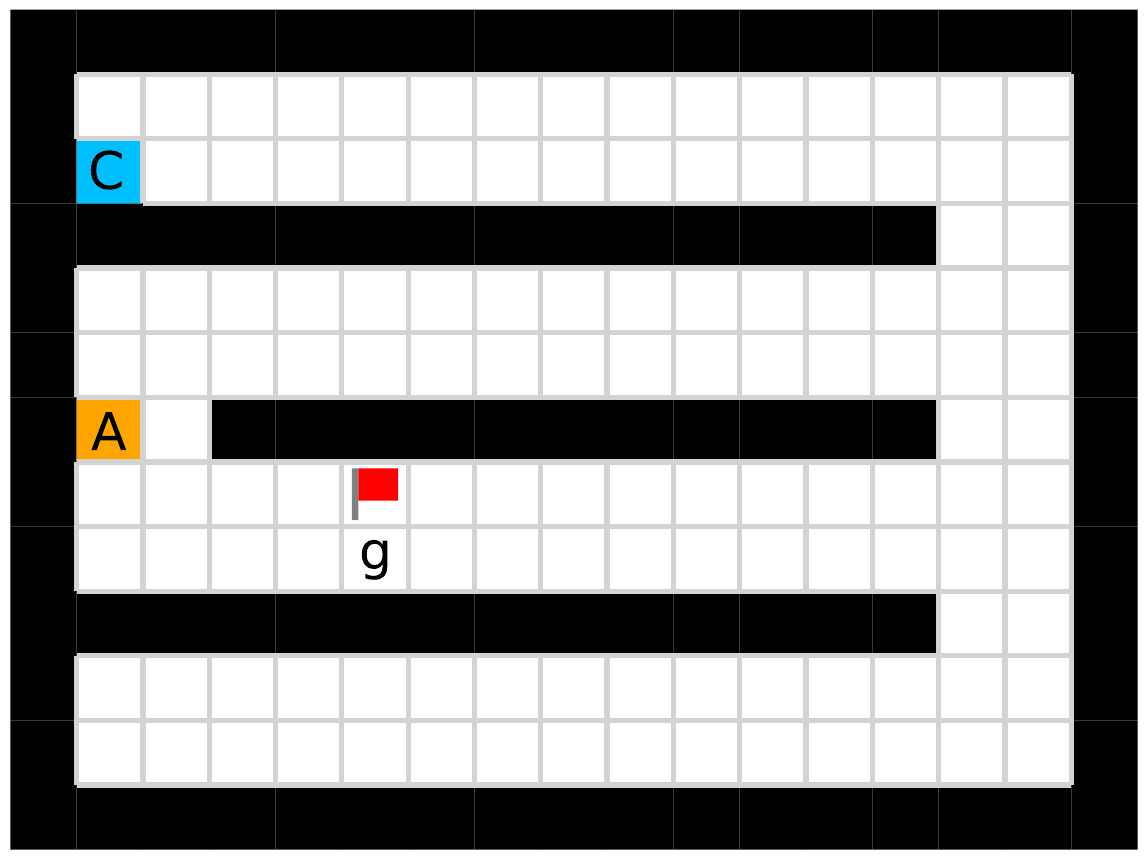}
    \hspace{0.4em}
    \includegraphics[width=0.17\linewidth]{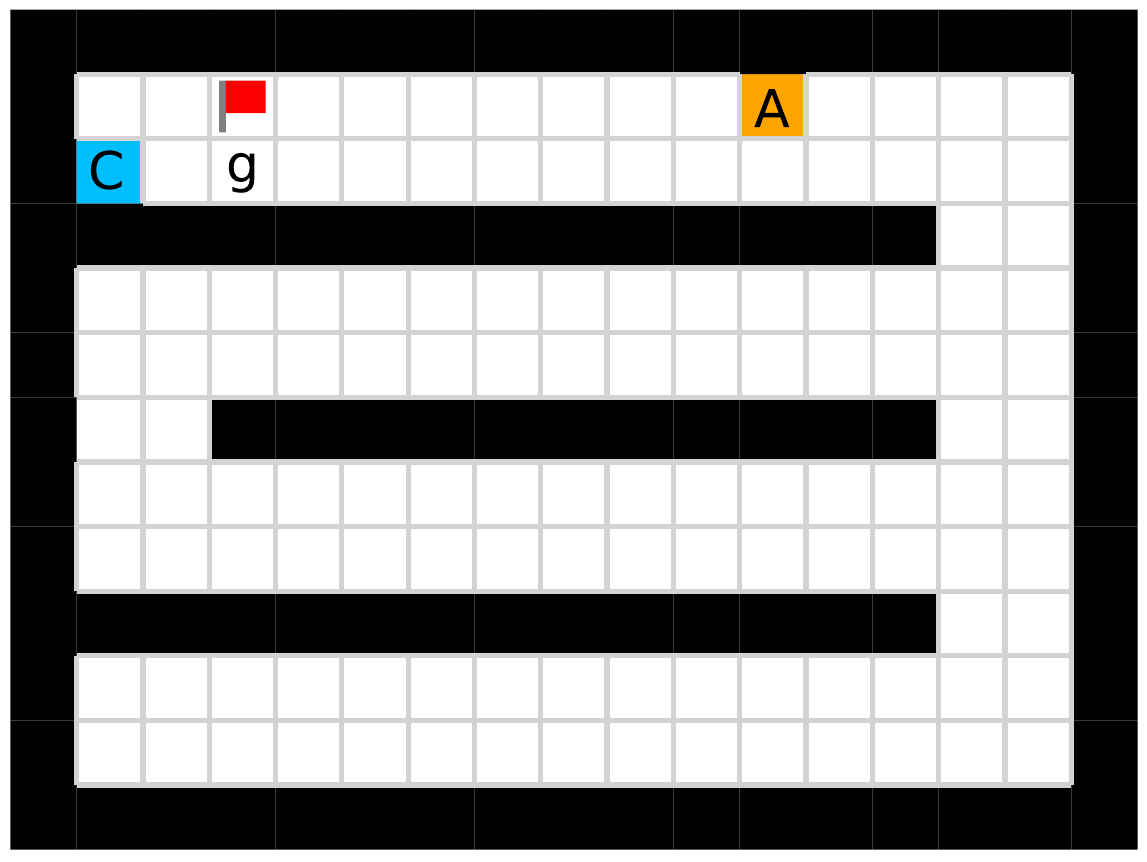}
    \hspace{0.5em}
    \includegraphics[width=0.19\linewidth]{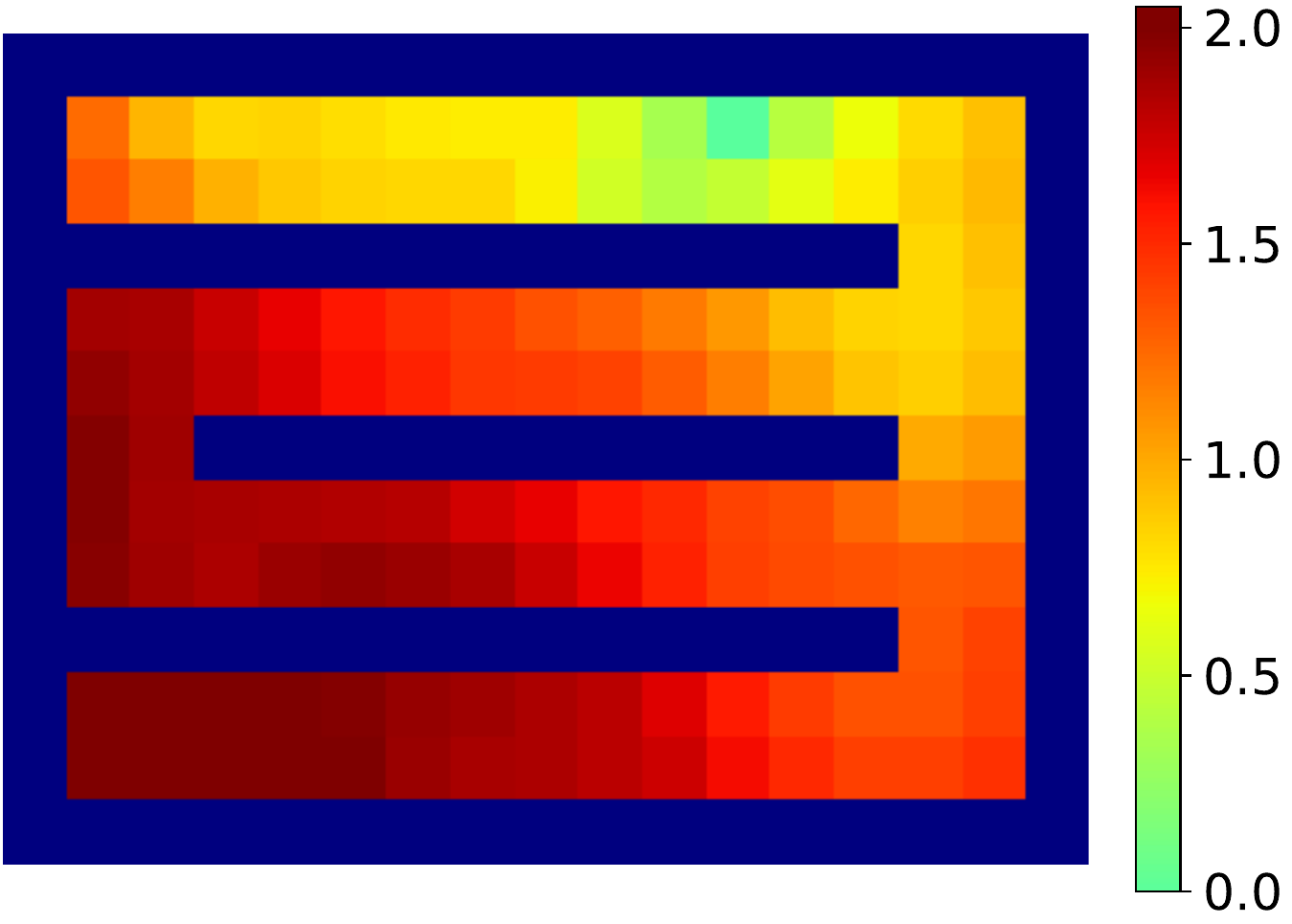} \\}
    \hspace{0.1em}\includegraphics[width=0.81\linewidth]{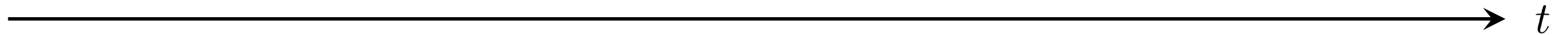}
    \end{minipage}
    \caption{Visualizations on the Key-Chest task, based on a single evaluation run. The agent (A), key (K), chest (C) and subgoal (g) at four different time steps are plotted. The adjacency heatmap is based on the fourth time step, where colder colors represent smaller shortest transition distances.}
    \label{fig:visualization}
\end{figure}

Finally, we visualize the subgoals generated by the high-level policy and the adjacency heatmap in Figure~\ref{fig:visualization}. Visualizations indicate that the agent does learn to generate adjacent and interpretable subgoals. We provide additional visualizations in the supplementary material.

\subsection{Empirical Study in Stochastic Environments}

\begin{figure}
\centering
\includegraphics[width=0.18\linewidth]{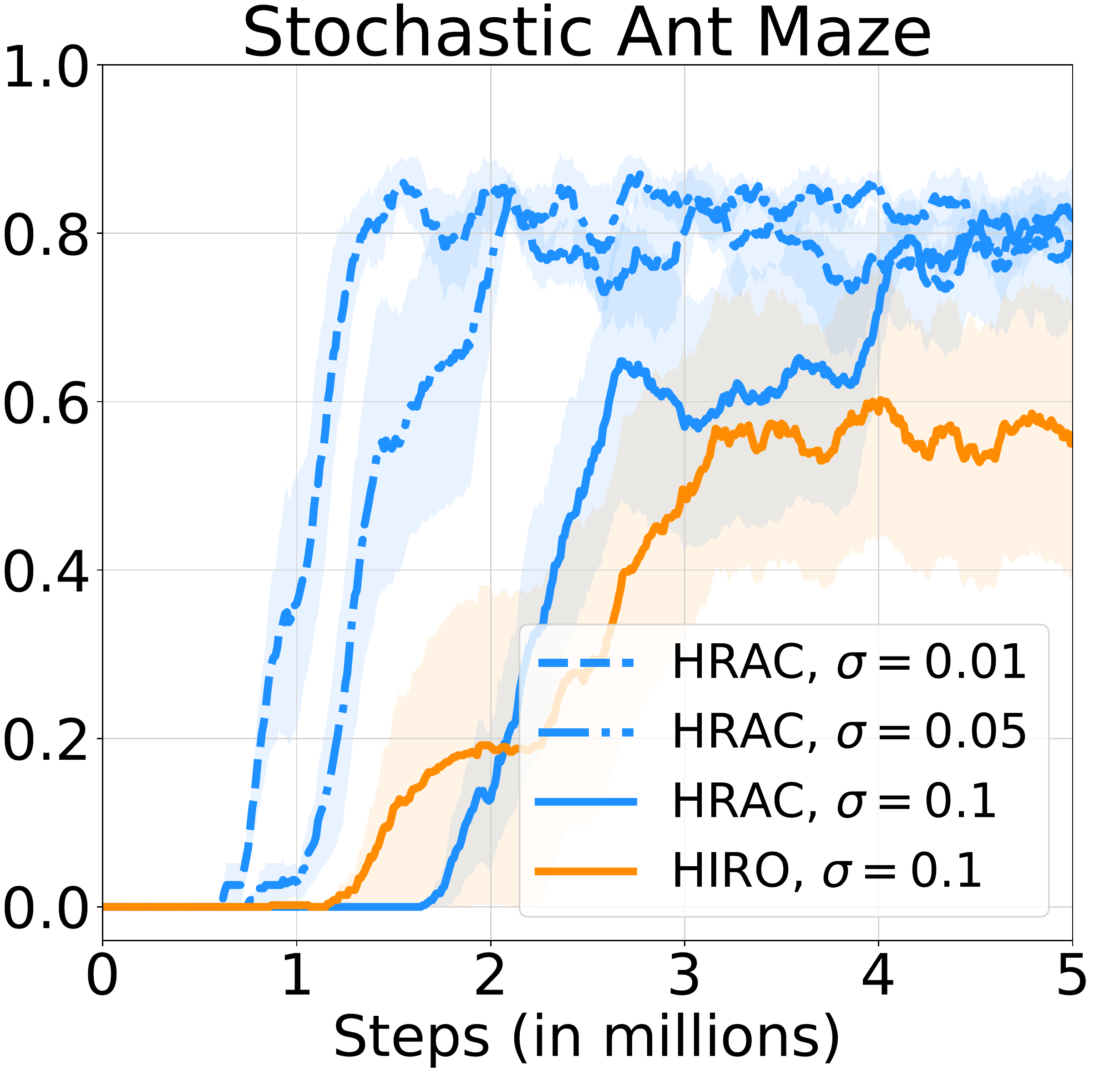}
\hspace{0.3em}
\includegraphics[width=0.18\linewidth]{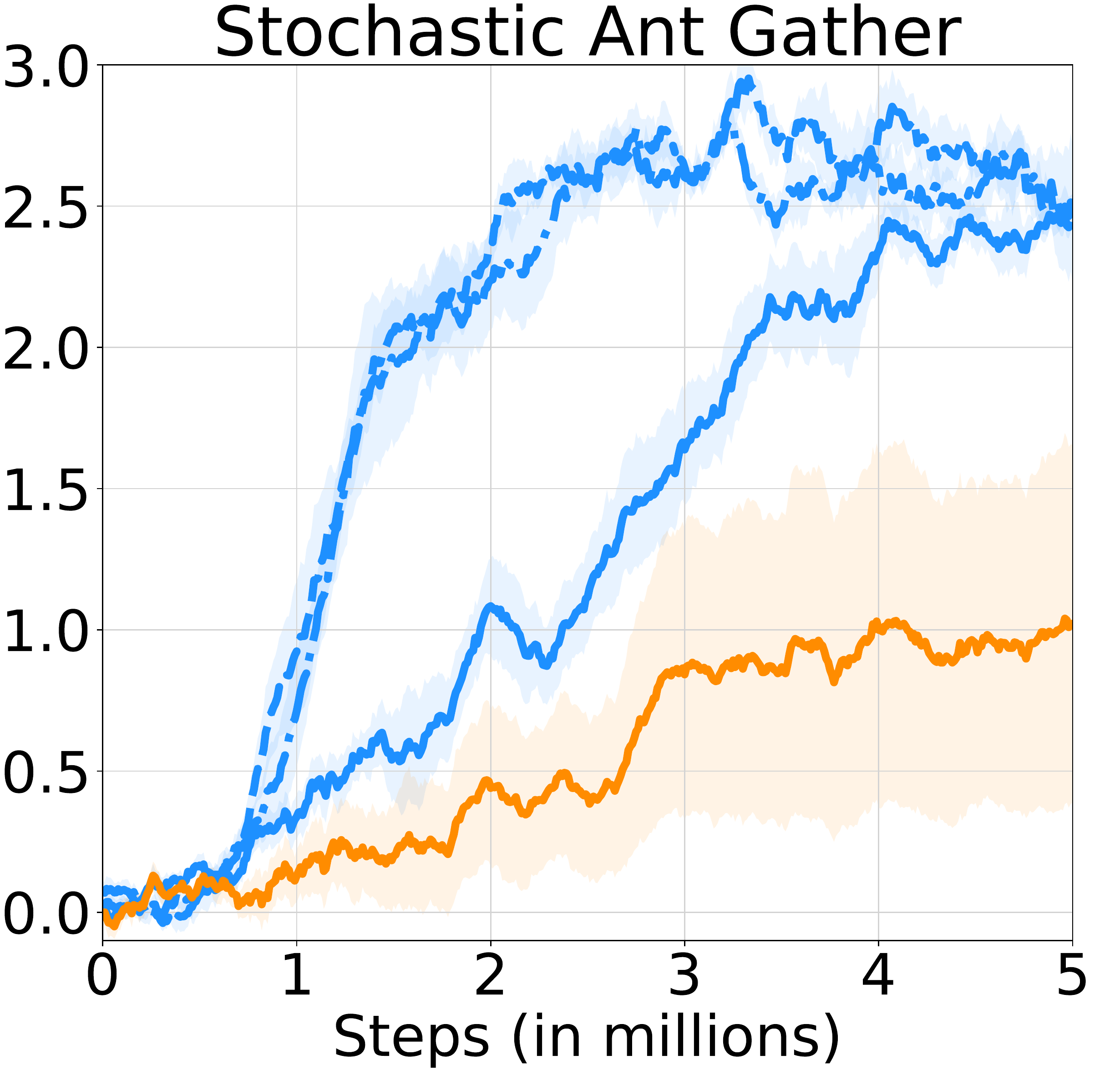}
\hspace{0.3em}
\includegraphics[width=0.18\linewidth]{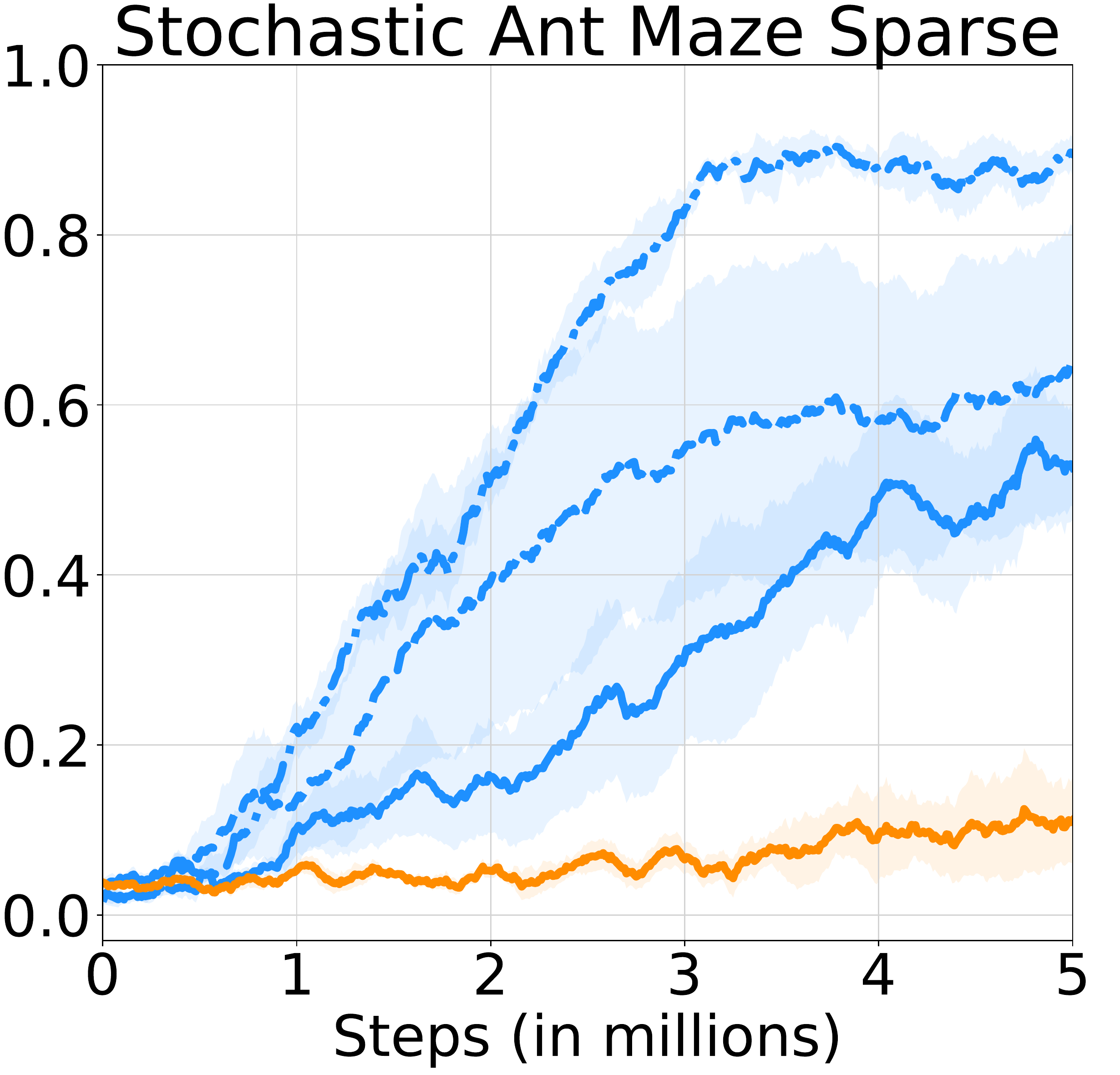}
\hspace{0.3em}
\caption{Learning curves in stochastic environments, averaged over 5 independent trials.}
\label{fig:stochastic}
\end{figure}

To empirically verify the stochasticity robustness of HRAC, we applied it to a set of stochastic tasks, including stochastic Ant Gather, Ant Maze and Ant Maze Sparse tasks, which are modified from the original ant tasks respectively. Concretely, we added Gaussian noise with different standard deviations $\sigma$ to the $(x,y)$ position of the ant robot at every step, including $\sigma=0.01,\sigma=0.05$ and $\sigma=0.1$, representing increasing environmental stochasticity. In these tasks we compare HRAC with the baseline HIRO, which has exhibited generally better performance than other baselines, in the most noisy scenario when $\sigma=0.1$. As displayed in Figure~\ref{fig:stochastic}, HRAC achieves similar asymptotic performances with different noise magnitudes in stochastic Ant Gather and Ant Maze tasks and consistently outperforms HIRO, exhibiting robustness to stochastic environments.

\section{Related Work}

Effectively learning policies with multiple hierarchies has been a long-standing problem in RL.
Goal-conditioned HRL~\cite{dayan_feudal_1993,schmidhuber_planning_1993,kulkarni_hierarchical_2016,vezhnevets_feudal_2017,nachum_data-efficient_2018,levy_learning_2019} aims to resolve this problem with a framework that separates high-level planning and low-level control using subgoals.
Recent advances in goal-conditioned HRL mainly focus on improving the learning efficiency of this framework. Nachum et al.~\cite{nachum_data-efficient_2018,nachum_near-optimal_2019} proposed an off-policy correction technique to stabilize training, and addressed the problem of goal space representation learning using a mutual-information-based objective. However, the subgoal generation process in their approaches is unconstrained and supervised only by the external reward, and thus these methods may still suffer from training inefficiency. Levy et al.~\cite{levy_learning_2019} used hindsight techniques~\cite{andrychowicz_hindsight_2017} to train multi-level policies in parallel and also penalized the high-level for generating subgoals that the low-level failed to reach. However, their method has no theoretical guarantee, and they directly obtain the reachability measure from the environment, using the environmental information that is not available in many scenarios.
There are also prior works focusing on unsupervised acquisition of subgoals based on potentially pivotal states~\cite{mcgovern_automatic_2001,simsek_identifying_2005,kulkarni_deep_2016,savinov_semi-parametric_2018,rafati_unsupervised_2019,huang_mapping_2019}. However, these subgoals are not guaranteed to be well-aligned with the downstream tasks and thus are often sub-optimal.

Several prior works have constructed an environmental graph for high-level planning used search nearby graph nodes as reachable subgoals for the low-level~\cite{savinov_semi-parametric_2018,eysenbach_search_2019,huang_mapping_2019,zhang_composable_2018}. However, these approaches hard-coded the high-level planning process based on domain-specific knowledge, e.g., treat the planning process as solving a shortest-path problem in the graph instead of a learning problem, and thus are limited in scalability. Nasiriany et al.~\cite{nasiriany_planning_2019} used goal-conditioned value functions to measure the feasibility of subgoals, but a pre-trained goal-conditioned policy is required. A more general topic of goal generation in RL has also been studied in the literature~\cite{florensa_automatic_2018,nair_visual_2018,ren_exploration_2019}. However, these methods only have a flat architecture and therefore cannot successfully solve tasks that require complex high-level planning.

Meanwhile, our method relates to previous research that studied transition distance or reachability~\cite{pong_temporal_2018,savinov_semi-parametric_2018,savinov_episodic_2019,florensa_self-supervised_2019,hartikainen_dynamical_2020}. Most of these works learn the transition distance based on RL~\cite{pong_temporal_2018,florensa_self-supervised_2019,hartikainen_dynamical_2020}, which tend to have a high learning cost. Savinov et al.~\cite{savinov_semi-parametric_2018,savinov_episodic_2019} proposed a supervised learning approach for reachability learning. However, the metric they learned depends on a certain policy used for interaction and thus could be sub-optimal compared to our learning method. There are also other metrics that can reflect state similarities in MDPs, such as successor represention~\cite{dayan_improving_1993,kulkarni_deep_2016} that depends on both the environmental dynamics and a specific policy, and bisimulation metrics~\cite{ferns_metrics_2004,castro_scalable_2019} that depend on both the dynamics and the rewards. Compared to these metrics, the shortest transition distance depends only on the dynamics and therefore may be seamlessly applied to multi-task settings.


\section{Conclusion}
We present a novel $k$-step adjacency constraint for goal-conditioned HRL framework to mitigate the issue of training inefficiency, with the theoretical guarantee of preserving the optimal policy in deterministic MDPs. We show that the proposed adjacency constraint can be practically implemented with an adjacency network. Experiments on several testbeds with discrete and continuous state and action spaces demonstrate the effectiveness and robustness of our method.

As one of the most promising directions for scaling up RL, goal-conditioned HRL provides an appealing paradigm for handling large-scale problems. However, some key issues involving how to devise effective and interpretable hierarchies remain to be solved, such as how to empower the high-level policy to learn and explore in a more semantically meaningful action space~\cite{nachum_why_2019}, and how to enable the subgoals to be shared and reused in multi-task settings. Other future work includes extending our method to tasks with high-dimensional state spaces, e.g., by encompassing modern representation learning schemes~\cite{higgins_beta-vae:_2017,nachum_near-optimal_2019,srinivas_curl_2020}, and leveraging the adjacency network to improve the learning efficiency in more general scenarios.

\section*{Broader Impact}
This work may promote the research in the field of HRL and RL, and has potential real-world applications such as robotics. The main uncertainty of the proposed method might be the fact that the RL training process itself is somewhat brittle, and may break in counterintuitive ways when the reward function is misspecified.
Also, since the training data of RL heavily depends on the training environments, designing unbiased simulators or real-world training environments is important for eliminating the biases in the data collected by the agents.



\begin{ack}
This work was supported in part by the National Natural Science Foundation of China under Grant 61671266, Grant 61836004, Grant 61836014 and in part by the Tsinghua-Guoqiang research program under Grant 2019GQG0006. The authors would also like to thank the anonymous reviewers for their careful reading and their many insightful comments.
\end{ack}

{\small
\bibliography{ref}
\bibliographystyle{plain}}




\newpage
\appendix

\section{Proofs of Theorems}

\subsection{Proof of Theorem~\ref{theo:low}}

\begin{proof}
Under the assumption that the MDP is deterministic and all states are strongly connected, there exists at least one shortest state trajectory from $s$ to $g$. Without loss of generality, we consider one shortest state trajectory $\tau^* = (s_0,s_1,s_2,\cdots,s_{n-1},s_n)$, where $s_0 = s,\,s_n = \varphi^{-1}(g)$ and $d_{\mathrm{st}}\left(s,\varphi^{-1}(g)\right) = n$. For all $k\in\mathbb{N}_+$ and $k\le d_{\mathrm{st}}\left(s,\varphi^{-1}(g)\right) = n$, let $\tilde{g} = \varphi(s_k)$, and let $\tau = (s_0,s_1,s_2,\cdots,s_k)$ be the $k$-step sub-trajectory of $\tau^*$ from $s_0$ to $s_k$. Since $s_0$ and $s_k$ is connected by $\tau$ in $k$ steps, we have that $d_{\mathrm{st}}\left(s_0,\varphi^{-1}(\tilde{g})\right) = d_{\mathrm{st}}\left(s_0,s_k \right) \le k$, i.e., $\tilde{g}\in\mathcal{G}_A(s,k)$. In the following, we will prove that $\pi^*(s_i,\tilde{g}) = \pi^*(s_i,g),\ \forall\, s_i\in \tau\,(i\ne k)$.

We first prove that the shortest transition distance $d_{\mathrm{st}}$ satisfies the triangle inequality, i.e., consider three arbitrary states $s_1,s_2,s_3\in\mathcal{S}$, then $d_{\mathrm{st}}(s_1,s_3) \le d_{\mathrm{st}}(s_1,s_2) + d_{\mathrm{st}}(s_2,s_3)$: let $\tau^*_{12}$ be one shortest state trajectory between $s_1$ and $s_2$ and let $\tau^*_{23}$ be one shortest state trajectory between $s_2$ and $s_3$. We can concatenate $\tau^*_{12}$ and $\tau^*_{23}$ to form a trajectory $\tau_{13} = (\tau^*_{12},\tau^*_{23})$ that connects $s_1$ and $s_3$. Then, by Definition~\ref{def:distance} we have $d_{\mathrm{st}}(s_1,s_3)\le d_{\mathrm{st}}(s_1,s_2) + d_{\mathrm{st}}(s_2,s_3)$.

Using the triangle inequality, we can prove that the sub-trajectory $\tau = (s_0,s_1,s_2,\cdots,s_k)$ is also a shortest trajectory from $s_0 = s$ to $s_k$: assume that this is not true and there exists a shorter trajectory from $s_0$ to $s_k$. Then, by Definition~\ref{def:distance} we have $d_{\mathrm{st}}(s_0,s_k) < k$. Since $(s_k,s_{k+1},\cdots,s_n)$ is a valid trajectory from $s_k$ to $s_n$, we have $d_{\mathrm{st}}(s_0,s_k) \le n-k$. Applying the triangle inequality, we have $d_{\mathrm{st}}(s_0,s_n) \le d_{\mathrm{st}}(s_0,s_k) + d_{\mathrm{st}}(s_k,s_n) < k + n - k = n$, which is in contradiction with $d_{\mathrm{st}}\left(s,\varphi^{-1}(g)\right) = d_{\mathrm{st}}(s_0,s_n) = n$. Thus, our original assumption must be false, and the trajectory $\tau = (s_0,s_1,s_2,\cdots,s_k)$ is a shortest trajectory from $s_0$ to $s_k$.

Finally, let $\alpha:\mathcal{S}\times\mathcal{S}\rightarrow\mathcal{A}$ be an inverse dynamics model, i.e., given state $s_t$ and the next state $s_{t+1}$, $\alpha(s_t,s_{t+1})$ outputs the action $a_t$ that is performed at $s_t$ to reach $s_{t+1}$. Then, employing Equation~\eqref{equ:goal_conditioned}, for $i=0,1,\cdots,k-1$ we have $\pi^*(s_i,g) = \alpha(s_i,s_{i+1})$ given that $\tau^*$ is a shortest trajectory from $s_0$ to $\varphi^{-1}(g)$, and $\pi^*(s_i,\tilde{g}) = \alpha(s_i,s_{i+1})$ given that $\tau$ is a shortest trajectory from $s_0$ to $\varphi^{-1}(\tilde{g})$. This indicates that $\pi^*(s_i,\tilde{g}) = \pi^*(s_i,g),\ \forall\, s_i\in \tau\,(i\ne k)$.
\end{proof}

\subsection{Proof of Theorem~\ref{theo:high}}

\begin{proof}
Using Theorem~\ref{theo:low}, we have that for each subgoal $g_{kt},\,t=0,1,\cdots,T-1$, there exists a subgoal $\tilde{g}_{kt}\in\mathcal{G}_A(s_{kt},k)$ that can induce the same low-level $k$-step action sequence as $g_{kt}$. This indicates that the agent's trajectory and the high-level reward $r_{kt}^h$ defined by Equation~\eqref{equ:r_high} remain the same for all $t$ when replacing $g_{kt}$ with $\tilde{g}_{kt}$. Then, using the high-level Bellman optimality equation for the optimal $Q$ function
\begin{equation}
\begin{aligned}
Q^*(s_{kt},g_{kt}) &= r_{kt}^h + \gamma\max_{g\in\mathcal{G}} Q^*(s_{k(t+1)},g)\\
&= r_{kt}^h + \gamma Q^*(s_{k(t+1)},g_{k(t+1)}),\quad t = 0,\,1\cdots,\,T-1
\end{aligned}
\end{equation}
and $Q^*(s_{kT},g) = 0,\ \forall\, g\in\mathcal{G}$ as $s_{kT}$ is the final state of $\tau^*$, we have $Q^*(s_{kt},\tilde{g}_{kt}) = Q^*(s_{kt}, g_{kt}),\,t=0,1,\cdots,T-1$.
\end{proof}

\section{Implementation Details}

\subsection{Adjacency Learning}
\paragraph{Constructing and updating the adjacency matrix.}
We use the agent's trajectories to construct and update the adjacency matrix. Concretely, the adjacency matrix is initialized to an empty matrix at the beginning of training. Each time when the agent explores a new state that it has never visited before, the adjacency matrix is augmented by a new row and a new column with zero elements, representing the $k$-step adjacent relation between the new state and explored states. When the temporal distance between two states in one trajectory is not larger than $k$, then the corresponding element in the adjacency matrix will be labeled to 1, indicating the adjacency. (The diagnoal of the adjacency matrix will always be labeled to 1.) Although the temporal distance between two states based on a single trajectory is often larger than the real shortest transition distance, it can be easily shown that the adjacency matrix with this labeling strategy can converge to the optimal adjacency matrix asymptotically with sufficient trajectories sampled by different policies. In practice, we employ a trajectory buffer to store newly-sampled trajectories, and update the adjacency matrix online in a fixed frequency using the stored trajectories. The trajectory buffer is cleared after each update.
\paragraph{Training the adjacency network.} The adjacency network is trained by minimizing the objective defined by Equation~\eqref{equ:contrastive}. We use states evenly-sampled from the adjacency matrix (i.e., from the set of all explored states) to approximate the expectation, and train the adjacency network each time after the adjacency matrix is updated with new trajectories. Note that by explicitly aggregating the adjacency information using an adjacency matrix, we are able to achieve the uniform sampling of all explored states and thus achieve a nearly unbiased estimation of the expectation, which cannot be realized when we directly sample state-pairs from the trajectories (see the following comparison with the work of Savinov et al.~\cite{savinov_semi-parametric_2018,savinov_episodic_2019} for details).

Embedding all subgoals with a single adjacency network is enough to express adjacency when the environment is reversible. However, when this condition is not satisfied, it is insufficient to express directional adjacency using one adjacency network, as the parameterized approximation defined by Equation~\eqref{equ:adjacency_net} is symmetric for $s_1$ and $s_2$. In this case, one can use two separate sub-networks to embed $g_1$ and $g_2$ in Equation~\eqref{equ:adjacency_net} respectively using the structure proposed in UVFA~\cite{schaul_universal_2015}.

\begin{figure}
\centering
\subfigure[]{
  \includegraphics[width=0.2\linewidth]{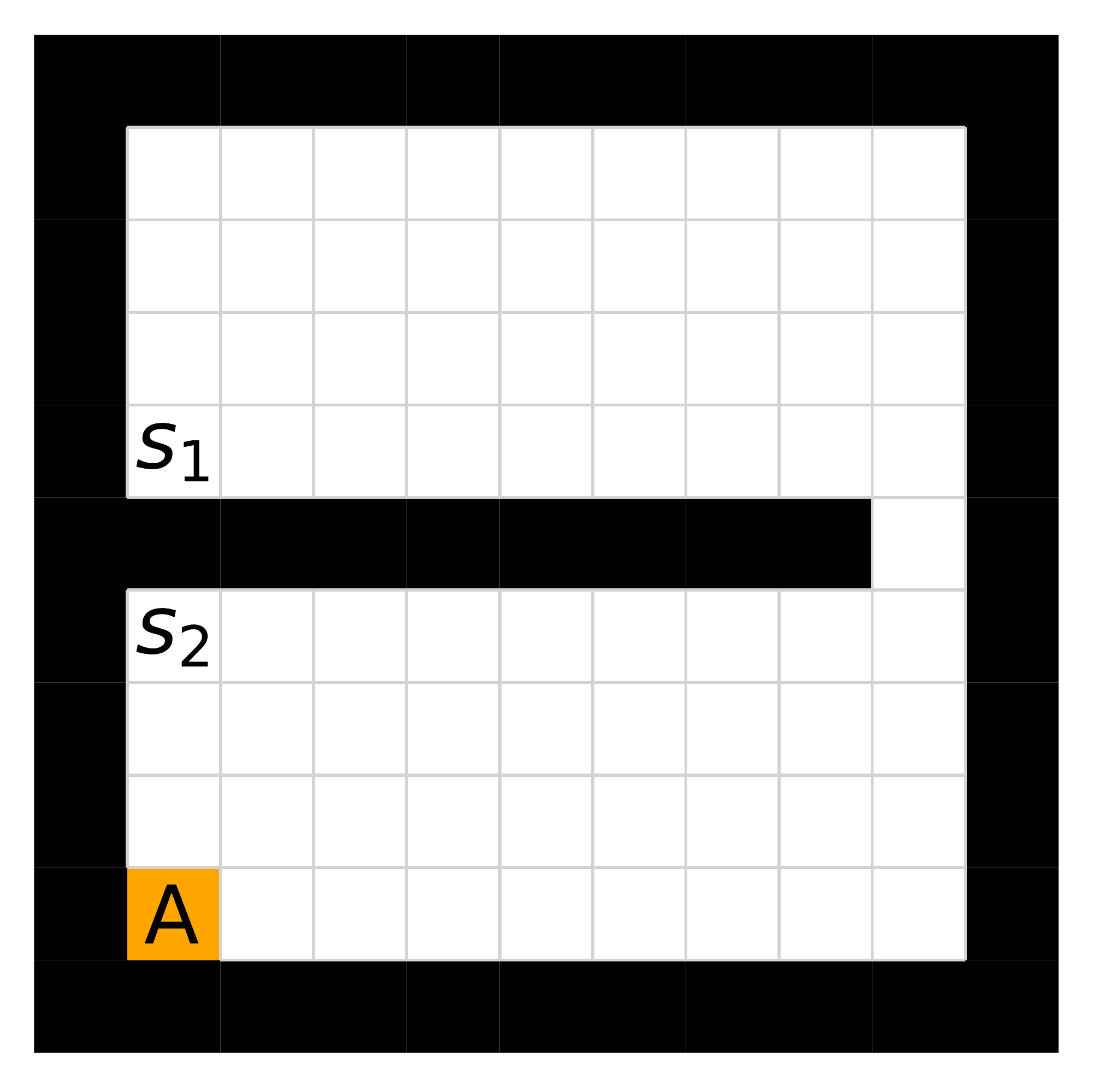}
  \label{subfig:gridworld}
}
\subfigure[]{
  \includegraphics[width=0.19\linewidth]{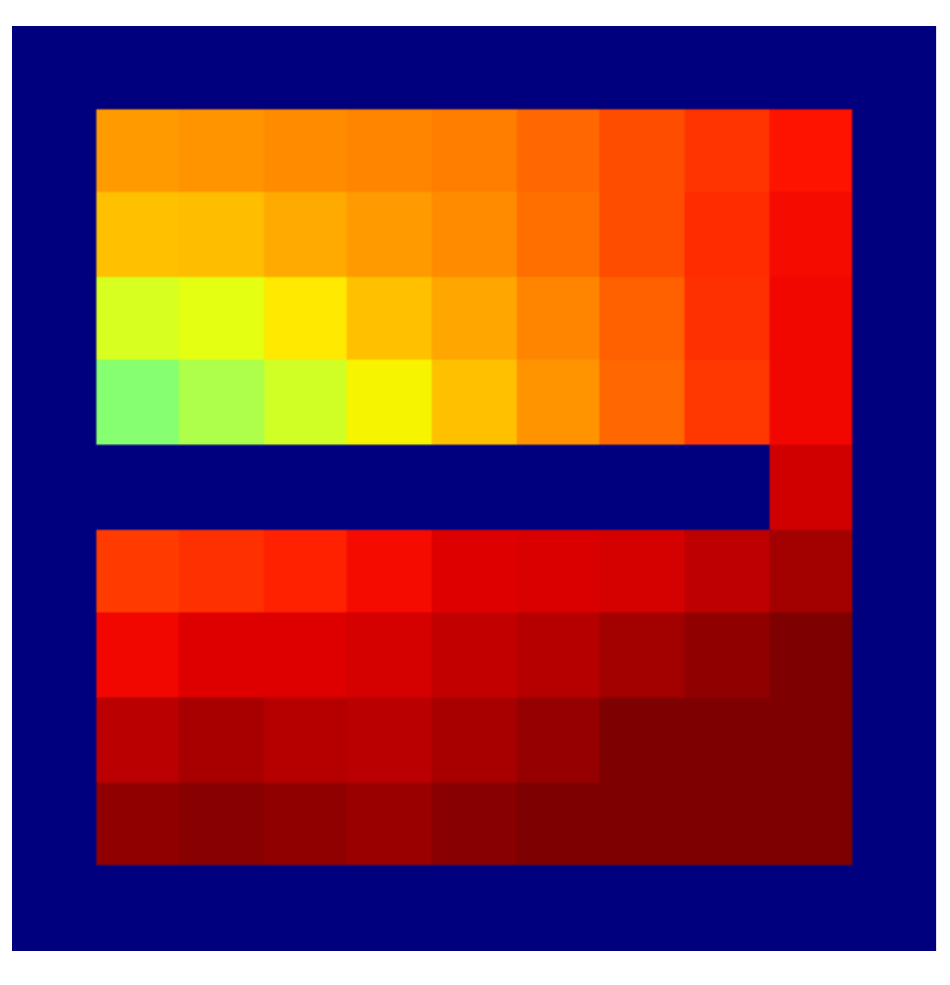}
  \hspace{0.8em}
  \includegraphics[width=0.19\linewidth]{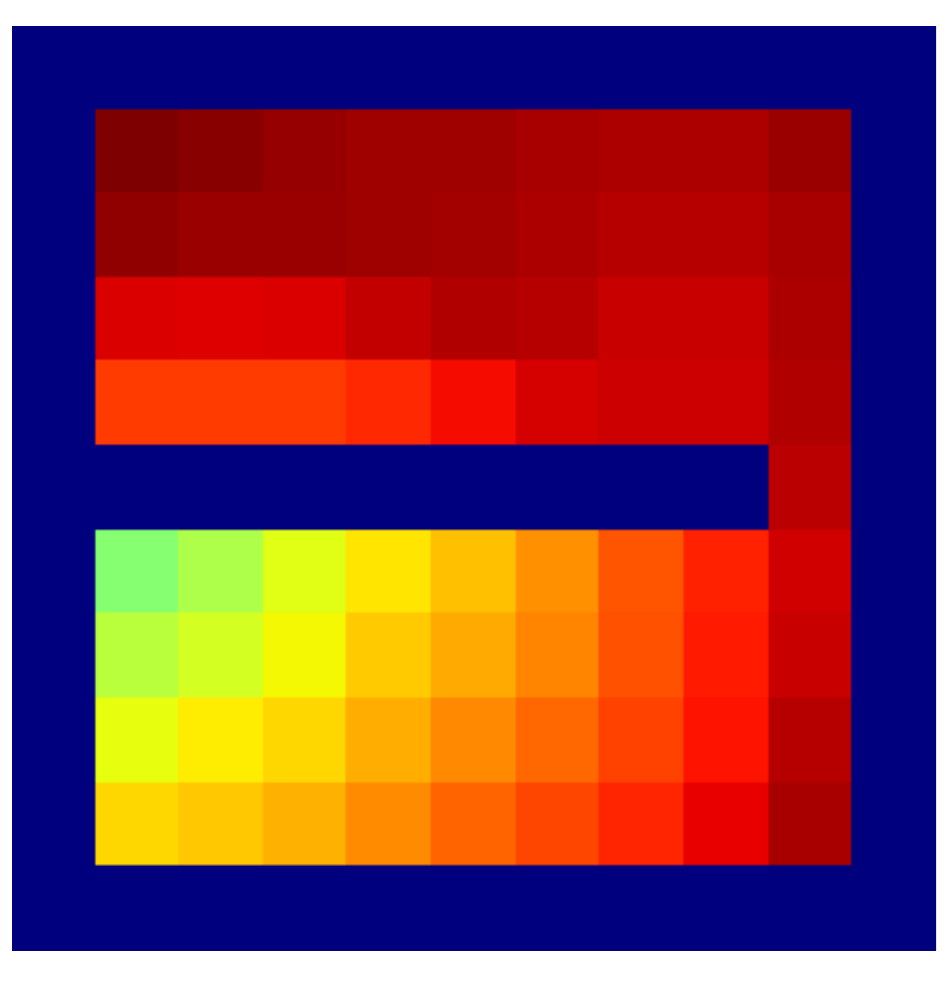}
  \hspace{0.8em}
  \includegraphics[width=0.19\linewidth]{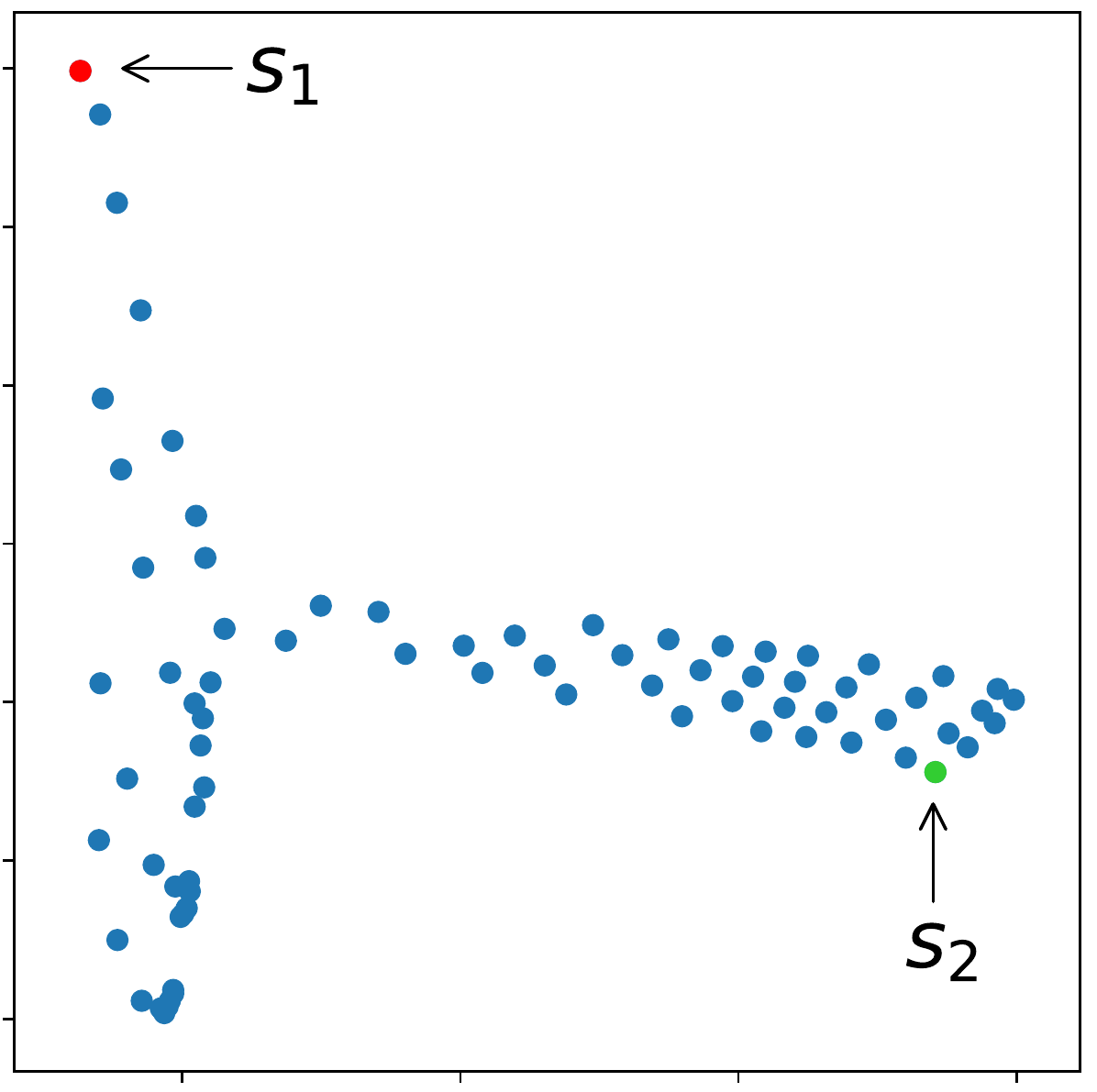}
  \label{subfig:lle_ours}
  }\\
\subfigure[]{
  \hspace{0.1em}
  \includegraphics[width=0.19\linewidth]{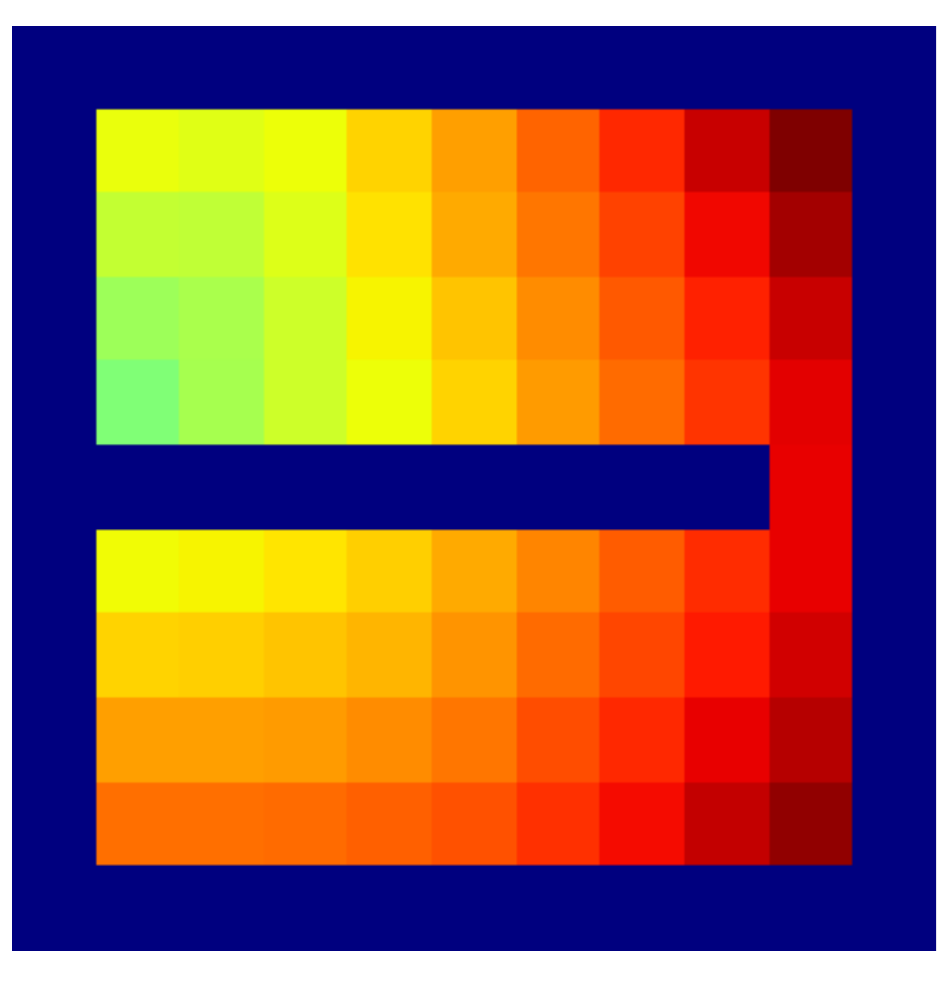}
  \hspace{0.8em}
  \includegraphics[width=0.19\linewidth]{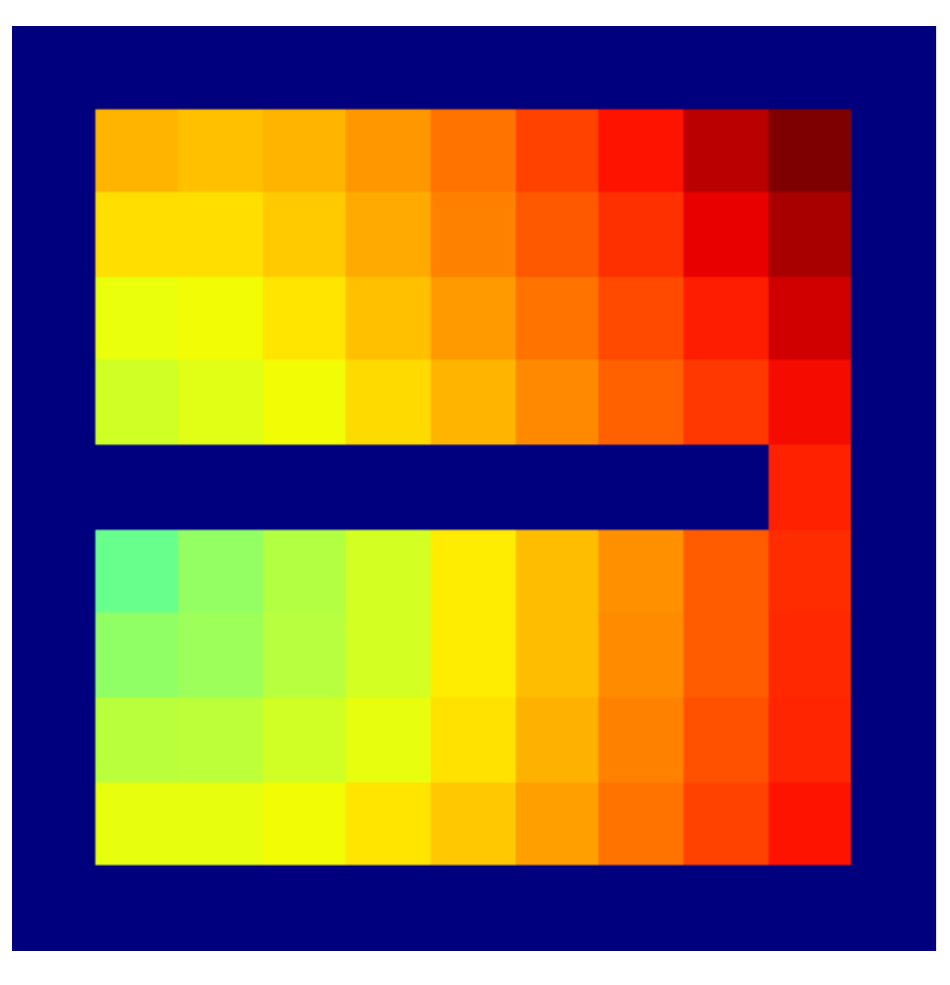}
  \hspace{0.8em}
  \includegraphics[width=0.19\linewidth]{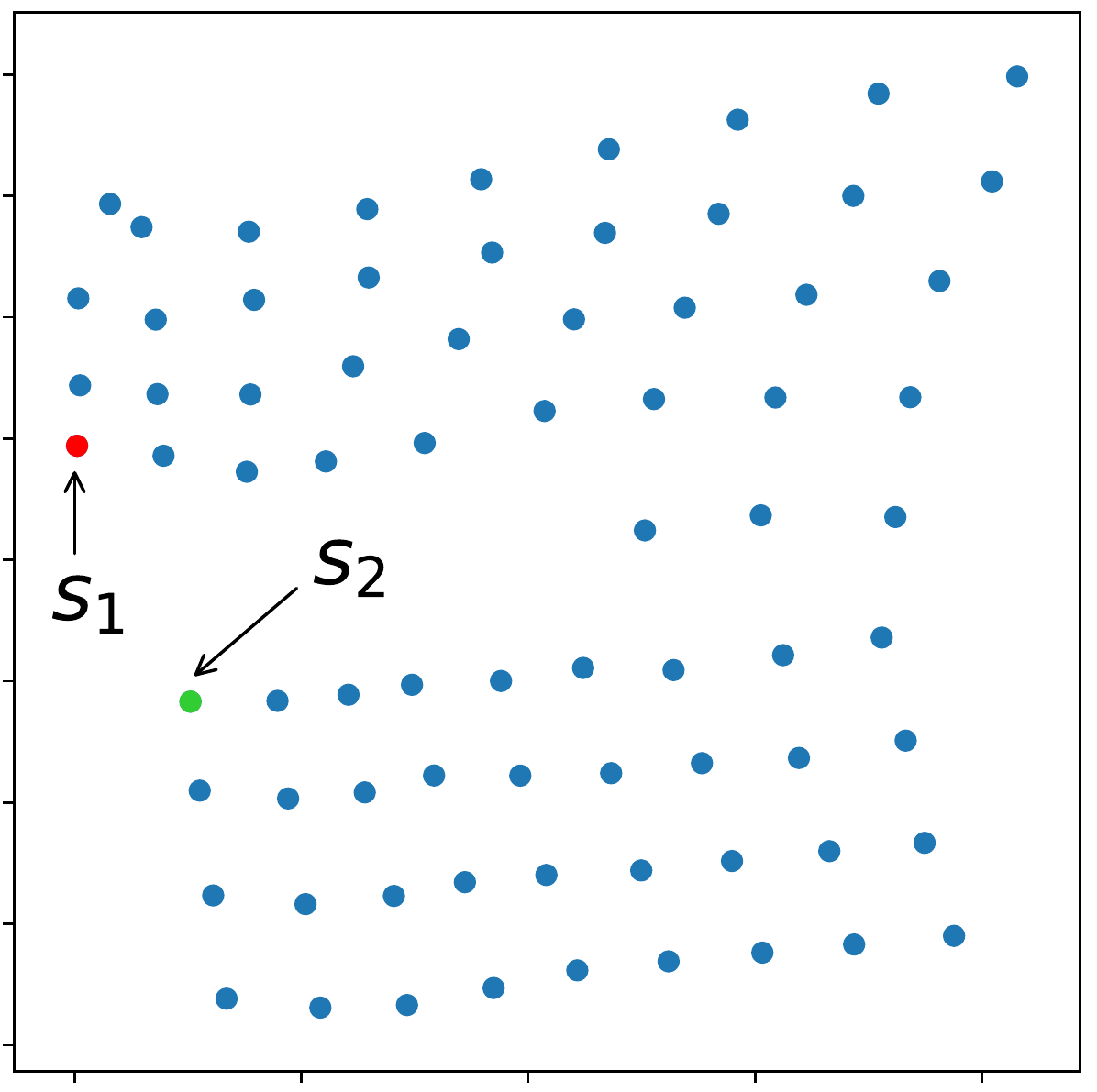}
  \label{subfig:lle_savinov}
  }
\caption{Qualitative comparison of adjacency learning methods. (a) Environment layout. The agent starts from the grid A. (b) Results of our method, including the adjacency heatmaps from states $s_1$, $s_2$ and the LLE visualization of state embeddings. (c) Results of the method proposed by Savinov et al.~\cite{savinov_semi-parametric_2018,savinov_episodic_2019}, including the adjacency heatmaps from states $s_1$, $s_2$ and the LLE visualization of state embeddings.}
\label{fig:r_compare}
\end{figure}

\paragraph{Comparison with the work of Savinov et al.} Savinov et al.~\cite{savinov_semi-parametric_2018,savinov_episodic_2019} also propose a supervised learning approach for learning the adjacency between states. The main differences between our method and theirs are: 1) We use trajectories sampled by multiple policies to construct training samples, while they only use trajectories sampled by one specific policy; 2) We use an adjacency matrix to explicitly aggregate the adjacency information and sample training pairs based on the adjacency matrix, while they directly sample training pairs from trajectories. These differences lead to two advantages of our method: 1) By using multiple policies, we achieve a more accurate adjacency approximation, as shown by Equation~\eqref{equ:approximation}; 2) By maintaining an adjacency matrix, we can uniformly sample from the set of all explored states and realize a nearly unbiased estimation of the expectation in Equation~\eqref{equ:contrastive}, while the estimation by sampling state-pairs from trajectories is biased. As an example, consider a simple grid world in Figure~\ref{subfig:gridworld}, where states are represented by their $(x,y)$ positions. In this environment, states $s_1$ and $s_2$ are non-adjacent since they are separated by a wall. However, it is hard for the method by Savinov et al. to handle this situation as these two states rarely emerge in the same trajectory due to the large distance, and thus the loss induced by this state-pair is very likely to be dominated by the loss of other nearer state-pairs. Meanwhile, our method treat the loss of all state-pairs equally, and can therefore alleviate this phenomenon.
Empirically, we employed a random agent (since the random policy is stochastic, it can be viewed as multiple deterministic policies, and is enough for adjacency learning in this simple environment) to interact with the environment for $20,000$ steps, and trained the adjacency network with collected samples using both methods.
We visualize the LLE of state embeddings and two adjacency distance heatmaps by both methods respectively in Figure~\ref{subfig:lle_ours} and~\ref{subfig:lle_savinov}. Visualizations validate our analysis, showing that our method does learn a better adjacency measure in this scenario.

\subsection{Algorithm Pseudocode}

We provide Algorithm~\ref{algo:hrac} to show the training procedure of HRAC. Some training details are omitted for brevity, e.g., the concrete training flow of the low-level policy.
\begin{algorithm}[tb]
   \caption{HRAC}
   \label{algo:hrac}
\begin{algorithmic}
   \STATE {\bfseries Input:} High-level policy $\pi^h_{\theta_h}$ parameterized by $\theta_h$, low-level policy $\pi^l_{\theta_l}$ parameterized by $\theta_l$, adjacency network $\psi_\phi$ parameterized by $\phi$, state-goal mapping function $\varphi$, goal transition function $h$, high-level action frequency $k$, number of training episodes $N$, adjacency learning frequency $C$, empty adjacency matrix $\mathcal{M}$, empty trajectory buffer $\mathcal{B}$.
   \vspace{0.5em}
   \STATE Sample and store trajectories in the trajectory buffer $\mathcal{B}$ using a random policy.
   \STATE Construct the adjacency matrix $\mathcal{M}$ using the trajectory buffer $\mathcal{B}$.
   \STATE Pre-train $\psi_\phi$ using $\mathcal{M}$ by minimizing Equation~\eqref{equ:contrastive}.
   \STATE Clear $\mathcal{B}$.
   \FOR {$n=1$ {\bfseries to} $N$}
       \STATE Reset the environment and sample the initial state $s_0$.
       \STATE $t = 0$.
       \REPEAT
       \IF {$t\equiv 0\,(\mathrm{mod}\ k)$}
         \STATE Sample subgoal $g_t \sim \pi^h_{\theta_h}(g|s_t)$.
       \ELSE
           \STATE Perform subgoal transition $g_t = h(g_{t-1},\,s_{t-1},\,s_t)$.
       \ENDIF
       \STATE Sample low-level action $a_t \sim \pi^l_{\theta_l}(a|s_t,\,g_t)$.
       \STATE Sample next state $s_{t+1} \sim \mathcal{P}(s|s_t,\,a_t)$.
       \STATE Sample reward $r_t \sim \mathcal{R}(r|s_t,\,a_t)$.
       \STATE Sample episode end signal $done$.
       \STATE $t = t+1$.
     \UNTIL {$done$ is $true$.}
     \STATE Store the sampled trajectory in $\mathcal{B}$.
     \STATE Train high-level policy $\pi^h_{\theta_h}$ according to Equation~\eqref{equ:total_loss} and~\eqref{equ:goal_loss}.
     \STATE Train low-level policy $\pi^l_{\theta_l}$.
     \IF {$n\equiv 0\,(\mathrm{mod}\ C)$}
         \STATE Update the adjacency matrix $\mathcal{M}$ using the trajectory buffer $\mathcal{B}$.
         \STATE Fine-tune $\psi_\phi$ using $\mathcal{M}$ by minimizing Equation~\eqref{equ:contrastive}.
         \STATE Clear $\mathcal{B}$.
     \ENDIF
   \ENDFOR
\end{algorithmic}
\end{algorithm}



\subsection{Environment Details}

\paragraph{Maze.} This environment has a size of $13\times17$, with a discrete 2-dimensional state space representing the $(x,y)$ position of the agent and a discrete 4-dimensional action space corresponding to actions moving towards four directions. The agent is provided with a dense reward to facilitate exploration, i.e., $+0.1$ each step if the agent moves closer to the goal, and $-0.1$ each step if the agent moves farther. Each episode has a maximum length of 200. Environmental stochasticity is introduced by replacing the action of the agent by a random action each step with a probability of 0.25.

\paragraph{Key-Chest.} This environment has a size of $13\times17$, with a discrete 3-dimensional state space in which the first two dimensions represent the $(x,y)$ position of the agent respectively, and the third dimension represents whether the agent has picked up the key (1 if the agent has the key and 0 otherwise). The agent has the same action space as the Maze task. The agent is provided with sparse reward of $+1$ and $+5$, respectively for picking up the key and opening the chest. Each episode ends if the agent opens the chest or runs out of the step limit of 500. The random action probability of the environment is also 0.25.

\paragraph{Ant Gather.} This environment has a size of $20\times 20$, with a continuous state space including the current position and velocity, the current time step $t$, and the depth readings defined by the stardard Gather environment~\cite{duan_benchmarking_2016}. We use the ant robot pre-defined by Rllab, with a 8-dimensional continuous action space. The ant robot is spawned at the center of the map and needs to gather apples while avoiding bombs. Both apples and bombs are randomly placed in the environment at the beginning of each episode. The agent receives a positive reward of $+1$ for each apple and a negative reward of $-1$ for each bomb. Each episode terminates at 500 time steps.

\paragraph{Ant Maze.} This environment has a size of $24\times 24$, with a continuous state space including the current position and velocity, the target location, and the current time step $t$. In the training stage, the environment randomly samples a target position at the beginning of each episode, and the agent receives a dense reward at each time step according to its negative Euclidean distance from the target position. At evaluation stage, the target position is fixed to $(0,16)$, and the success is defined as being within an Euclidean distance of 5 from the target. Each episode ends at 500 time steps. In practice, we scale the environmental reward by 0.1 equally for all methods.

\paragraph{Ant Maze Sparse.} This environment has a size of $20\times20$, with the same state and action spaces as the Ant Maze task. The target position (goal) is set at the position $(2.0,\,9.0)$ in the center corridor. The agent is rewarded by $+1$ only if it reachs the goal, which is defined as having a Euclidean distance that is smaller than 1 from the goal. At the beginning of each episode, the agent is randomly placed in the maze except at the goal position. Each episode is terminated if the agent reaches the goal or after 500 steps.

\subsection{HRAC and Baseline Details}

We use PyTorch to implement our method HRAC and all the baselines.\footnote{We use the open source PyTorch implementation of HIRO at~\url{https://github.com/bhairavmehta95/data-efficient-hrl}.}

\paragraph{HRAC.} For discrete control tasks, we adopt a binary intrinsic reward setting: we set the intrinsic reward to 1 when $|s_x - g_x| \le 0.5$ and $|s_y - g_y| \le 0.5$, where $(s_x,\,s_y)$ is the position of the agent and $(g_x,\,g_y)$ is the position of the desired subgoal. For continuous control tasks, we adopt a dense intrinsic reward setting based on the negative Euclidean distances $-\lVert s - g\rVert_2$ between states and subgoals.

\paragraph{HIRO.} Following Nachum et al.~\cite{nachum_data-efficient_2018}, we restrict the output of high-level to $(\pm10,\,\pm10)$, representing the desired shift of the agent's $(x,y)$ position. By limiting the range of directional subgoals generated by the high-level, HIRO can roughly control the Euclidean distance between the absolute subgoal and the current state in the raw goal space rather than the learned adjacency space. 

\paragraph{HRL-HER.} As HER cannot be applied to the on-policy training scheme in a straightforward manner, in discrete control tasks where the low level policy is trained using A2C, we modify its implementation so that it can be incorporated into the on-policy setting. For this on-policy variant, during the training phase, we maintain an additional episodic state memory. This memory stores states that the agent has visited from the beginning of each episode. When the high-level generates a new subgoal, the agent randomly samples a subgoal mapped from a stored state with a fixed probability 0.2 to substitute the generated subgoal for the low-level to reach. This implementation resembles the ``episode'' strategy introduced in the original HER. We still use the original HER in continuous control tasks.

\paragraph{NoAdj.} We follow the training pipeline proposed by Savinov et al.~\cite{savinov_semi-parametric_2018,savinov_episodic_2019}, where no adjacency matrix is maintained. Training pairs are constructed by randomly sampling state-pairs $(s_i,\,s_j)$ from the stored trajectories. The samples with $|i-j|\le k$ are labeled as positive with $l=1$, and the samples with $|i-j| \ge M k$ are negative ones with $l=0$. The hyper-parameter $M$ is used to create a gap between the two types of samples, where in practice we use $M = 4$.

\paragraph{NegReward.} In this variant, every time the high-level generates a subgoal, we use the adjacency network to judge whether it is $k$-step adjacent. If the subgoal is non-adjacent, the high-level will be penalized with a negative reward $-1$.

\subsection{Network Architecture}
For the hierarchical policy network, we employ the same architecture as HIRO~\cite{nachum_data-efficient_2018} in continuous control tasks, where both the high-level and the low-level use TD3~\cite{fujimoto_addressing_2018} algorithm for training. In discrete control tasks, we use two networks consisting of 3 fully-connected layers with ReLU nonlinearities as the low-level actor and critic networks of A2C (our preliminary results show that the performances using on-policy and off-policy methods for the low-level training are similar in the discrete control tasks we consider), and use the same high-level TD3 network architecture as the continuous control task. The size of the hidden layers of both low-level actor and critic is $(300,\,300)$. The output of high-level actor is activated using the \texttt{tanh} function and scaled to fit the size of the environments.

For the adjacency network, we use a network consisting of 4 fully-connected layers with ReLU nonlinearities in all tasks. Each hidden layer of the adjacency network has the size of $(128,\,128)$. The dimension of the output embedding is 32.

We use Adam optimizer for all networks.

\subsection{Hyper-parameters}
\label{subsec:hyperparam}

We list all hyper-parameters we use in the discrete and continuous control tasks respectively in Table~\ref{tab:param_disc} and Table~\ref{tab:param_cont}, and list the hyper-parameters used for adjacency network training in Table~\ref{tab:param_adj}. ``Ranges'' in the tables show the ranges of hyper-parameters considered, and the hyper-parameters without ranges are not tuned.

\section{Additional Visualizations}

\begin{figure}
\centering
\includegraphics[width=0.18\linewidth]{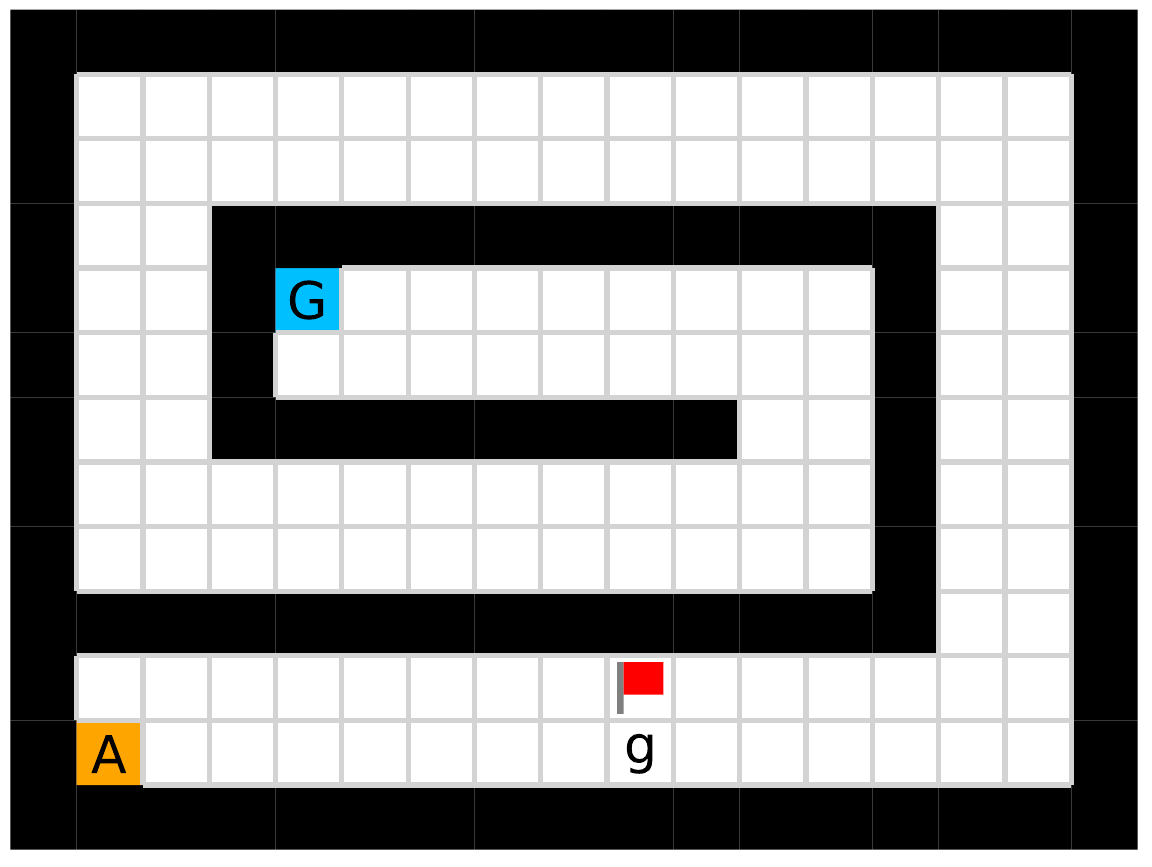}
\hspace{0.4em}
\vspace{0.5em}
\includegraphics[width=0.18\linewidth]{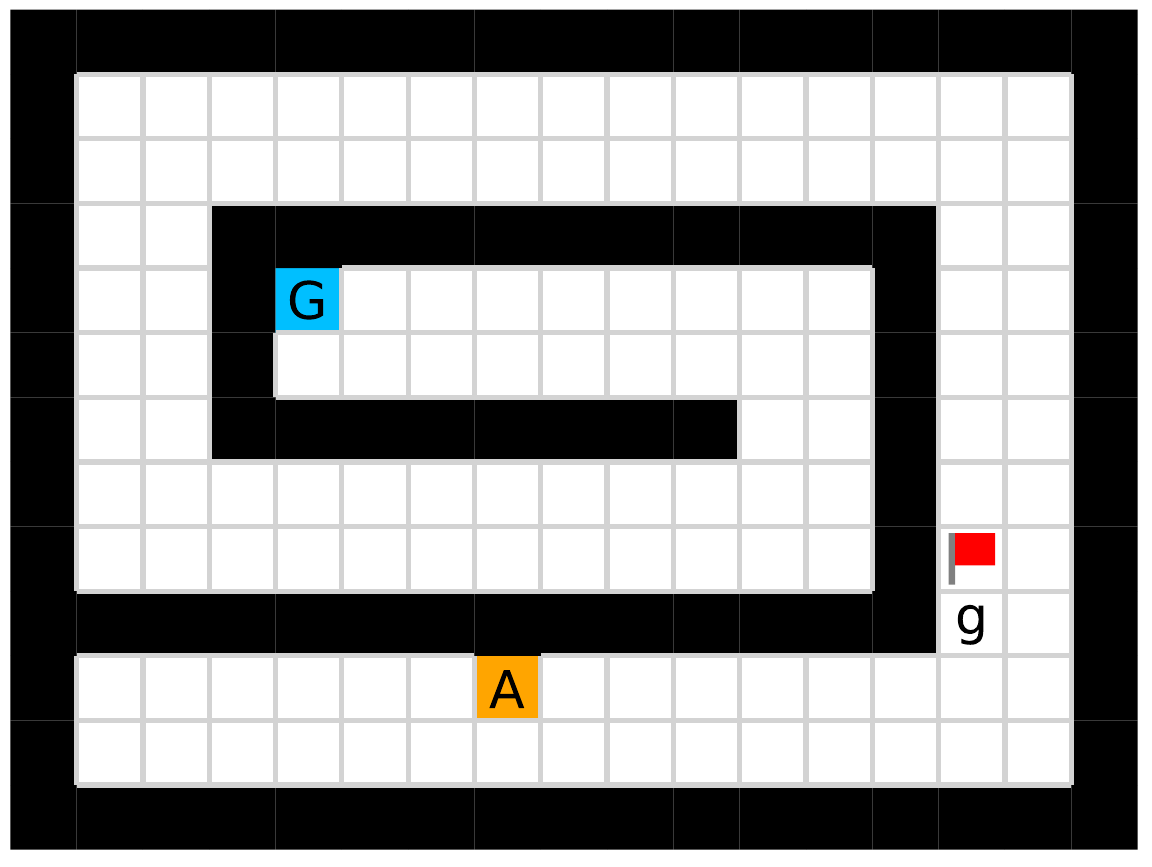}
\hspace{0.4em}
\includegraphics[width=0.18\linewidth]{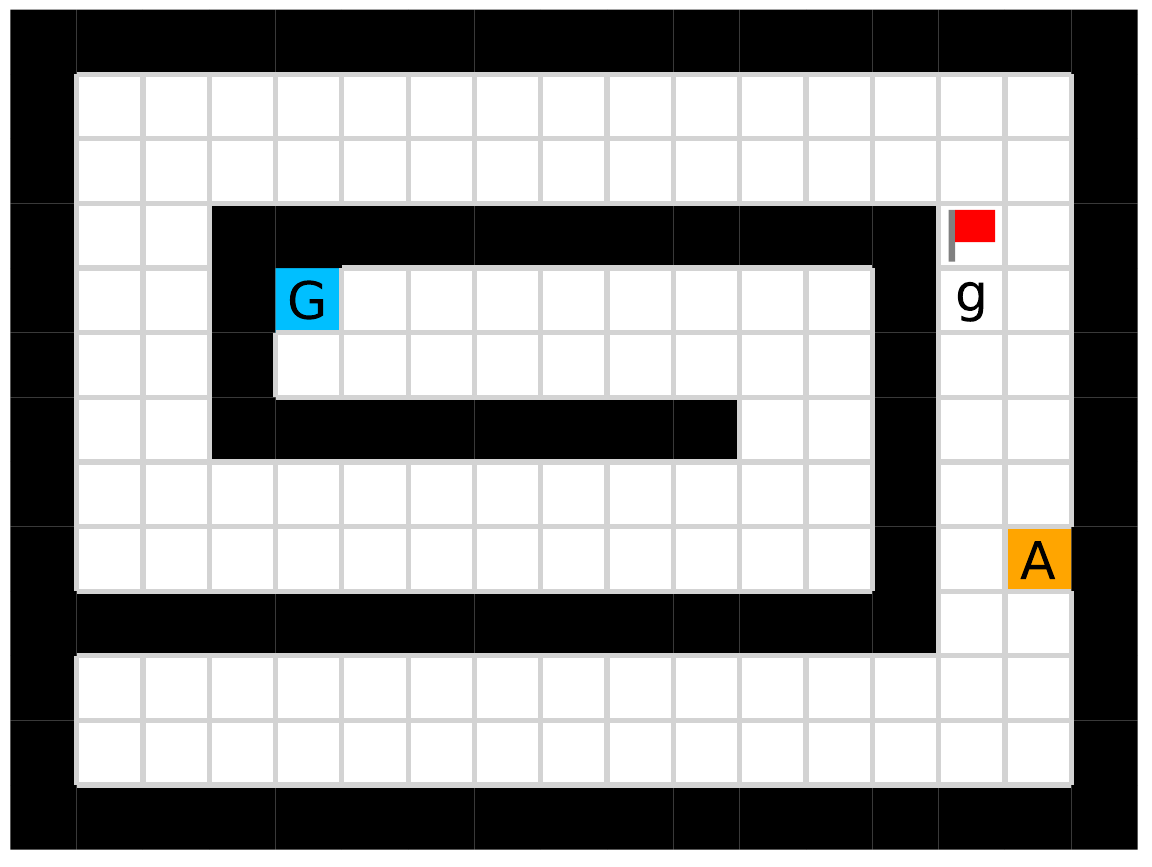}
\hspace{0.4em}
\includegraphics[width=0.18\linewidth]{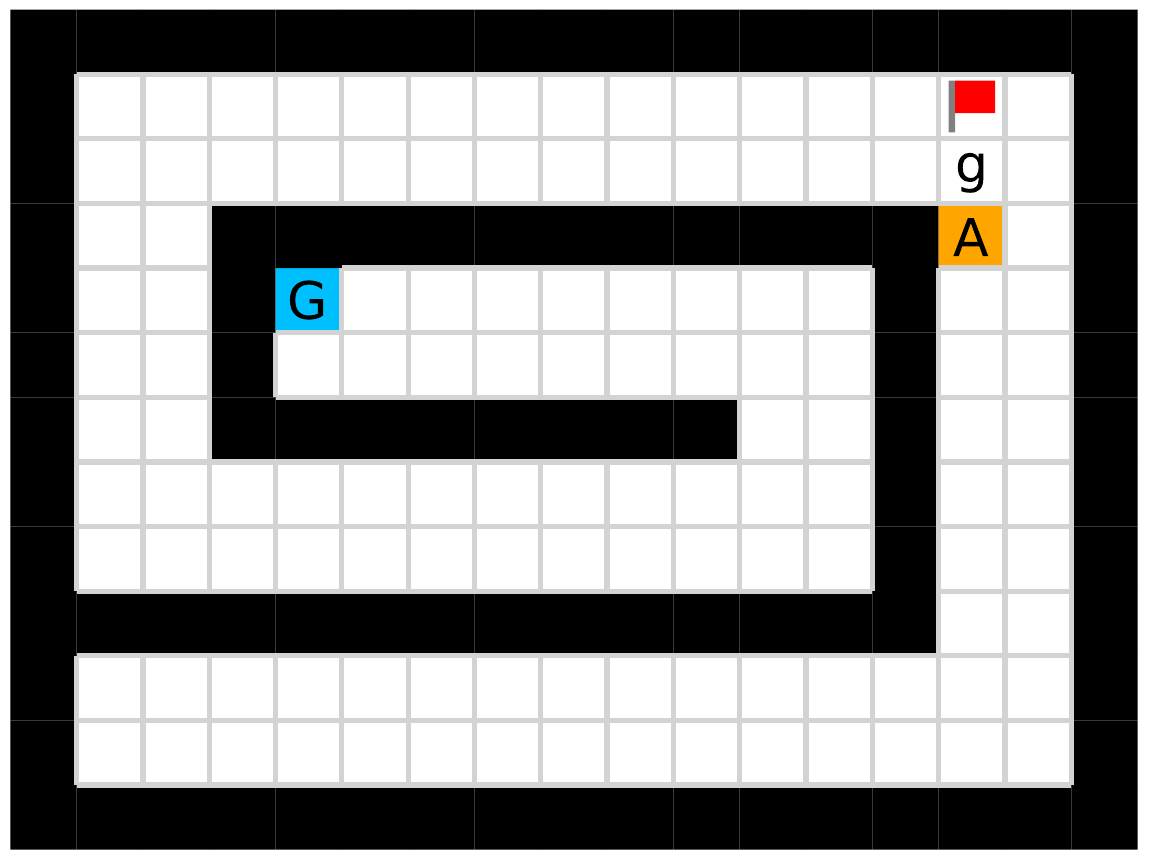}
\hspace{0.4em}
\includegraphics[width=0.18\linewidth]{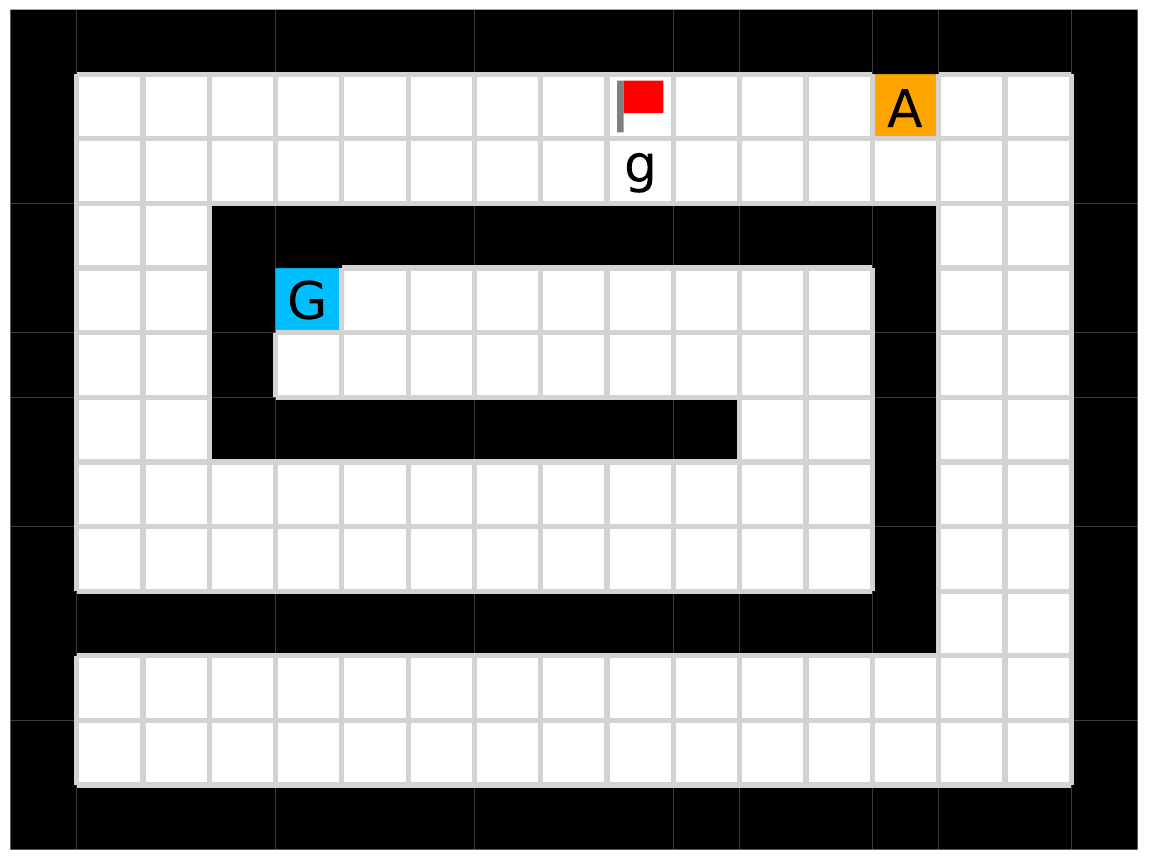}\\

\includegraphics[width=0.18\linewidth]{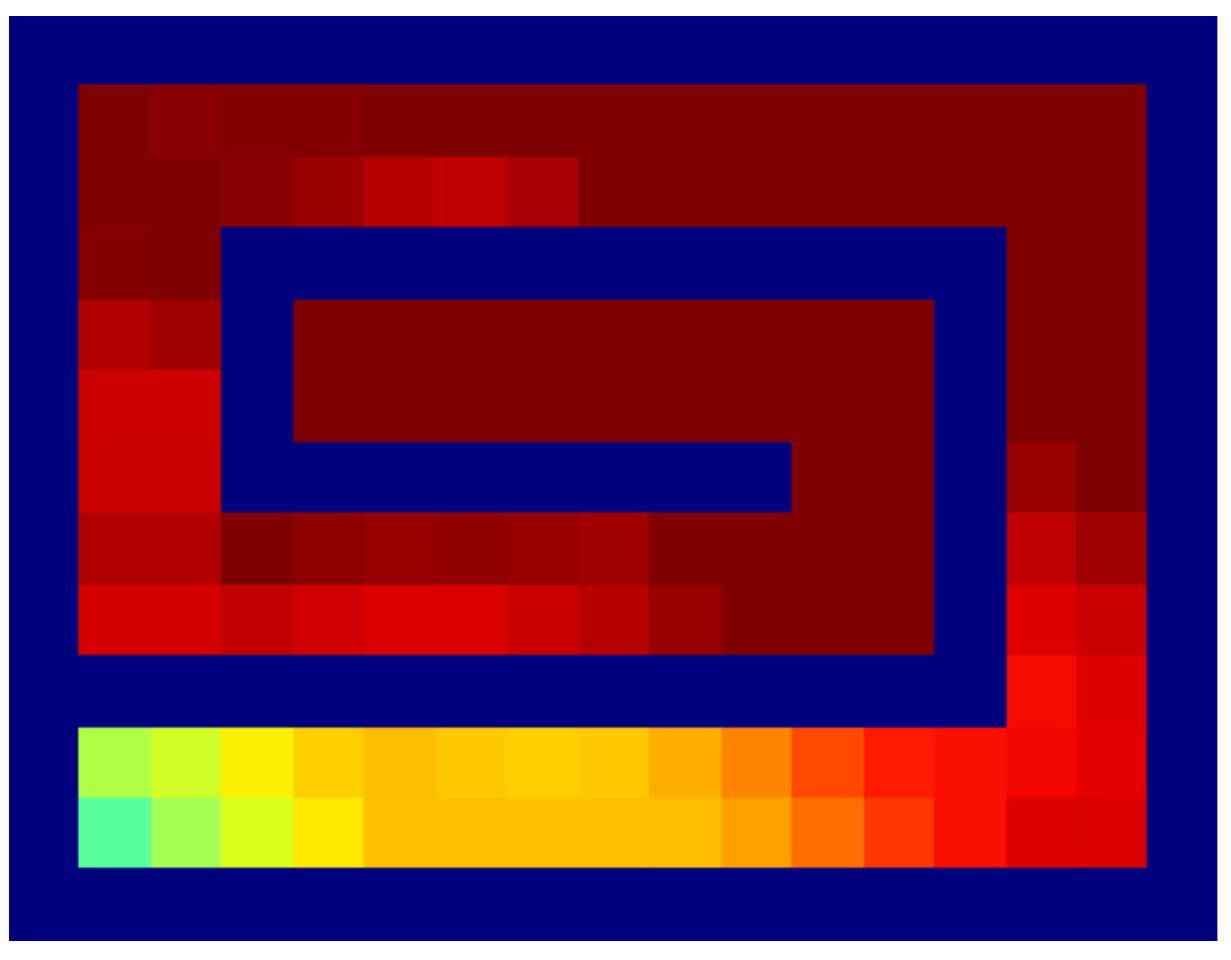}
\hspace{0.4em}
\vspace{2.5em}
\includegraphics[width=0.18\linewidth]{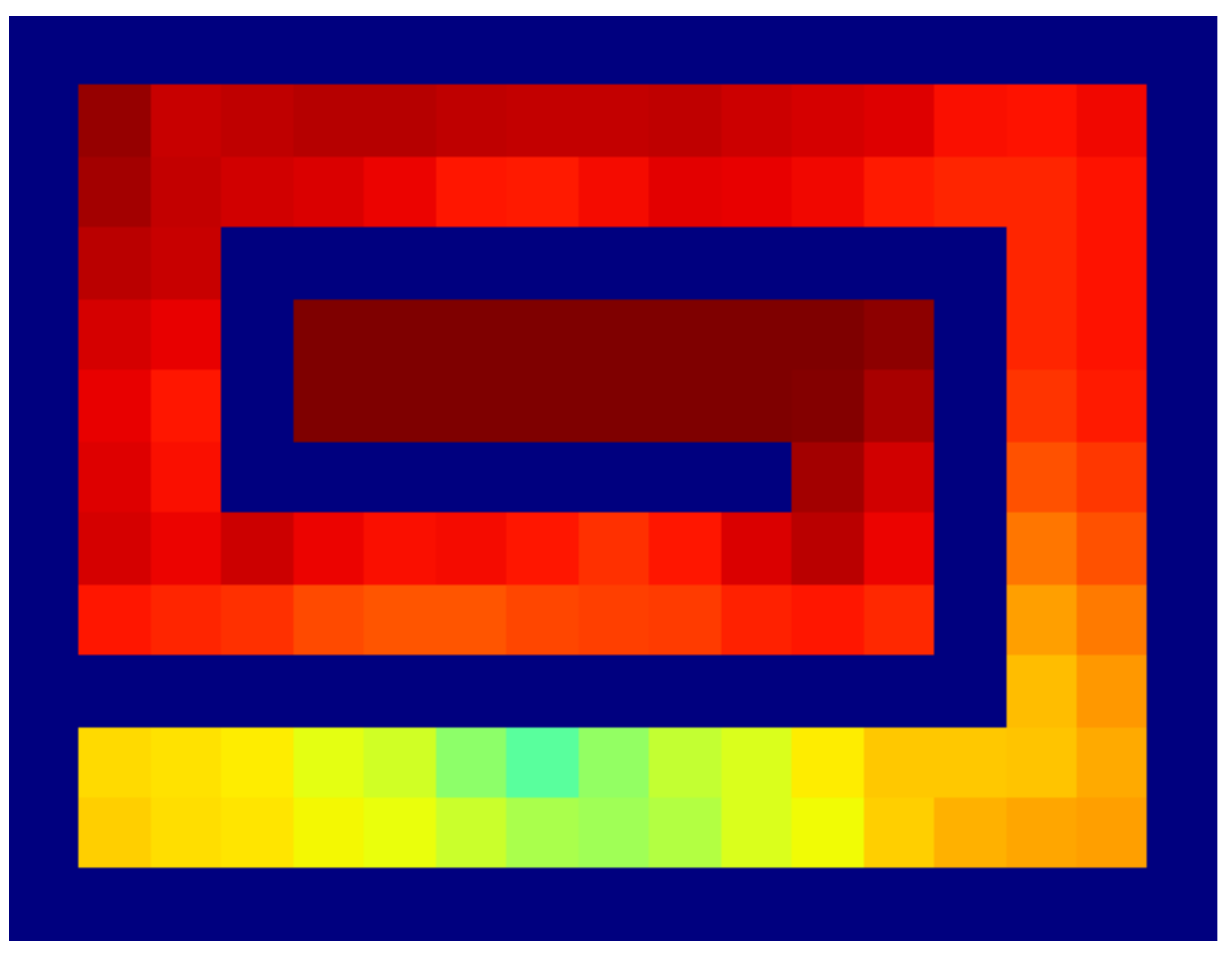}
\hspace{0.4em}
\includegraphics[width=0.18\linewidth]{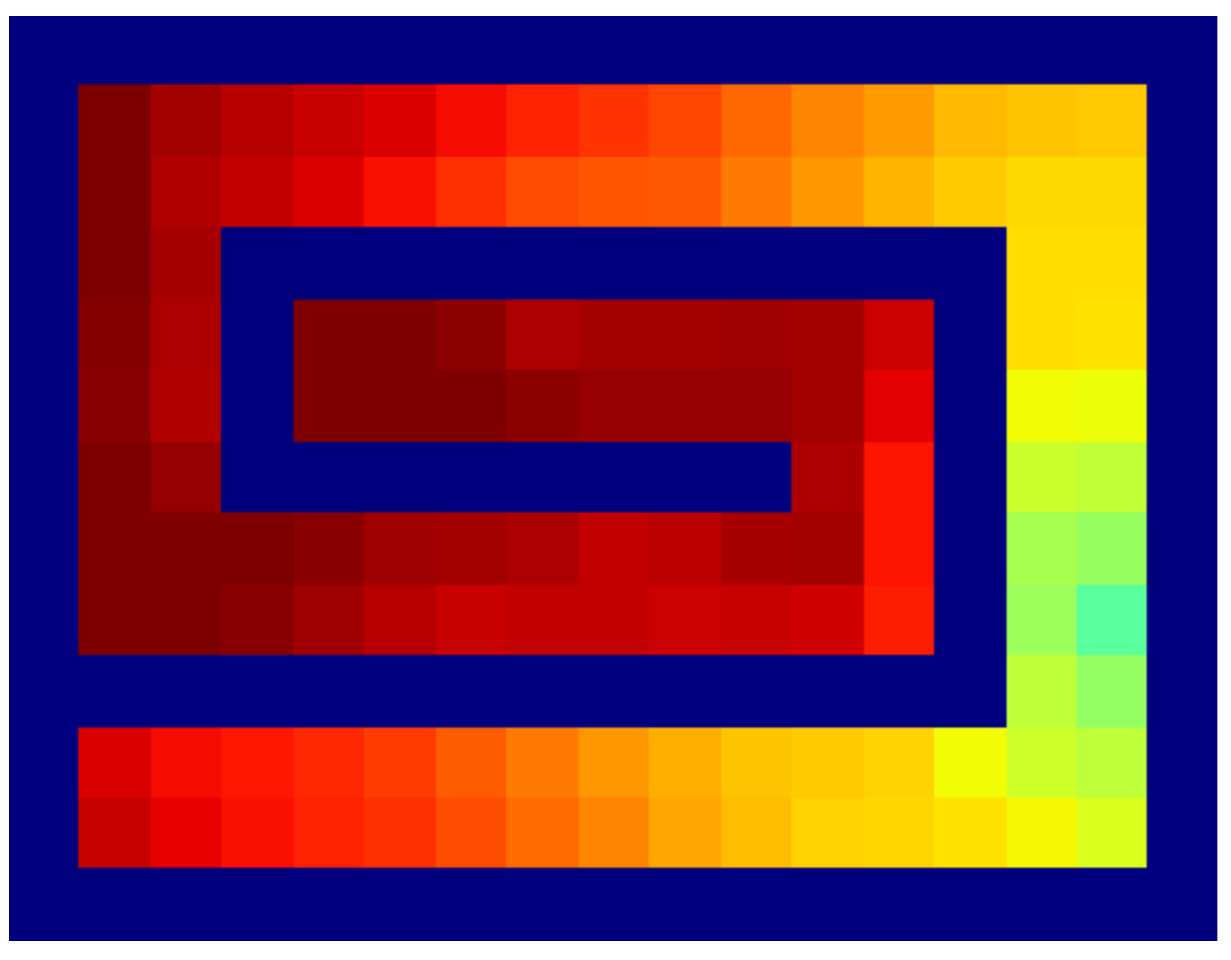}
\hspace{0.4em}
\includegraphics[width=0.18\linewidth]{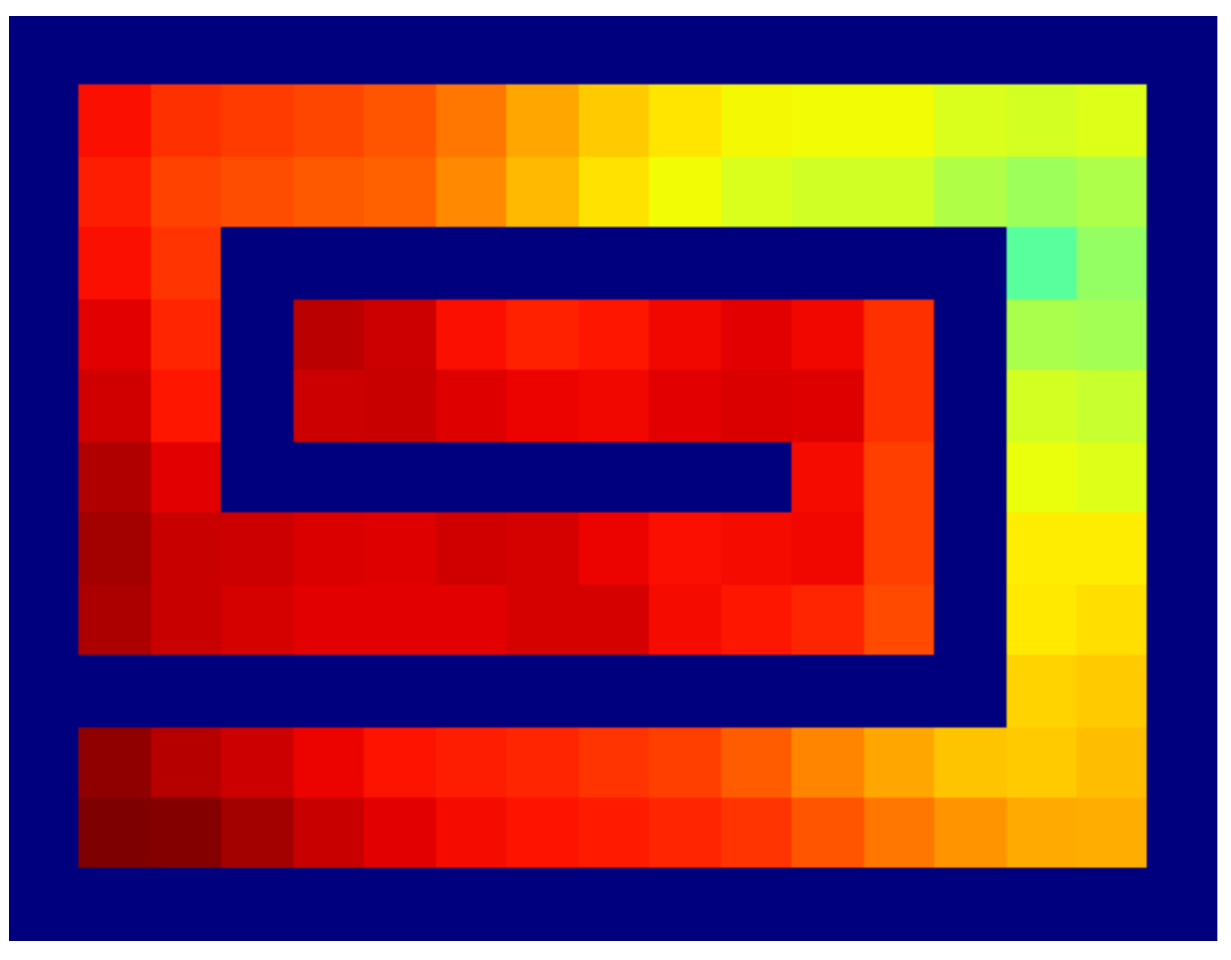}
\hspace{0.4em}
\includegraphics[width=0.18\linewidth]{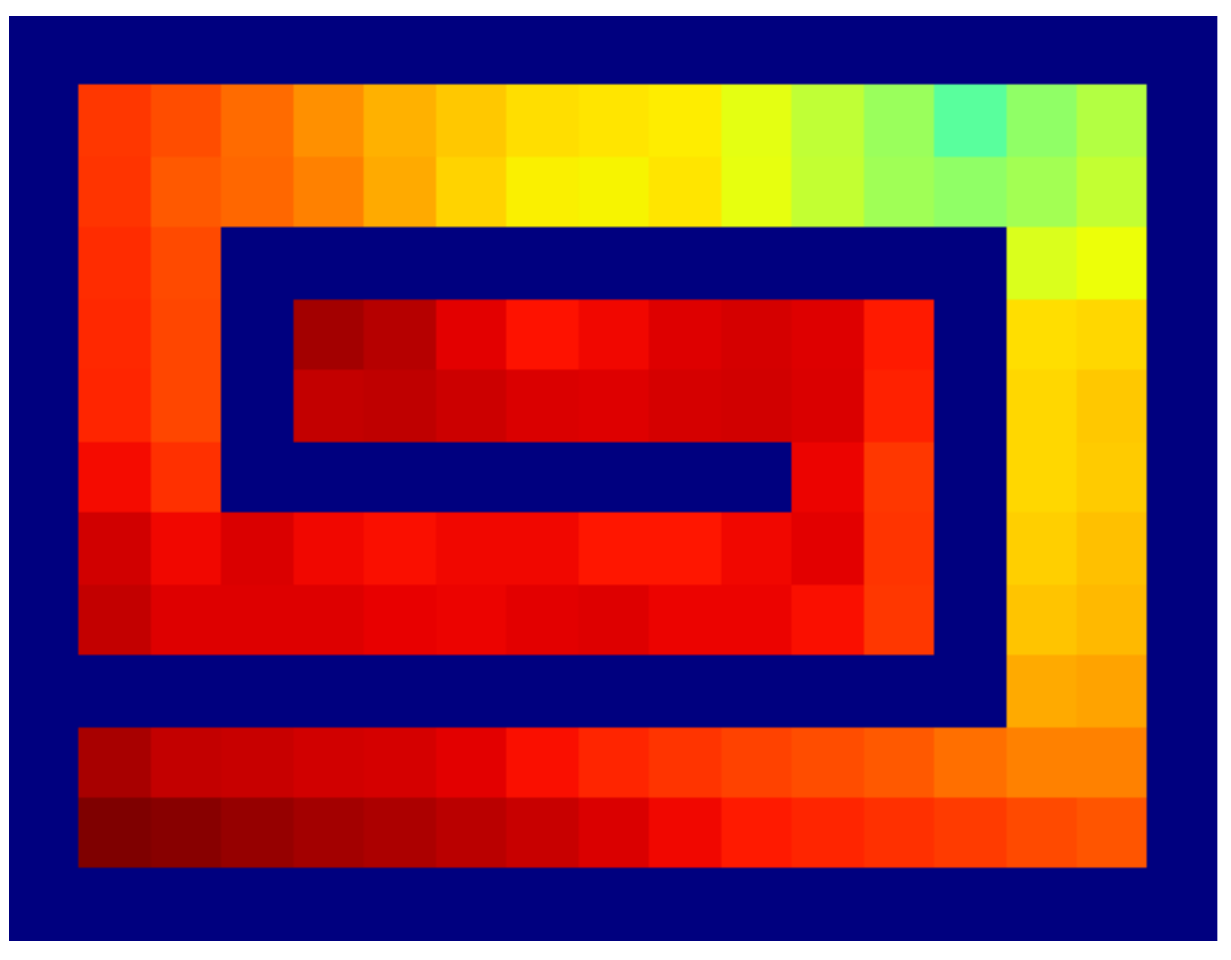}\\

\includegraphics[width=0.18\linewidth]{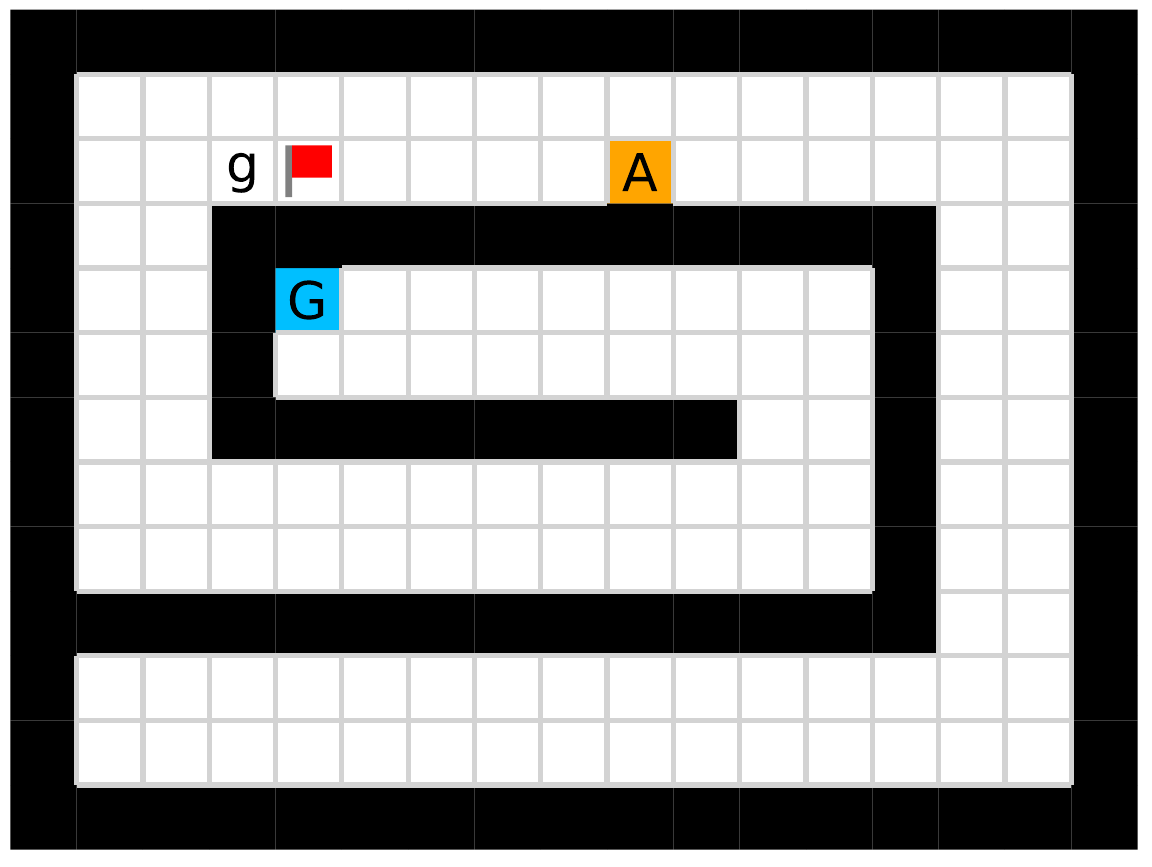}
\hspace{0.4em}
\vspace{0.5em}
\includegraphics[width=0.18\linewidth]{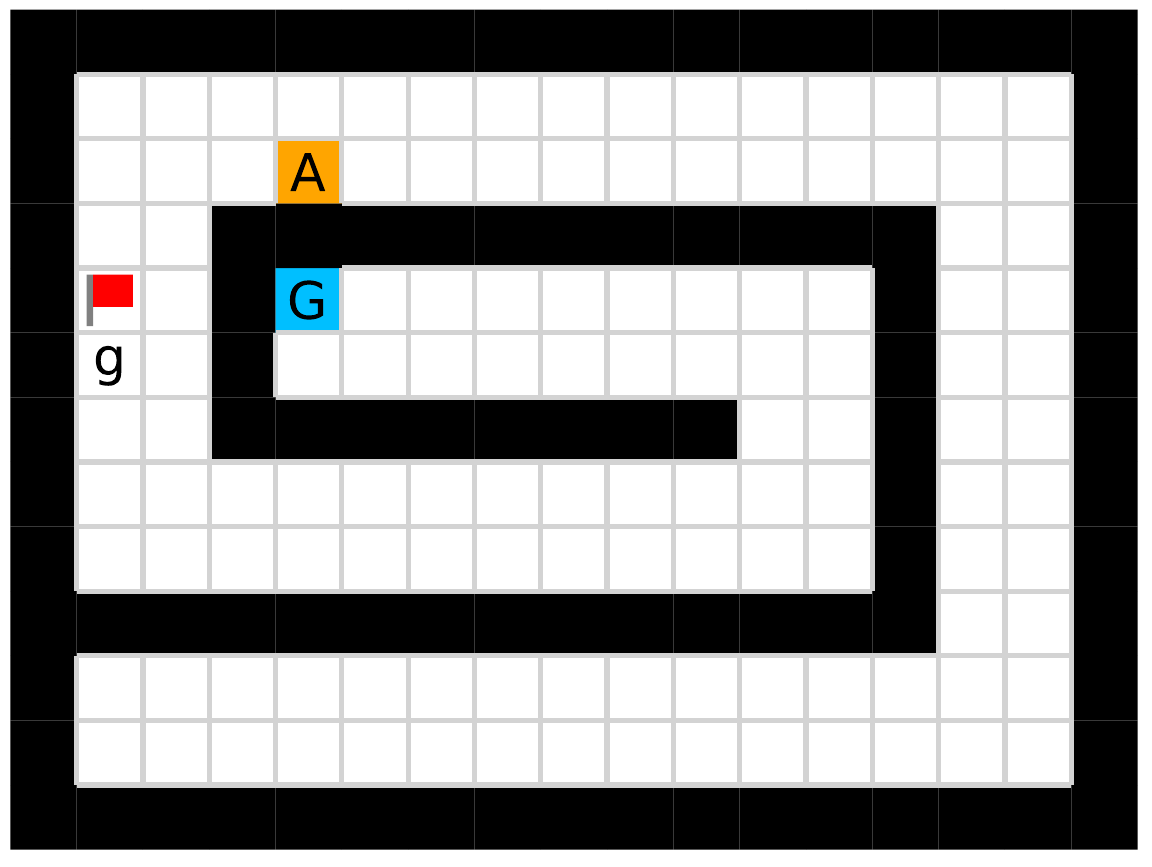}
\hspace{0.4em}
\includegraphics[width=0.18\linewidth]{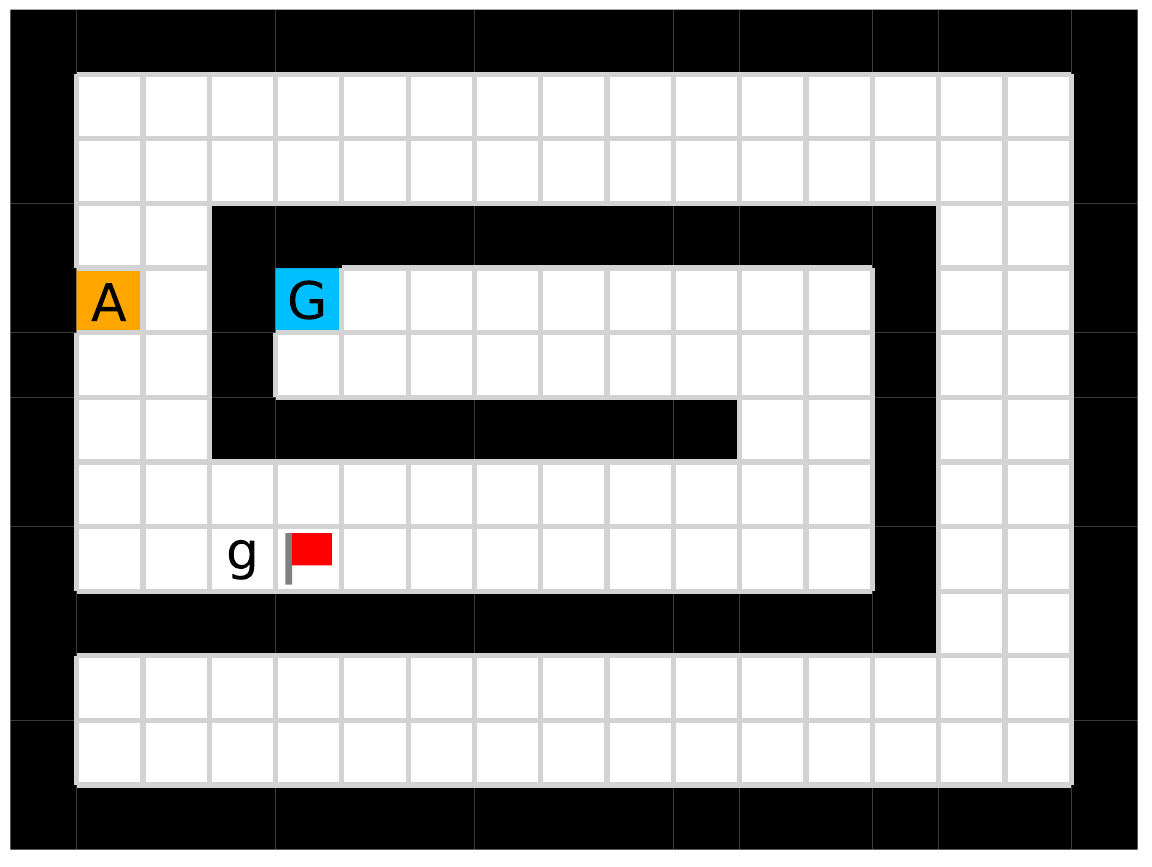}
\hspace{0.4em}
\includegraphics[width=0.18\linewidth]{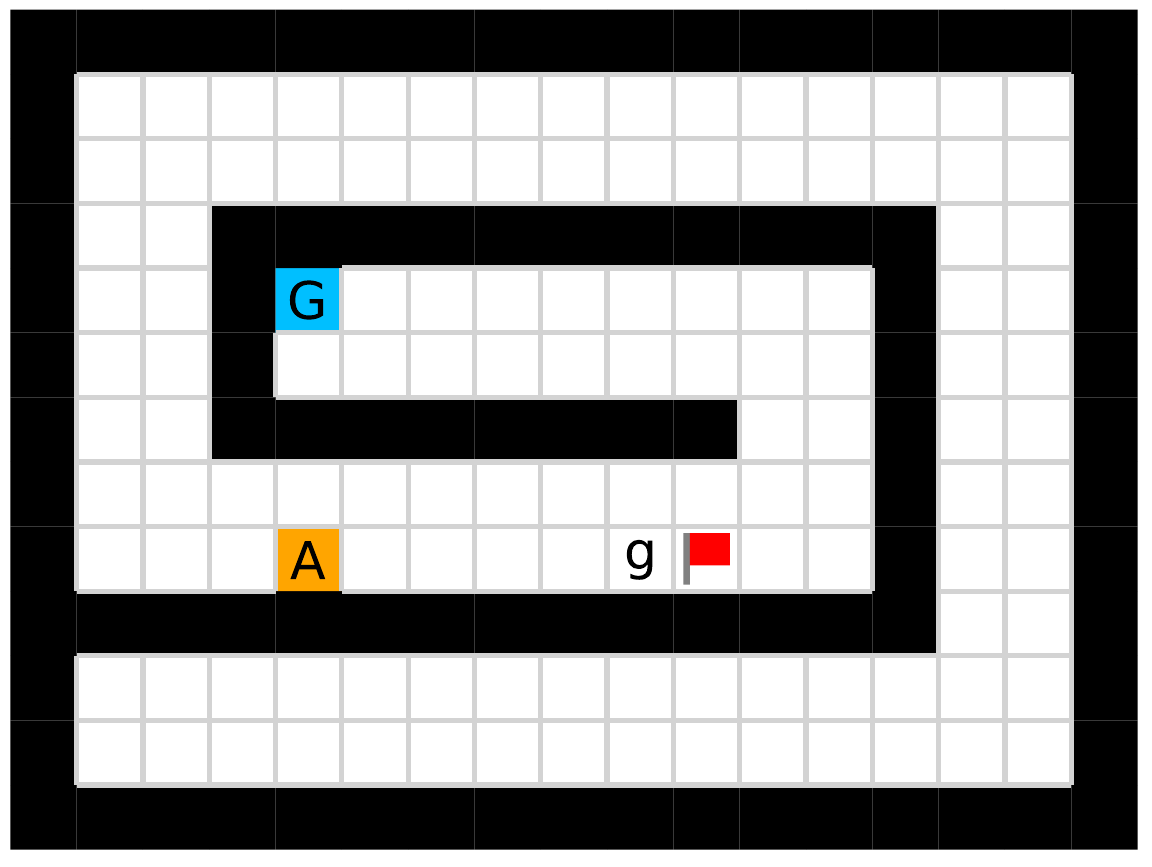}
\hspace{0.4em}
\includegraphics[width=0.18\linewidth]{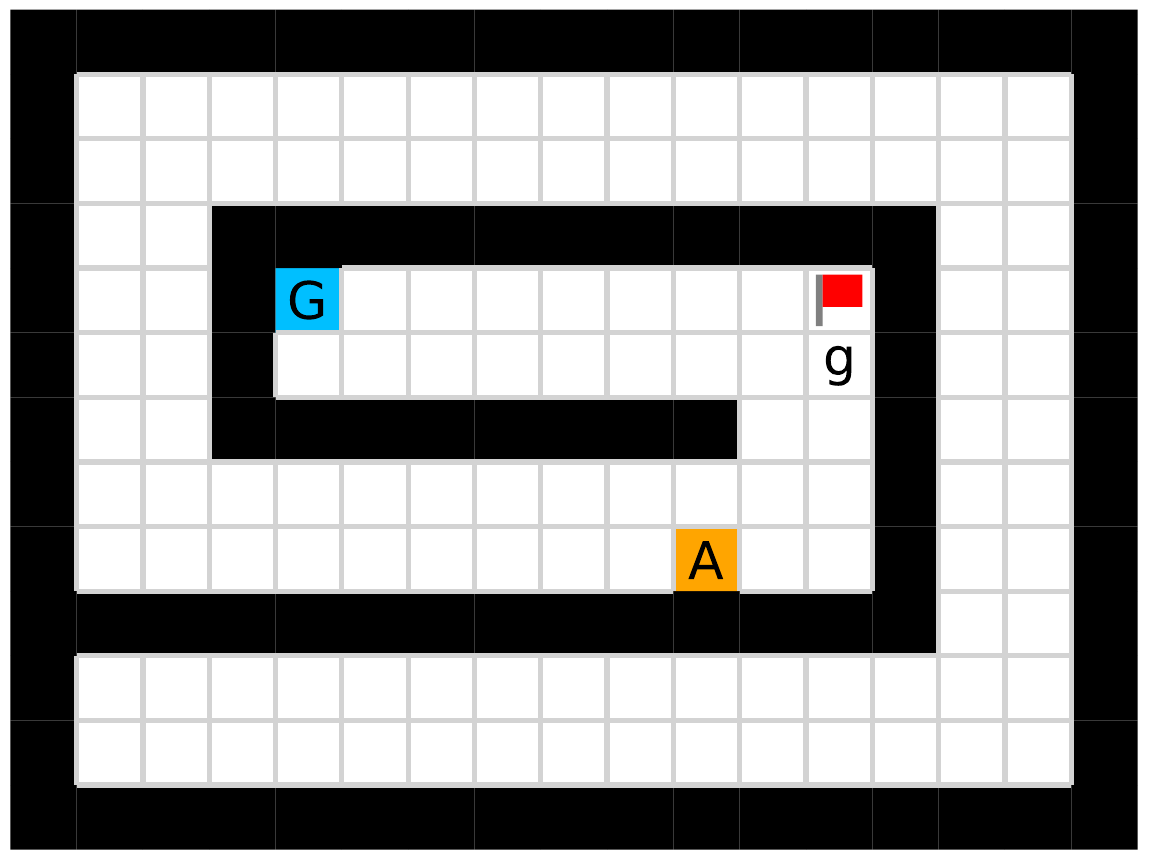}\\

\includegraphics[width=0.18\linewidth]{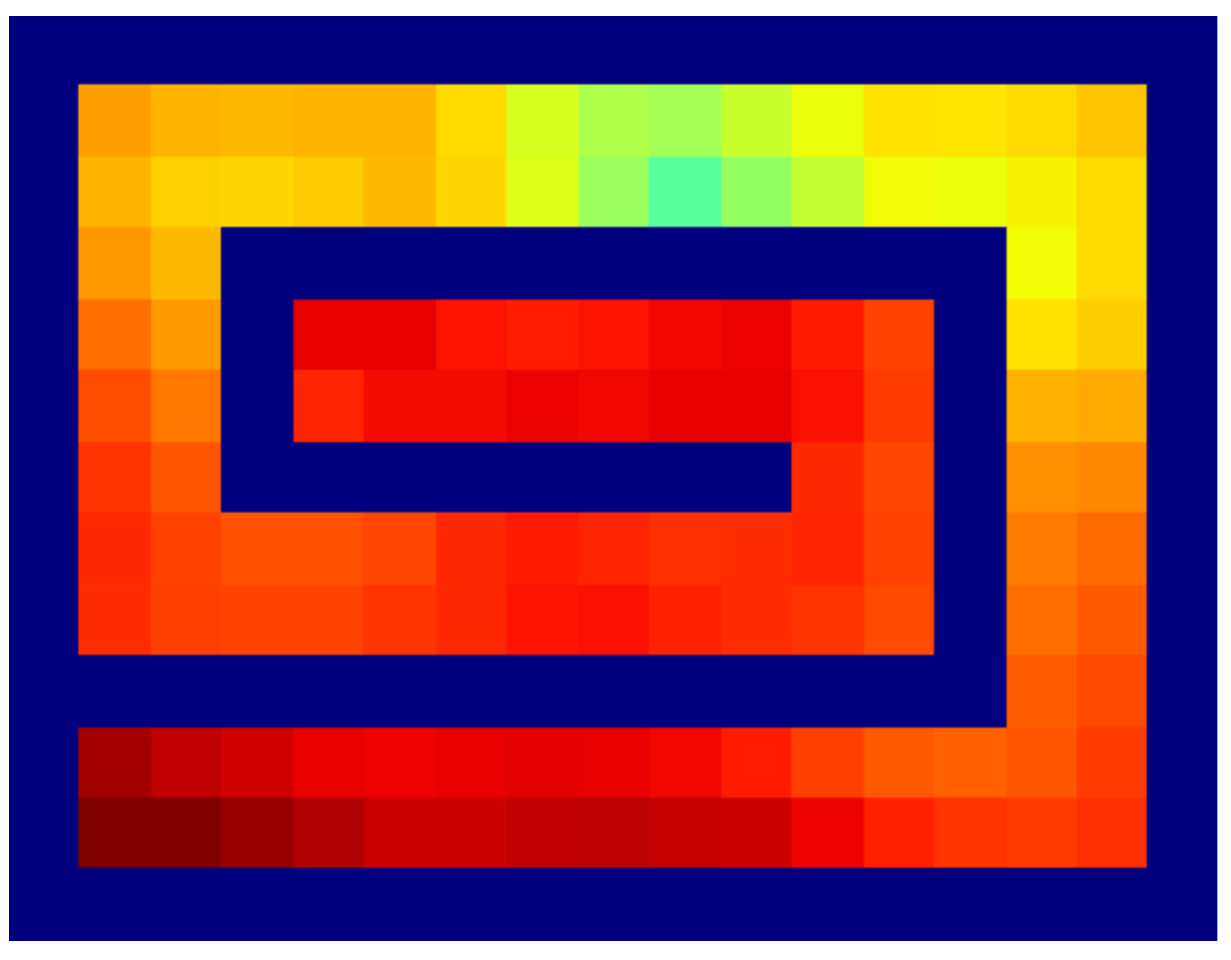}
\hspace{0.4em}
\vspace{2.5em}
\includegraphics[width=0.18\linewidth]{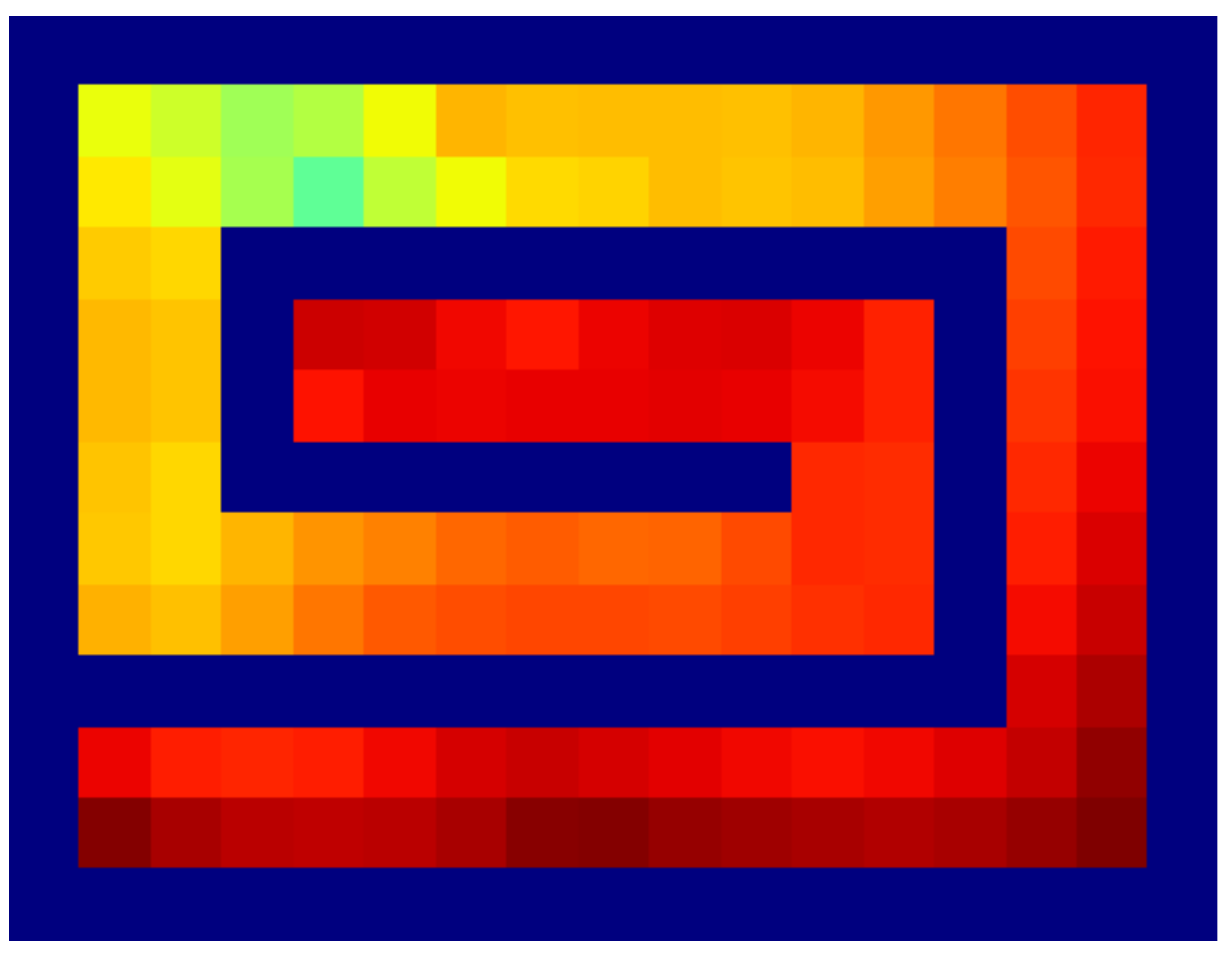}
\hspace{0.4em}
\includegraphics[width=0.18\linewidth]{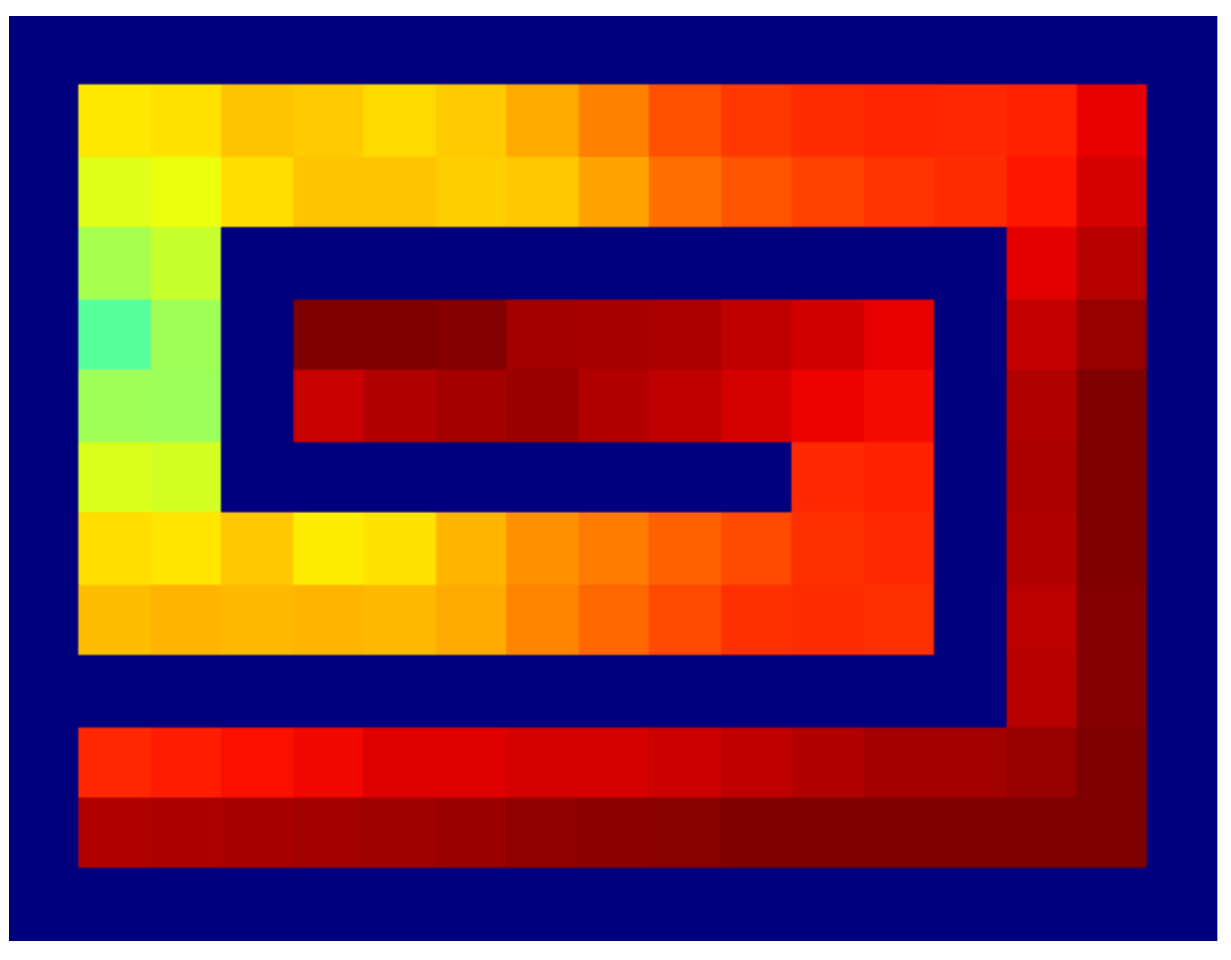}
\hspace{0.4em}
\includegraphics[width=0.18\linewidth]{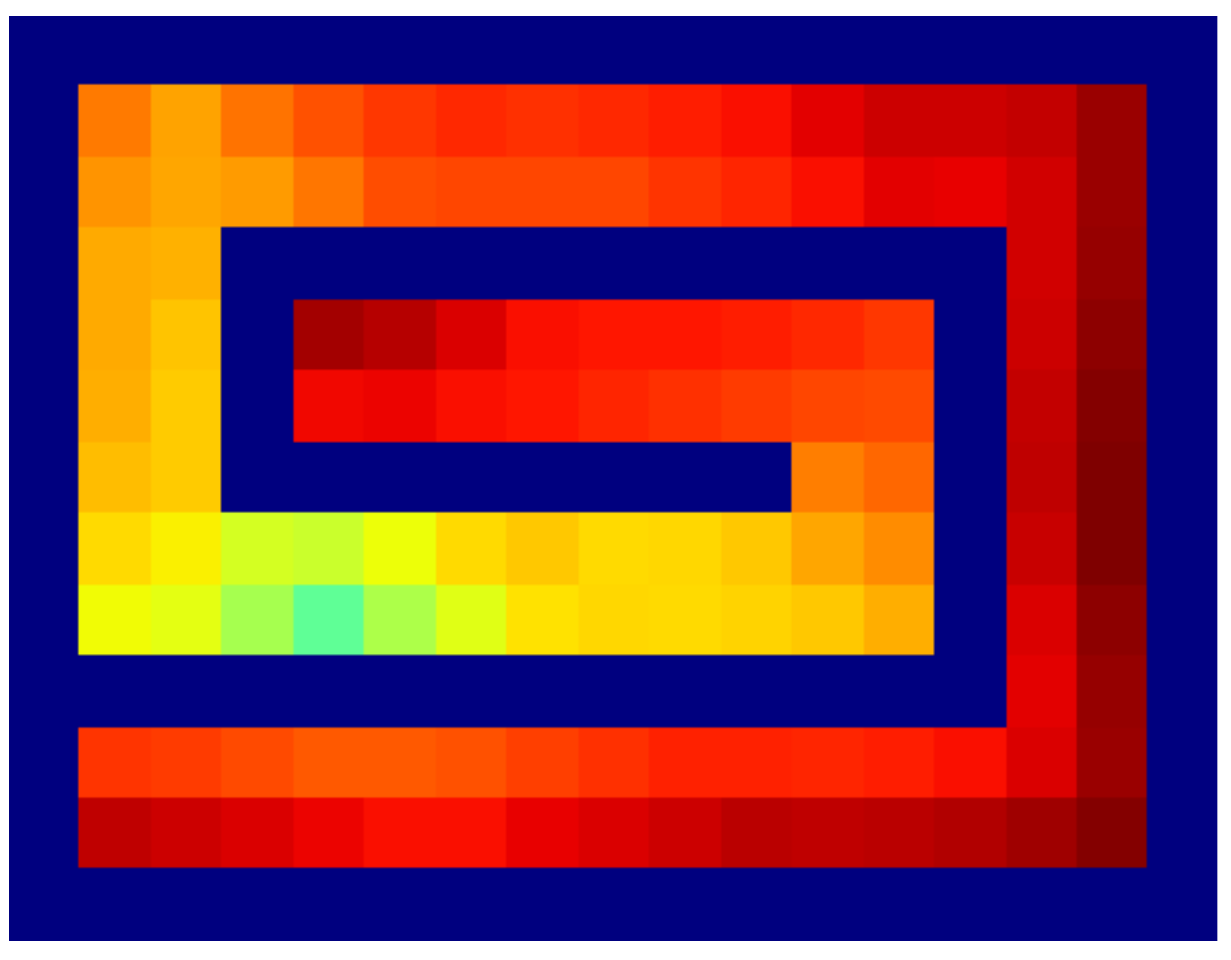}
\hspace{0.4em}
\includegraphics[width=0.18\linewidth]{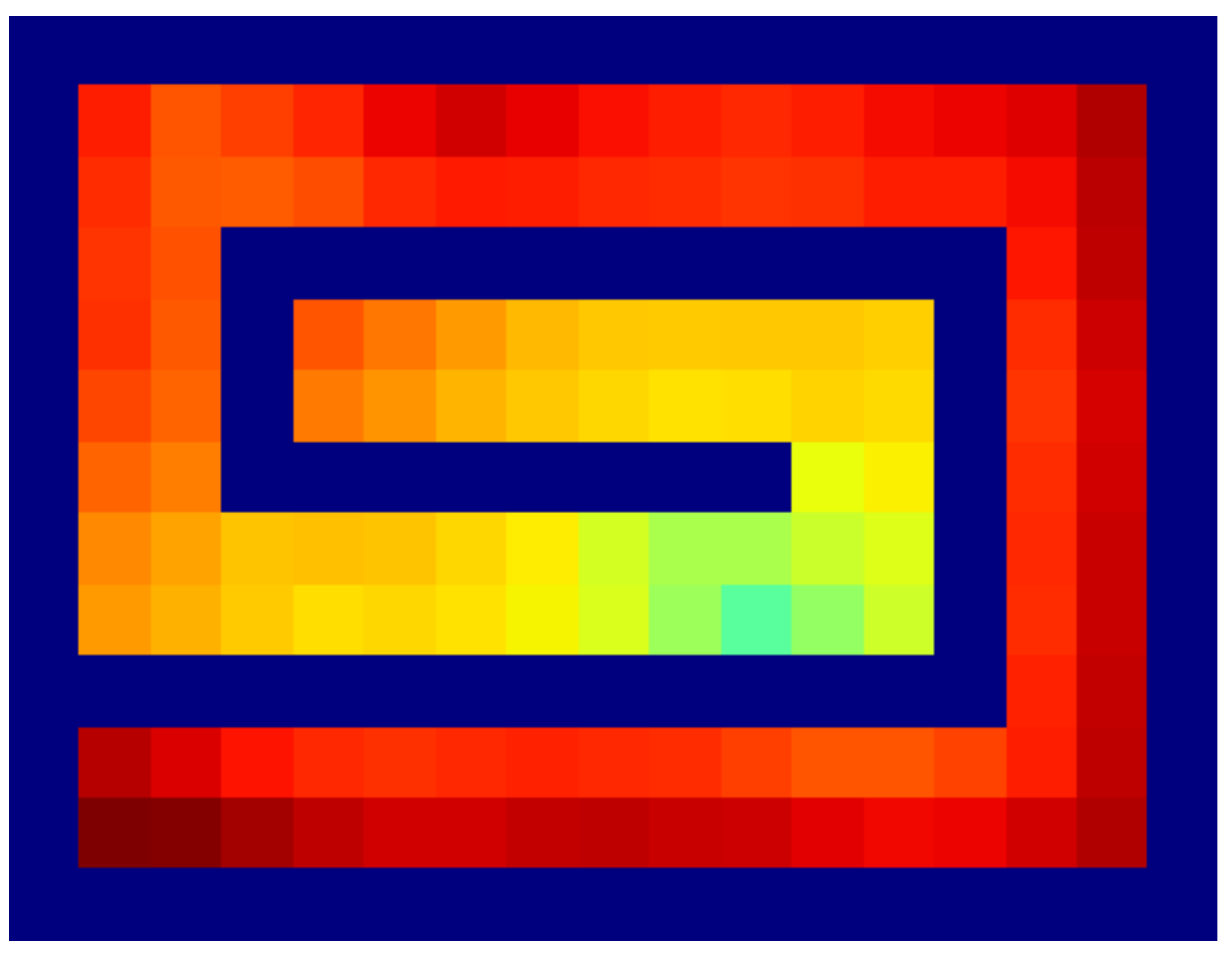}\\

\includegraphics[width=0.18\linewidth]{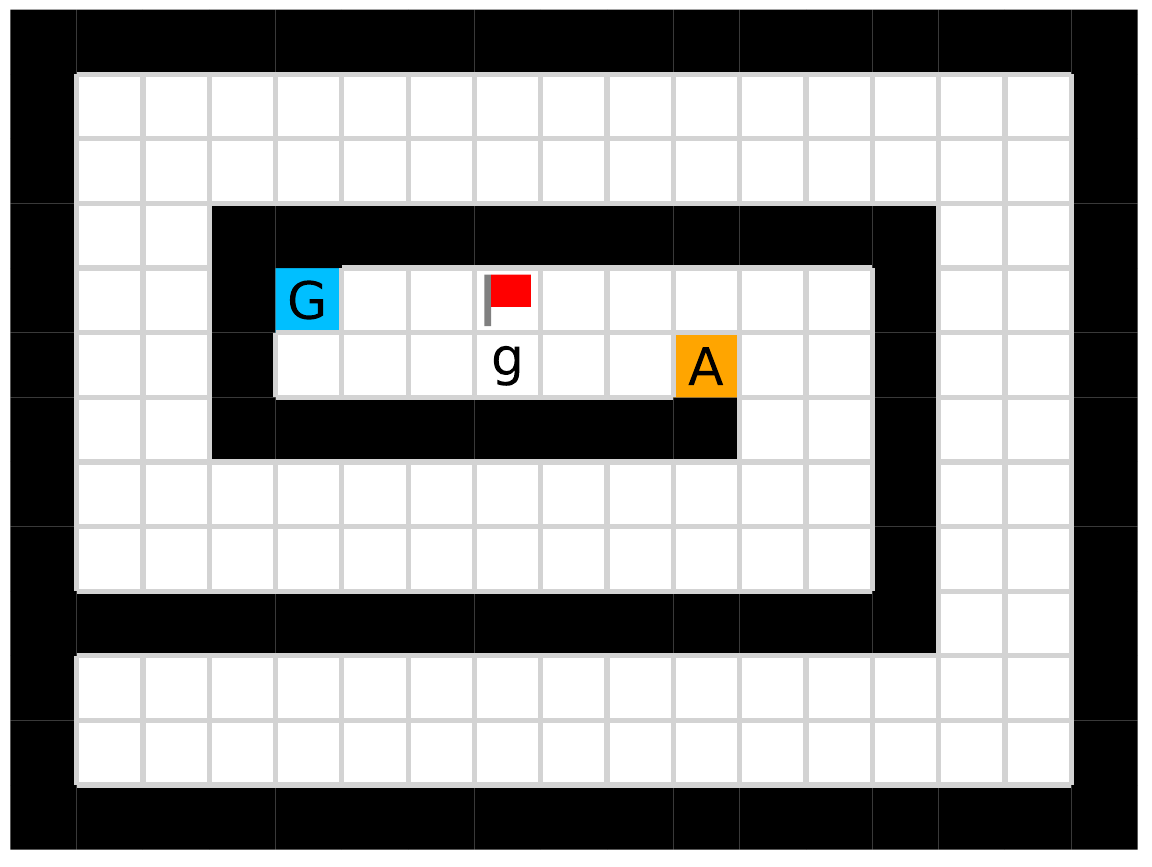}
\hspace{0.4em}
\vspace{0.5em}
\includegraphics[width=0.18\linewidth]{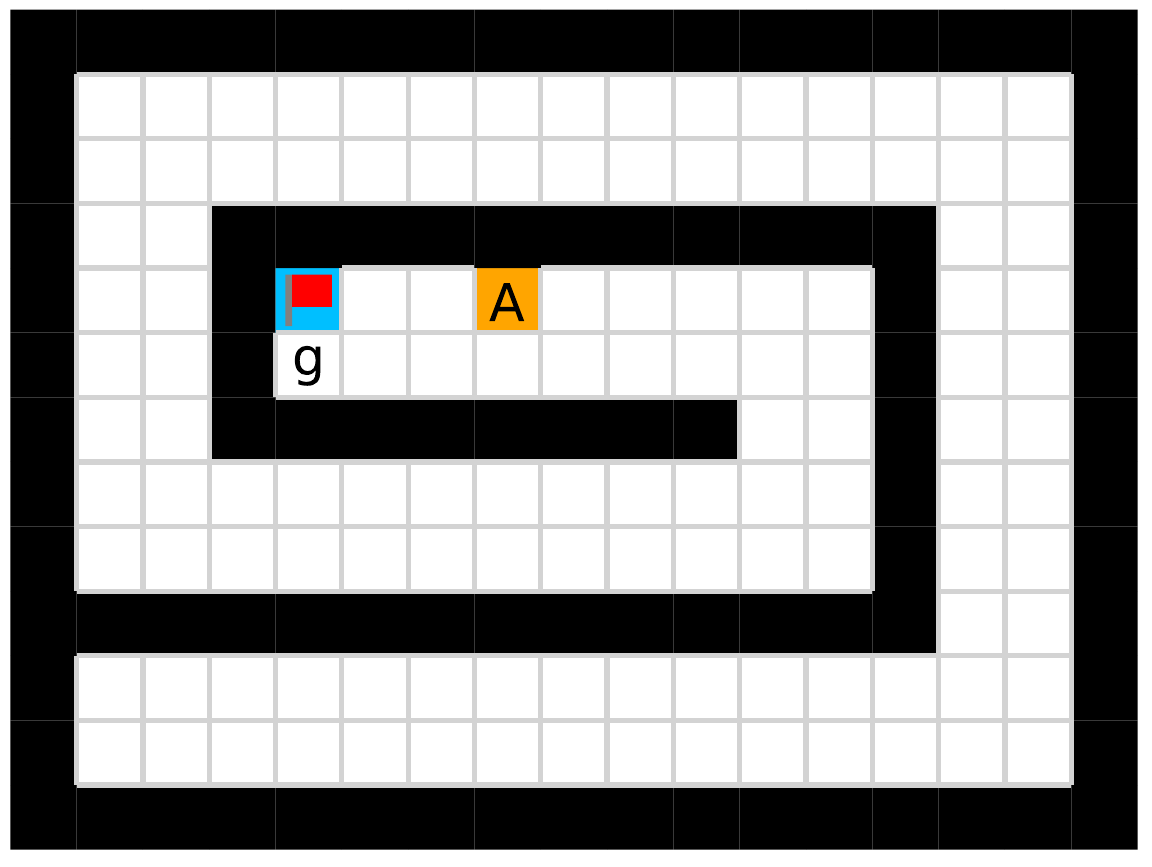}
\hspace{0.4em}
\includegraphics[width=0.18\linewidth]{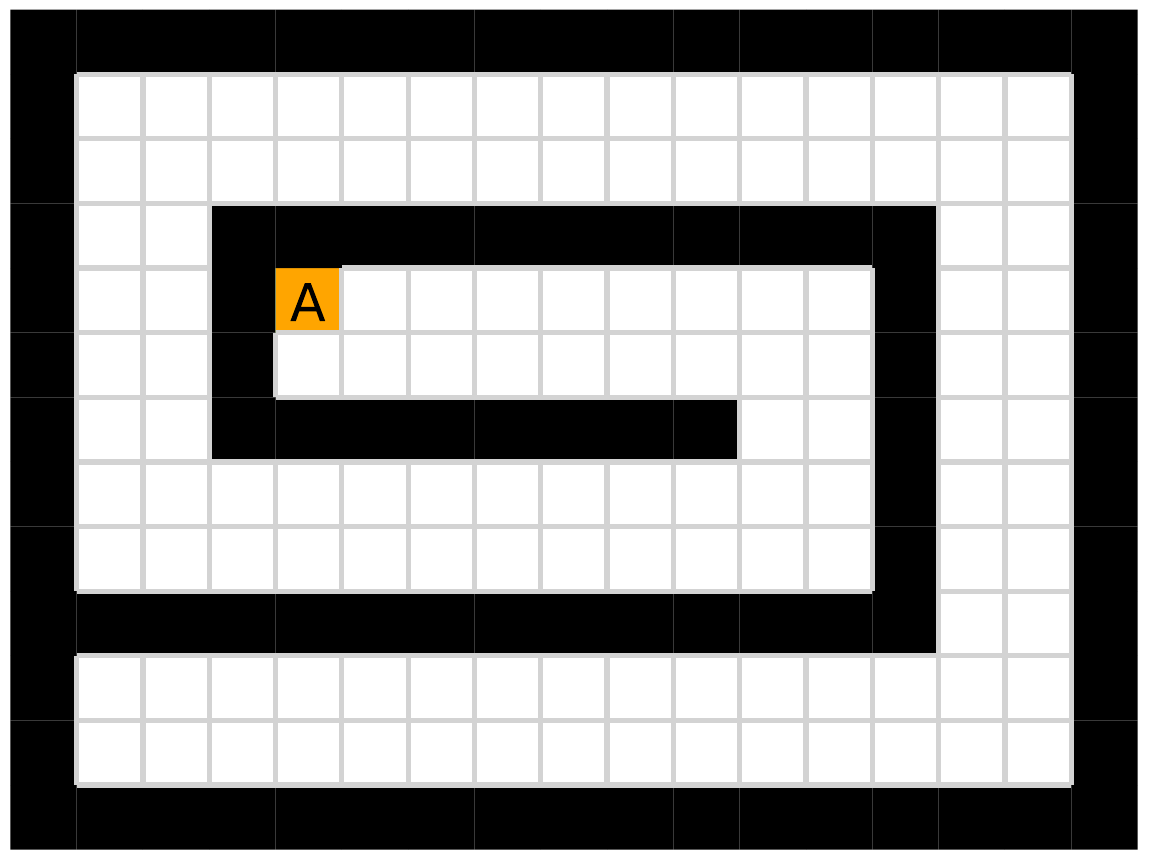}\\

\includegraphics[width=0.18\linewidth]{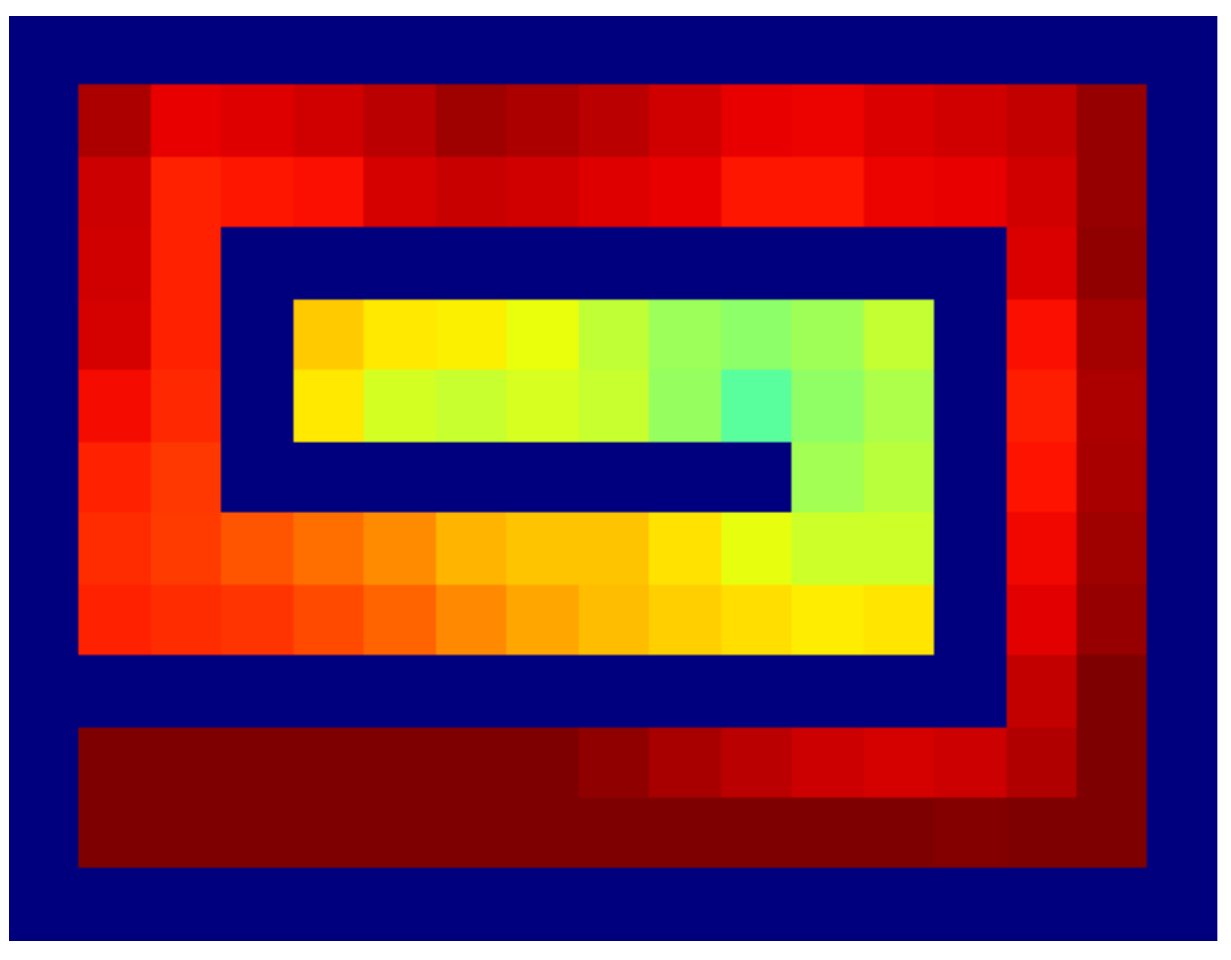}
\hspace{0.4em}
\vspace{0.5em}
\includegraphics[width=0.18\linewidth]{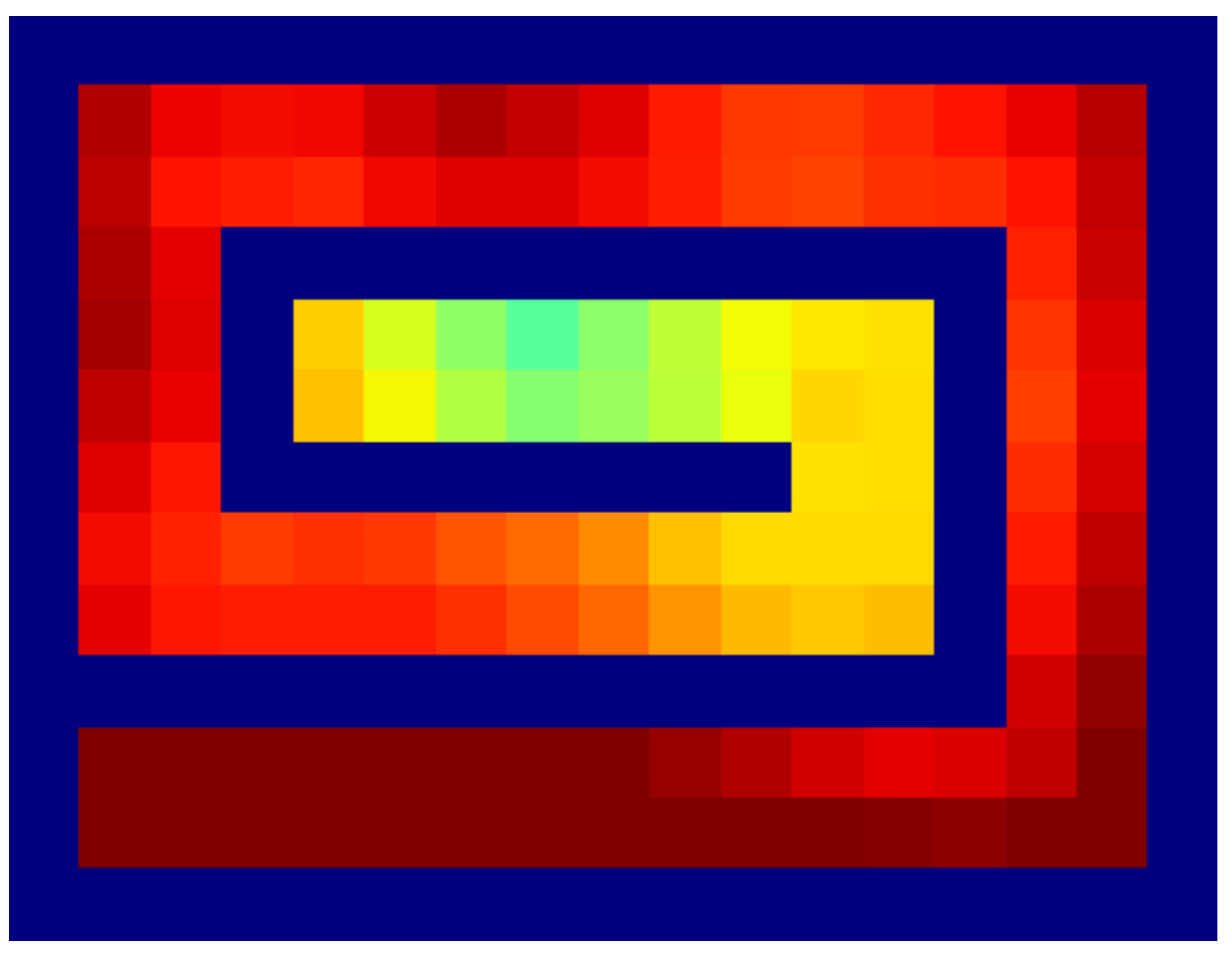}
\hspace{0.4em}
\includegraphics[width=0.18\linewidth]{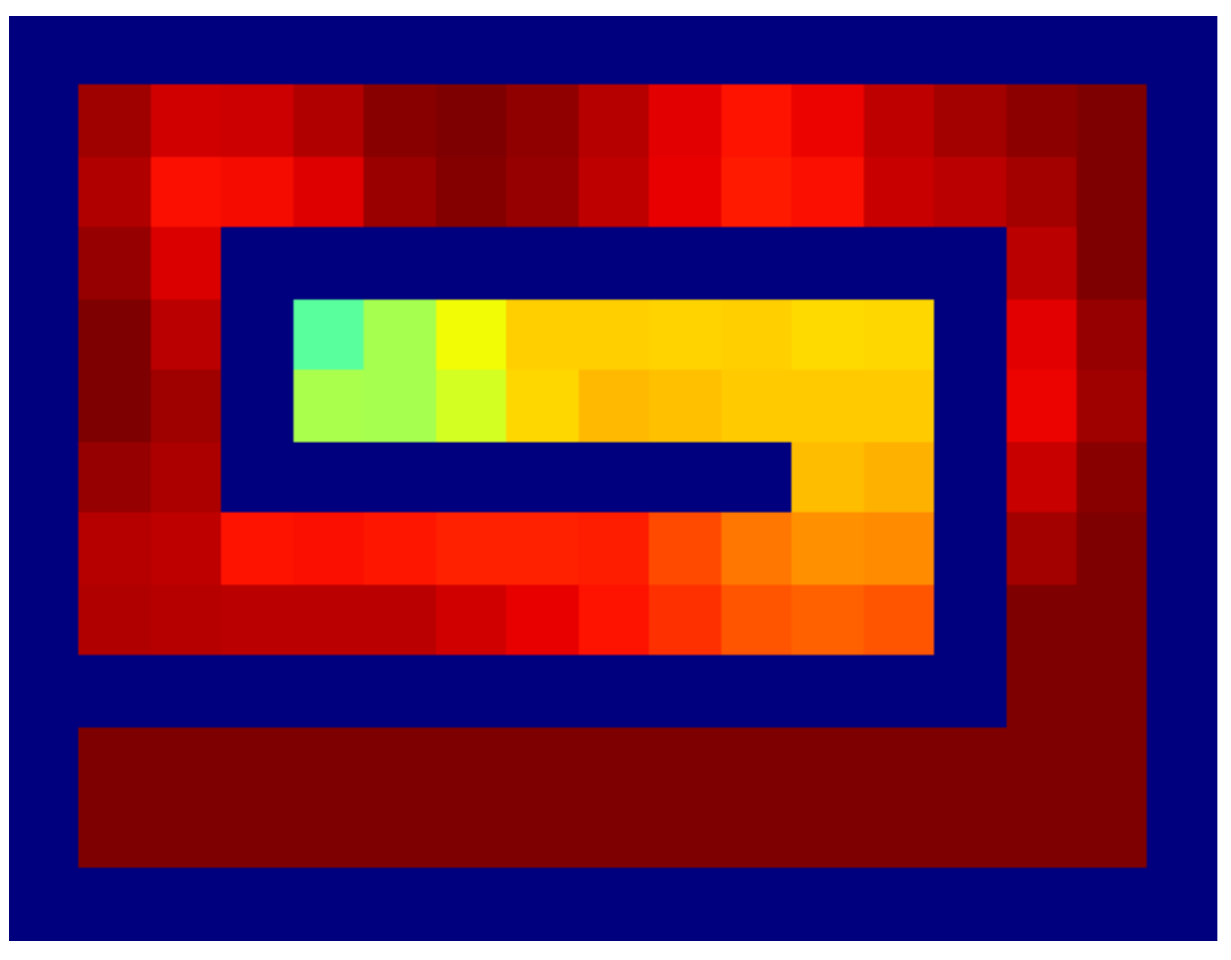}\\

\caption{Additional subgoal and adjacency heatmap visualizations of the Maze task, based on a single evaluation run. The agent (A), goal (G) and subgoal (g) at different time steps in one episode are plotted.  Colder colors in the adjacency heatmaps represent smaller shortest transition distances.}
\label{fig:visualization_supp_maze}
\end{figure}

\begin{figure}
\centering
\includegraphics[width=0.18\linewidth]{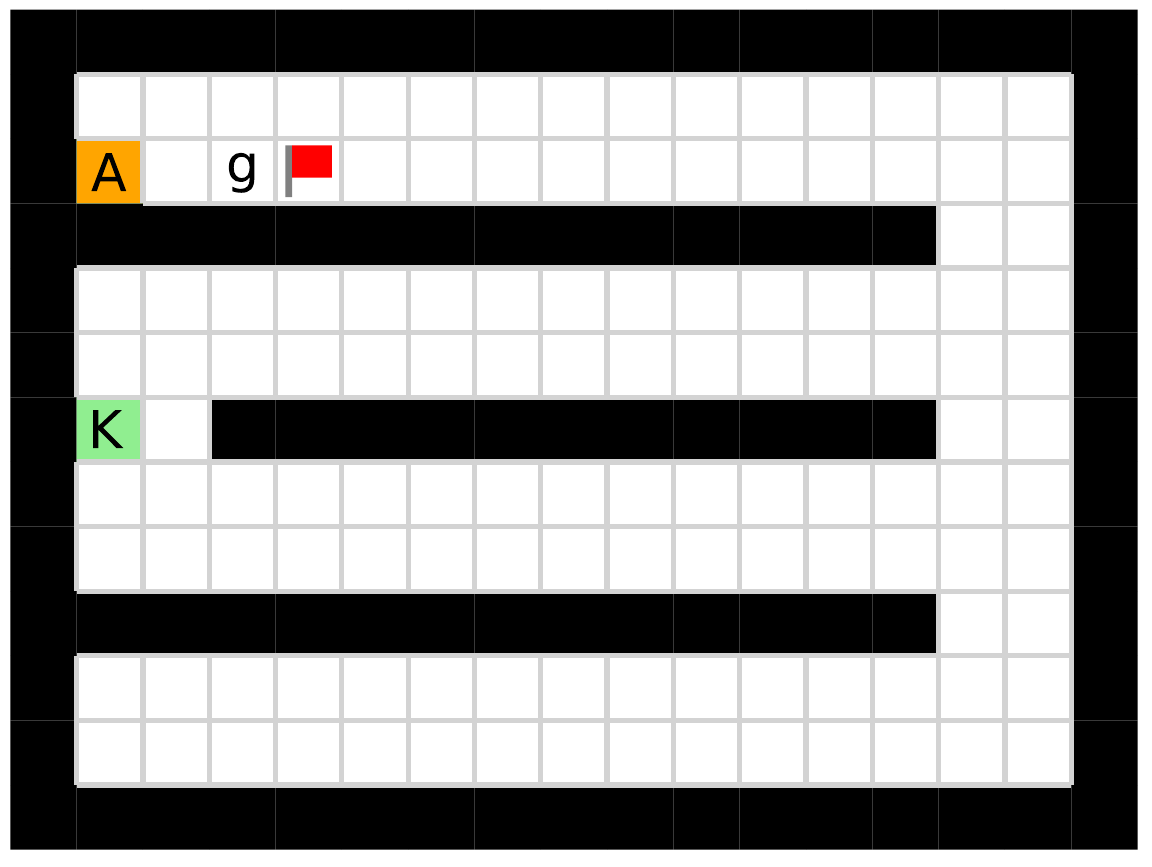}
\hspace{0.4em}
\vspace{0.4em}
\includegraphics[width=0.18\linewidth]{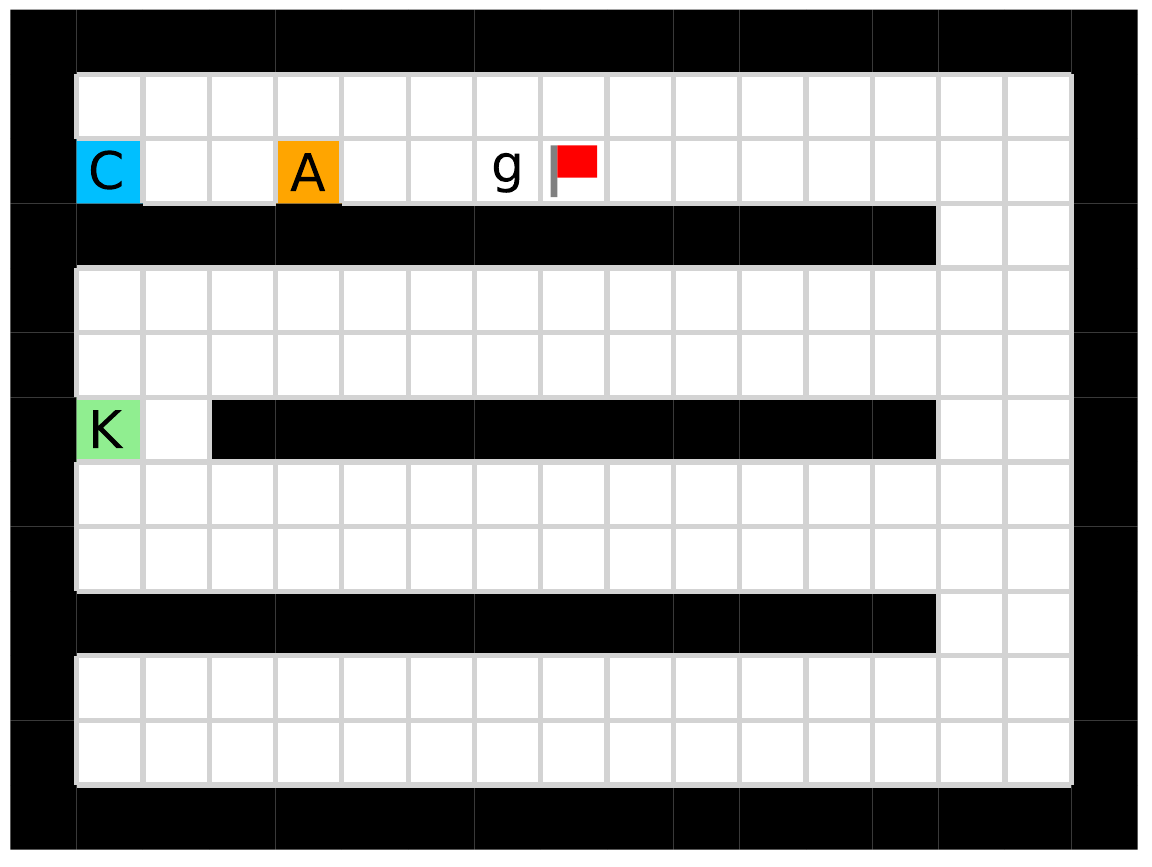}
\hspace{0.4em}
\includegraphics[width=0.18\linewidth]{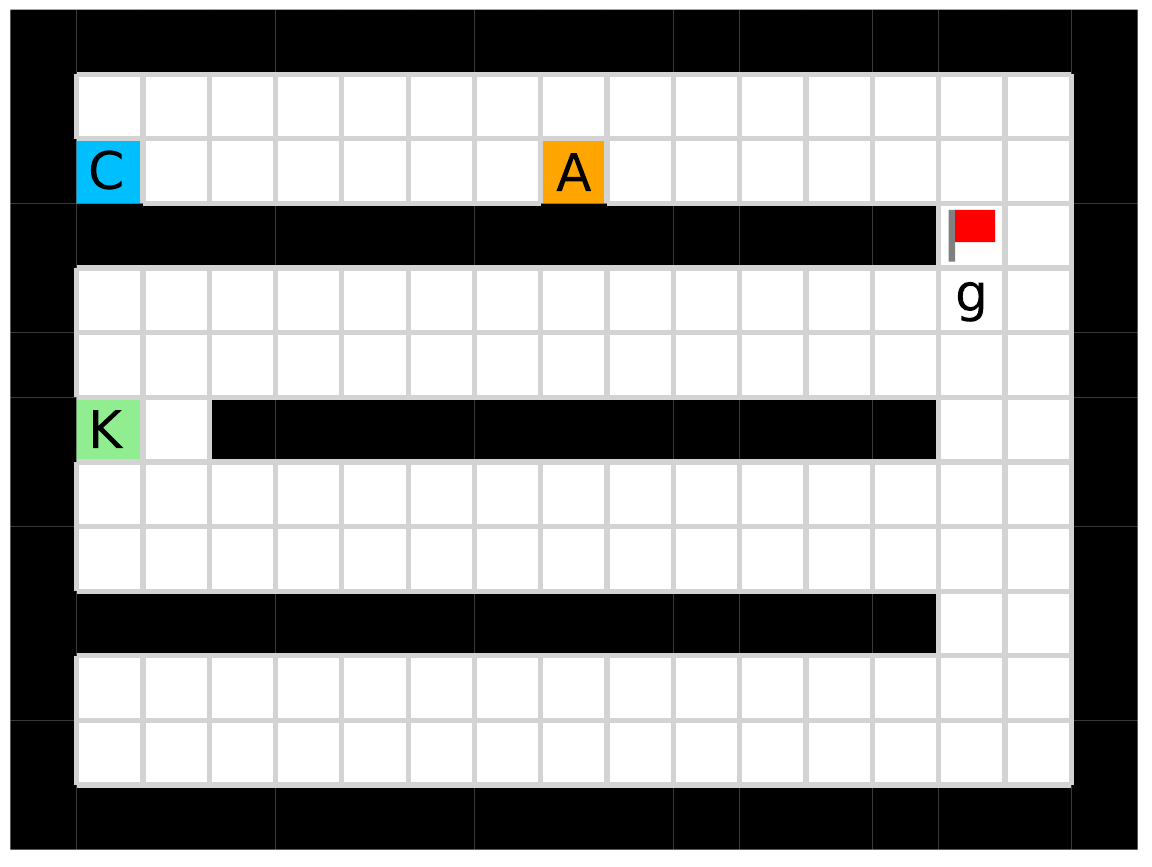}
\hspace{0.4em}
\includegraphics[width=0.18\linewidth]{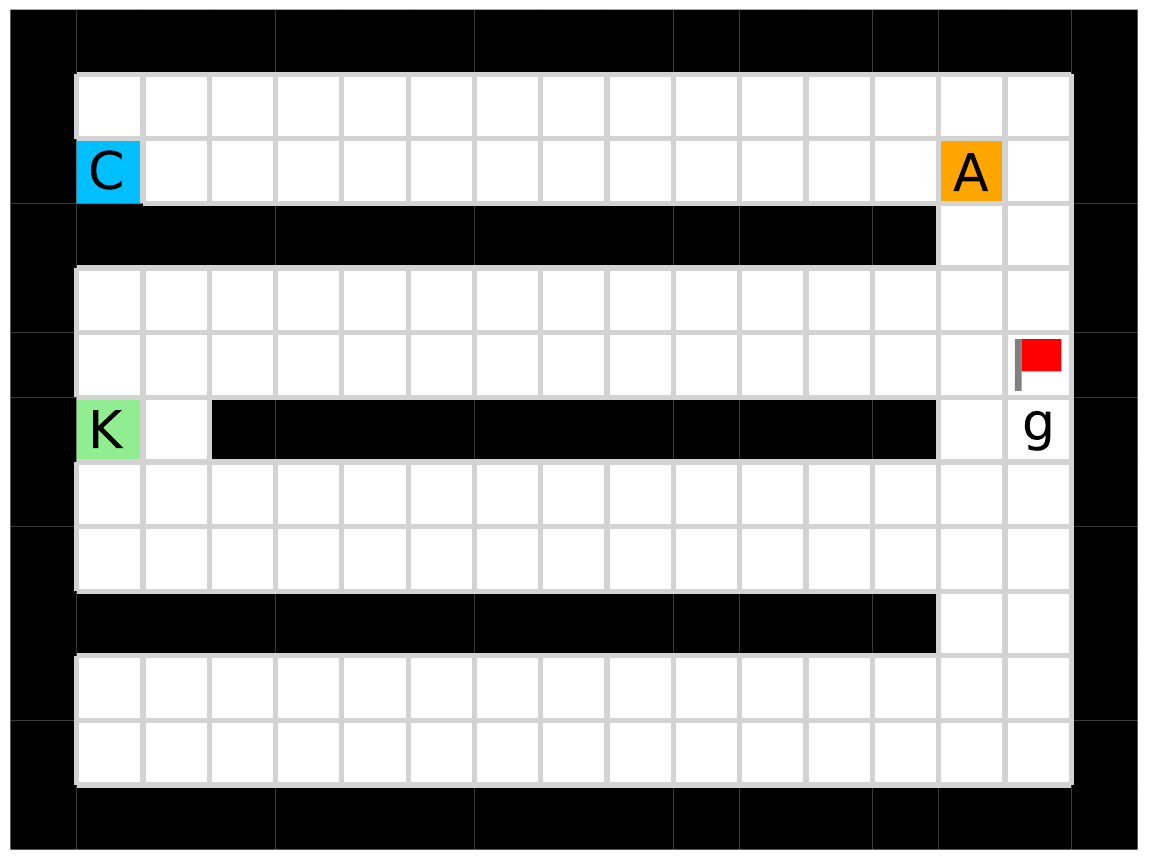}
\hspace{0.4em}
\includegraphics[width=0.18\linewidth]{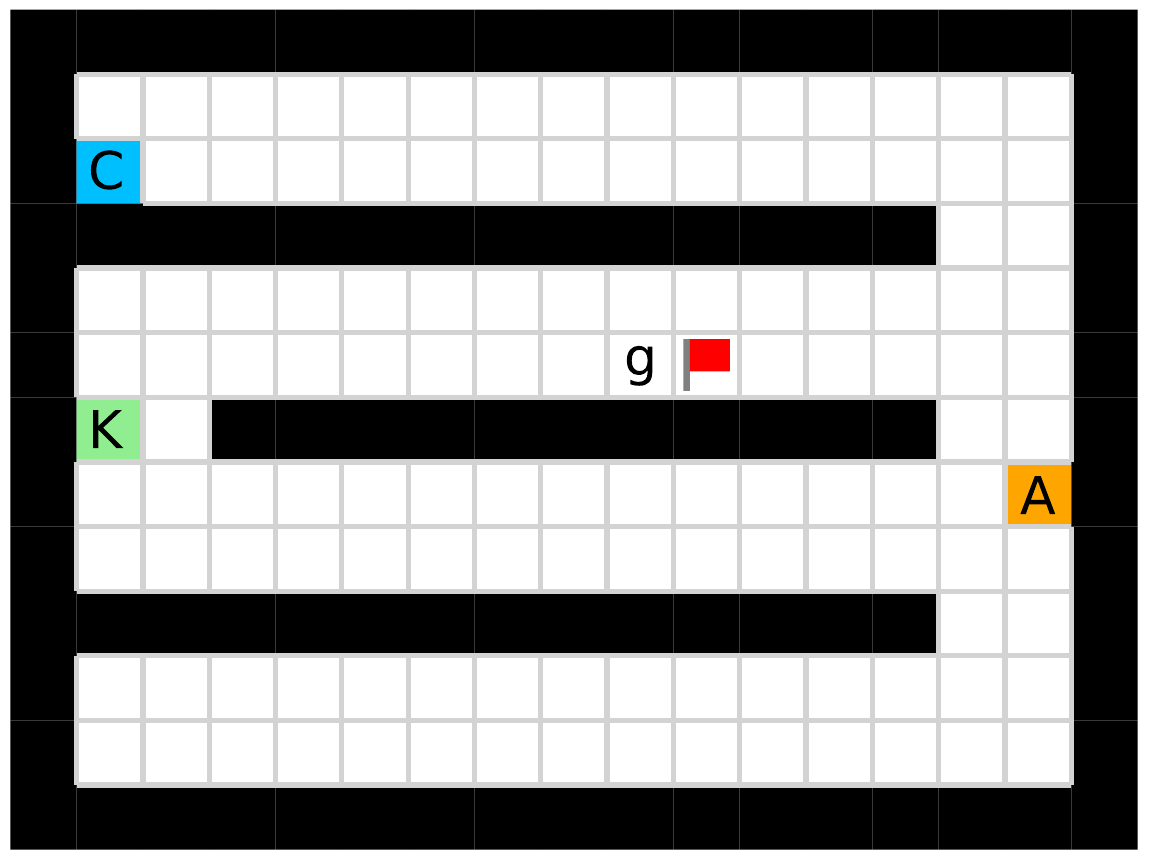}\\

\includegraphics[width=0.18\linewidth]{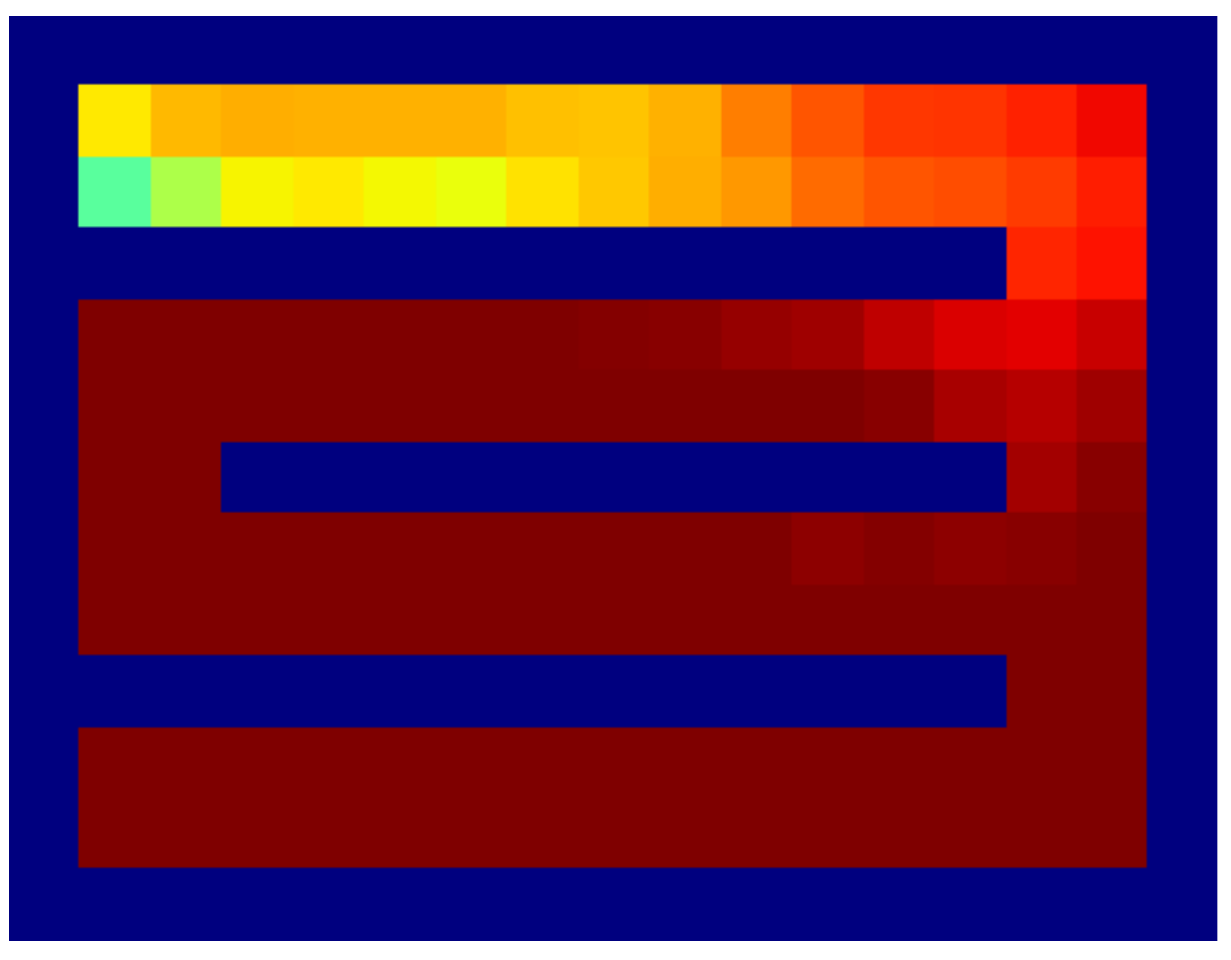}
\hspace{0.4em}
\vspace{1.7em}
\includegraphics[width=0.18\linewidth]{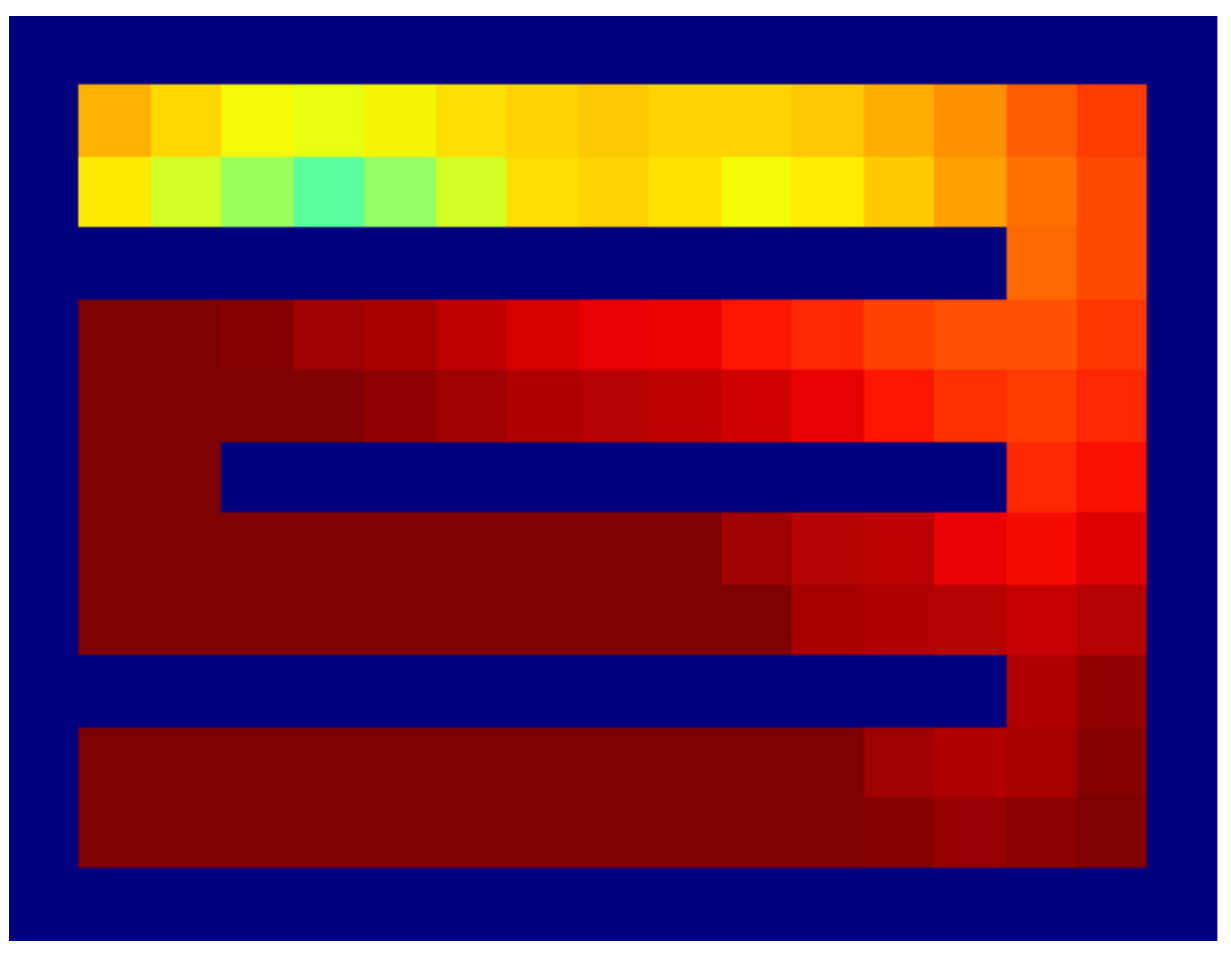}
\hspace{0.4em}
\includegraphics[width=0.18\linewidth]{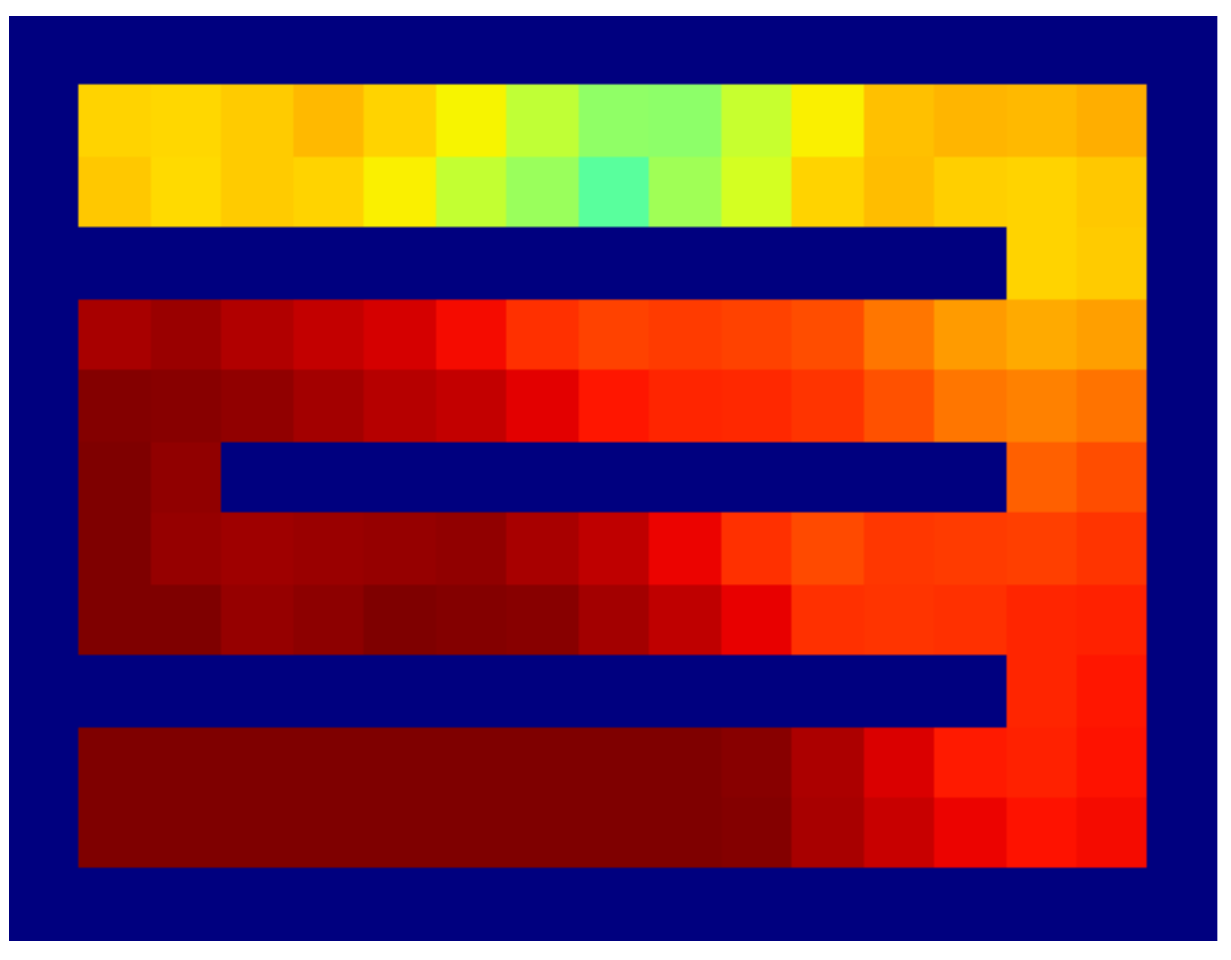}
\hspace{0.4em}
\includegraphics[width=0.18\linewidth]{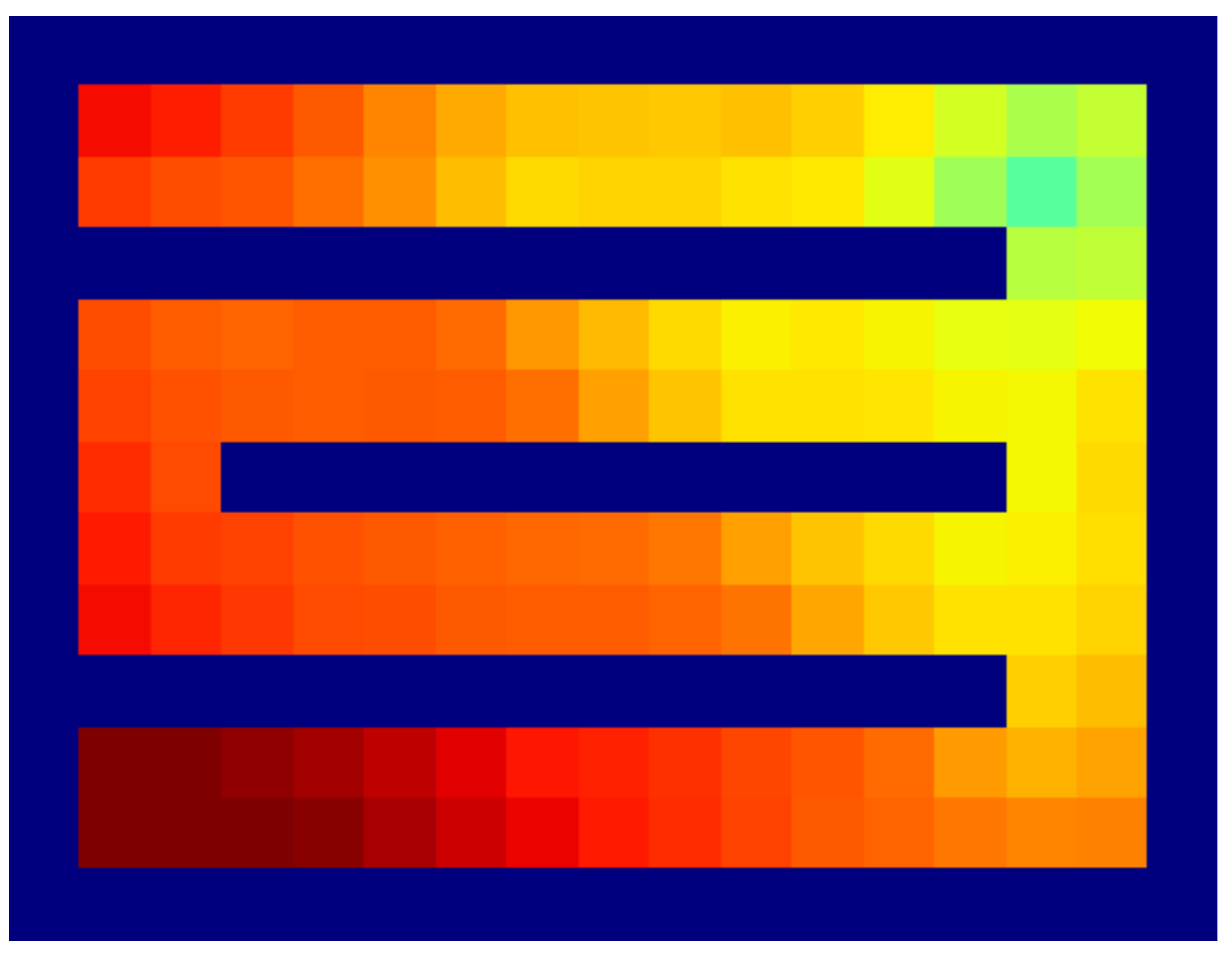}
\hspace{0.4em}
\includegraphics[width=0.18\linewidth]{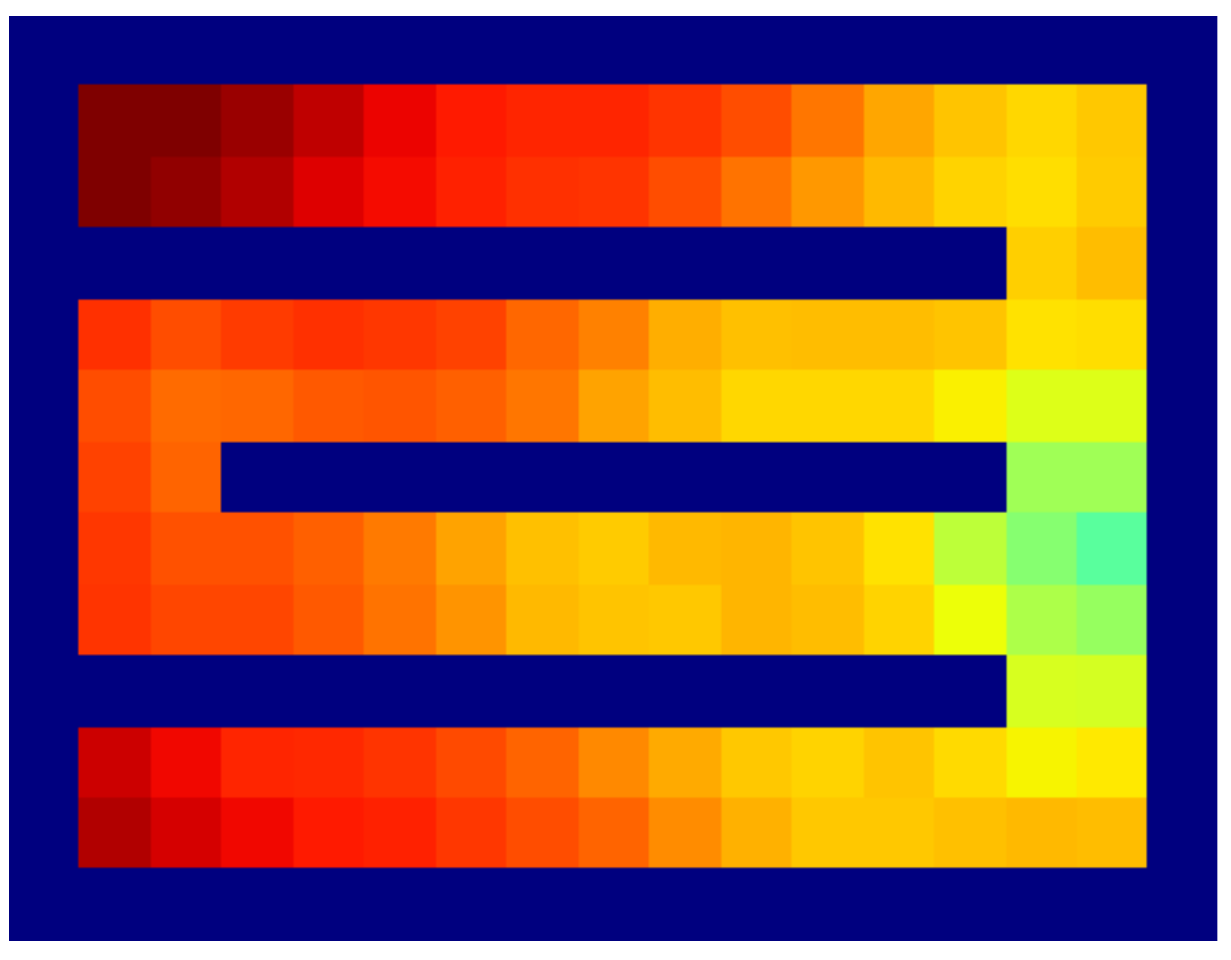}\\

\includegraphics[width=0.18\linewidth]{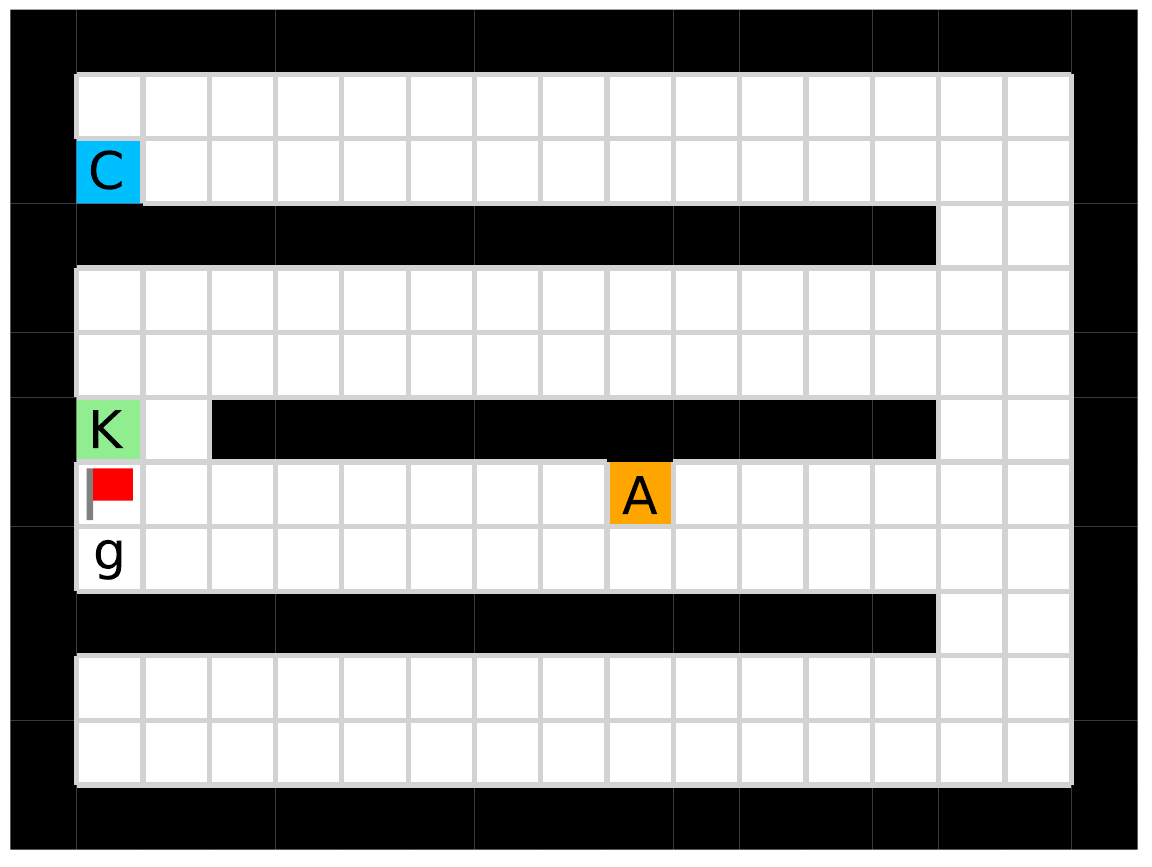}
\hspace{0.4em}
\vspace{0.4em}
\includegraphics[width=0.18\linewidth]{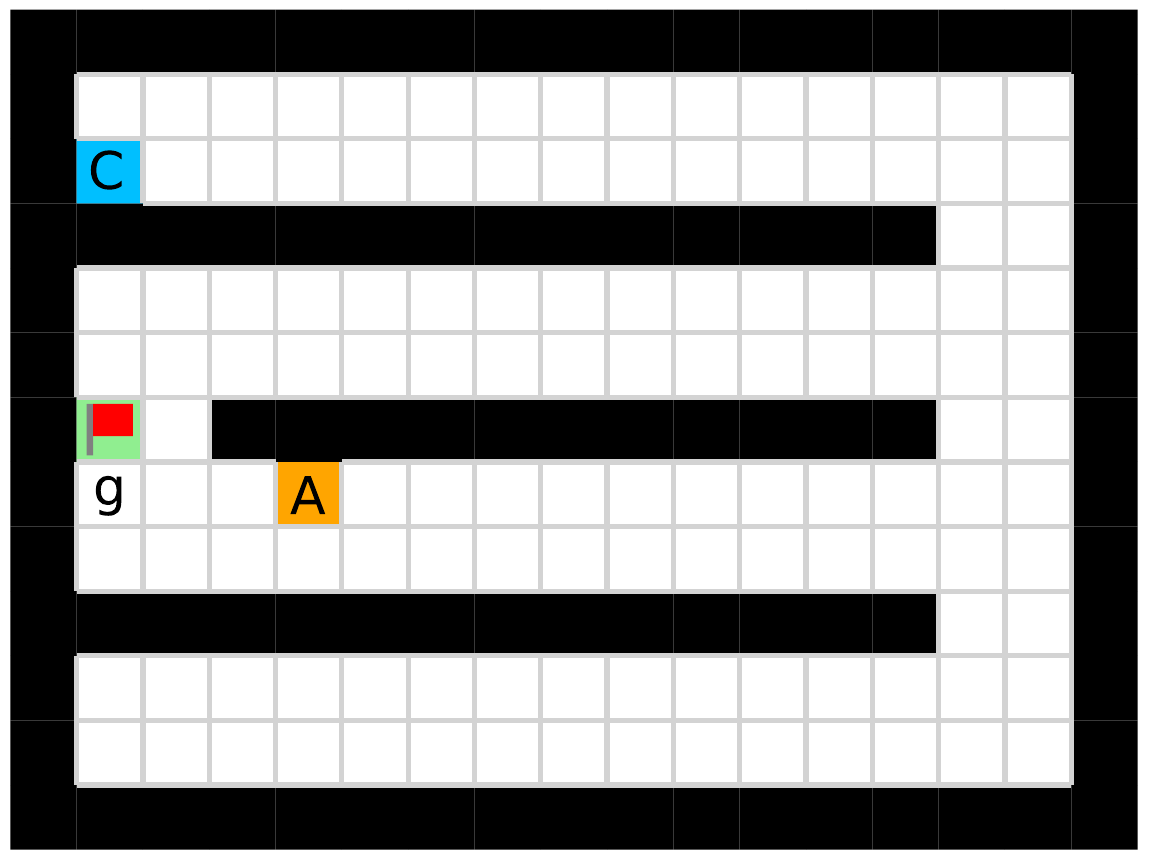}
\hspace{0.4em}
\includegraphics[width=0.18\linewidth]{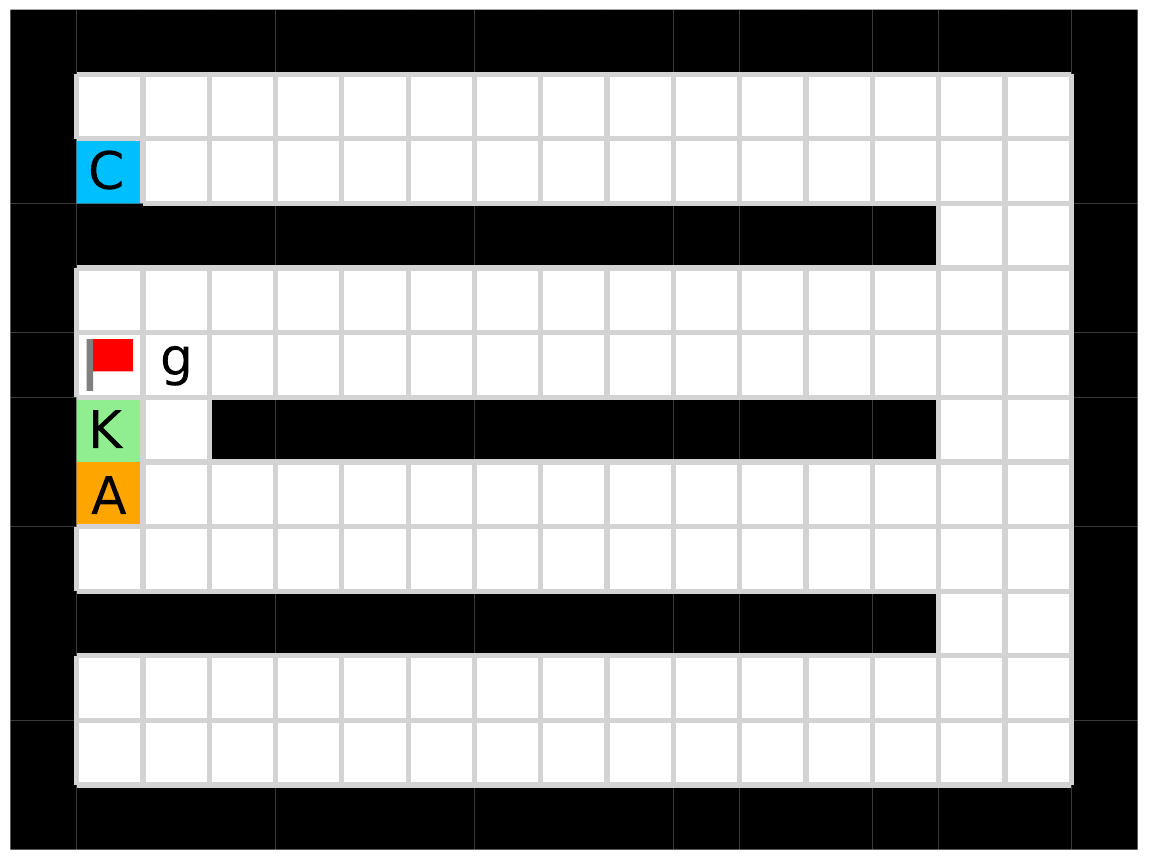}
\hspace{0.4em}
\includegraphics[width=0.18\linewidth]{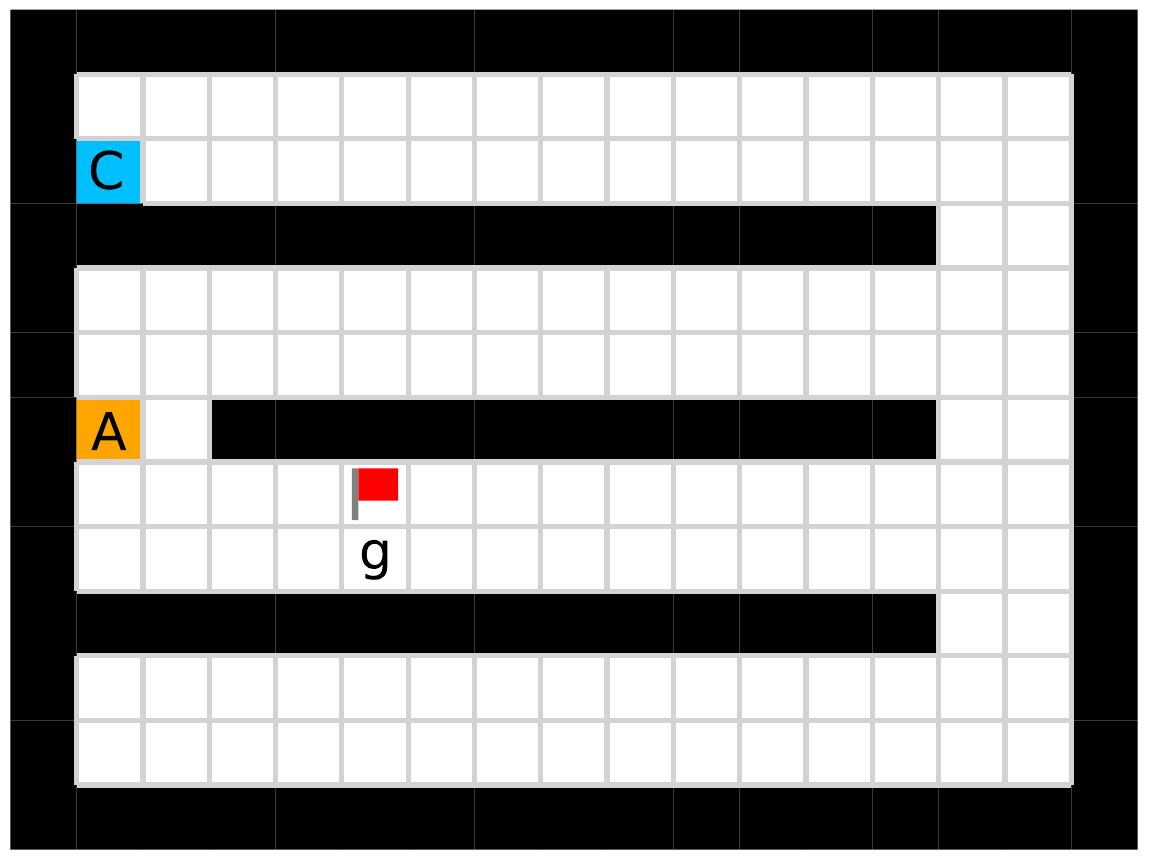}
\hspace{0.4em}
\includegraphics[width=0.18\linewidth]{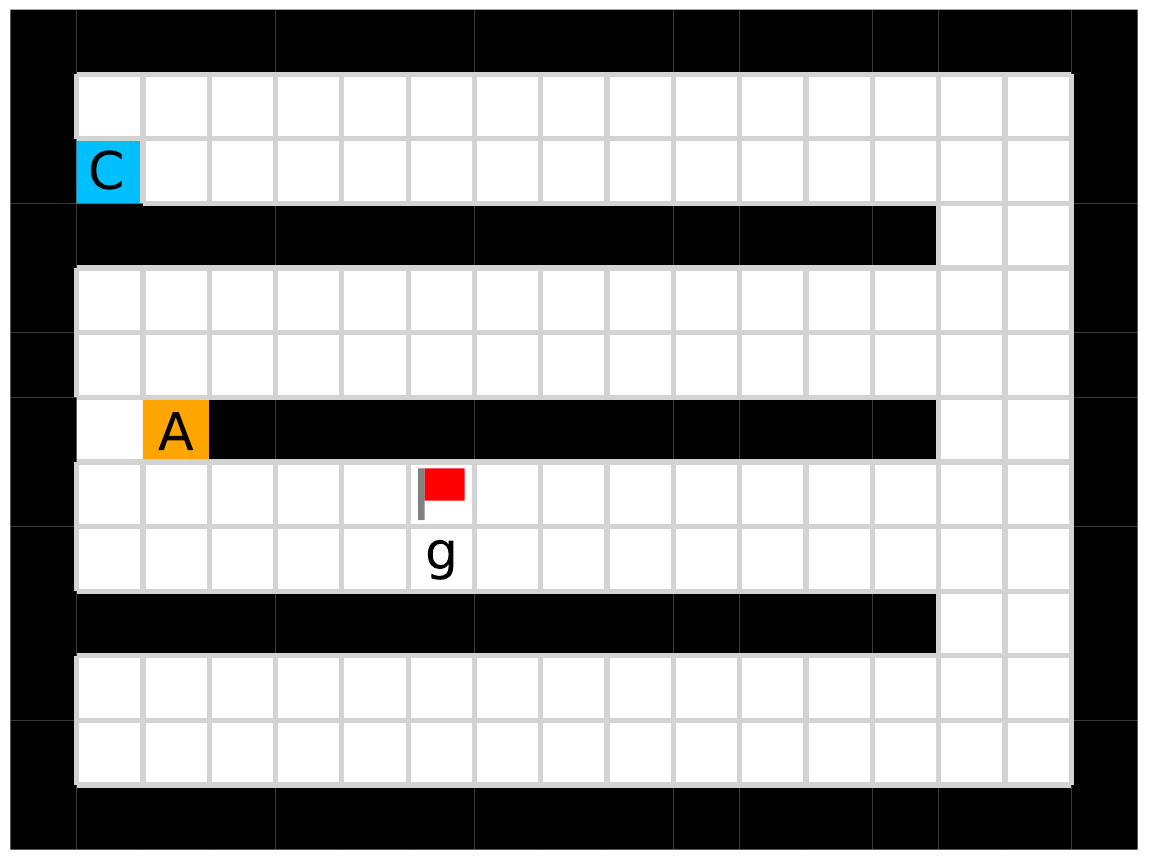}\\

\includegraphics[width=0.18\linewidth]{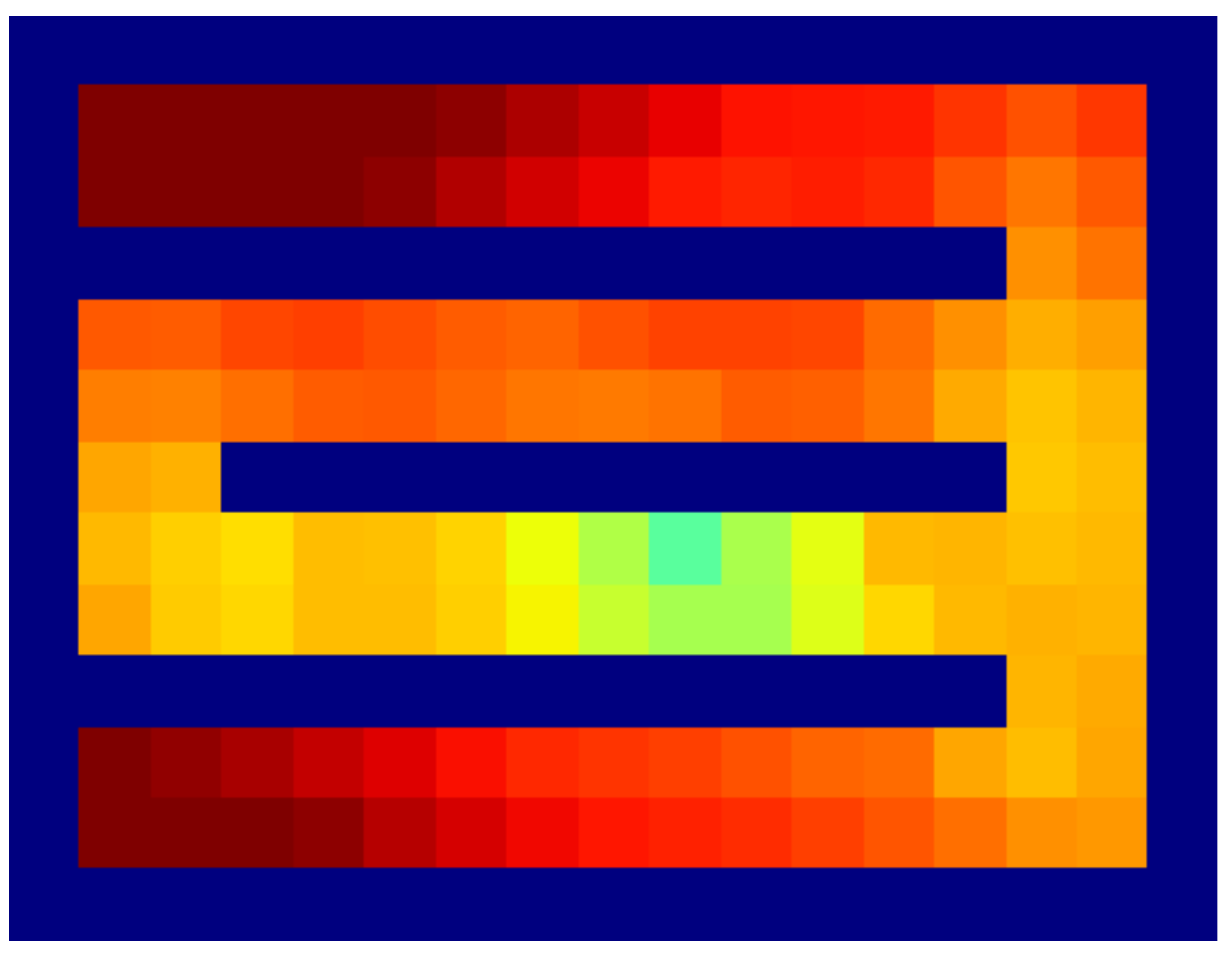}
\hspace{0.4em}
\vspace{1.7em}
\includegraphics[width=0.18\linewidth]{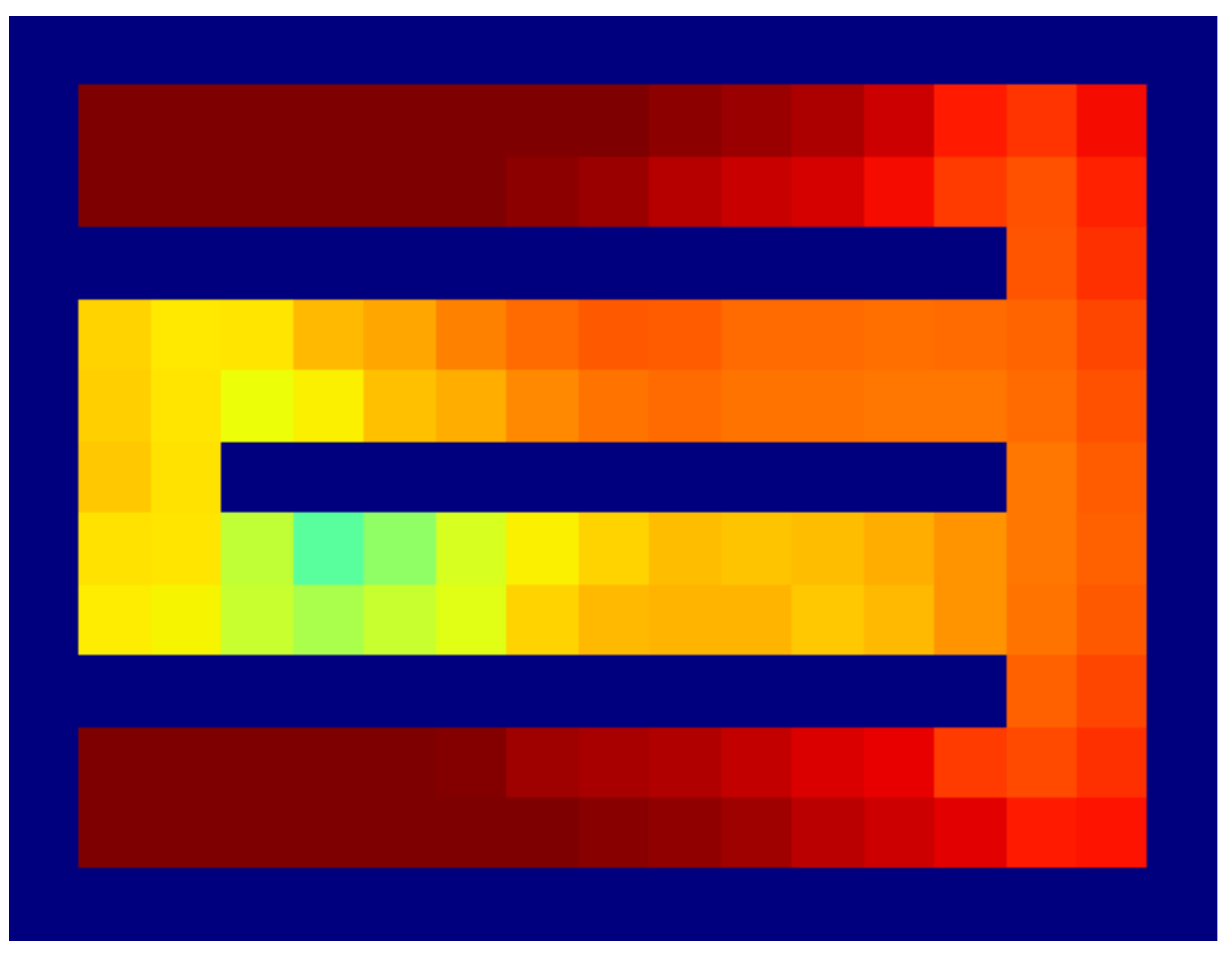}
\hspace{0.4em}
\includegraphics[width=0.18\linewidth]{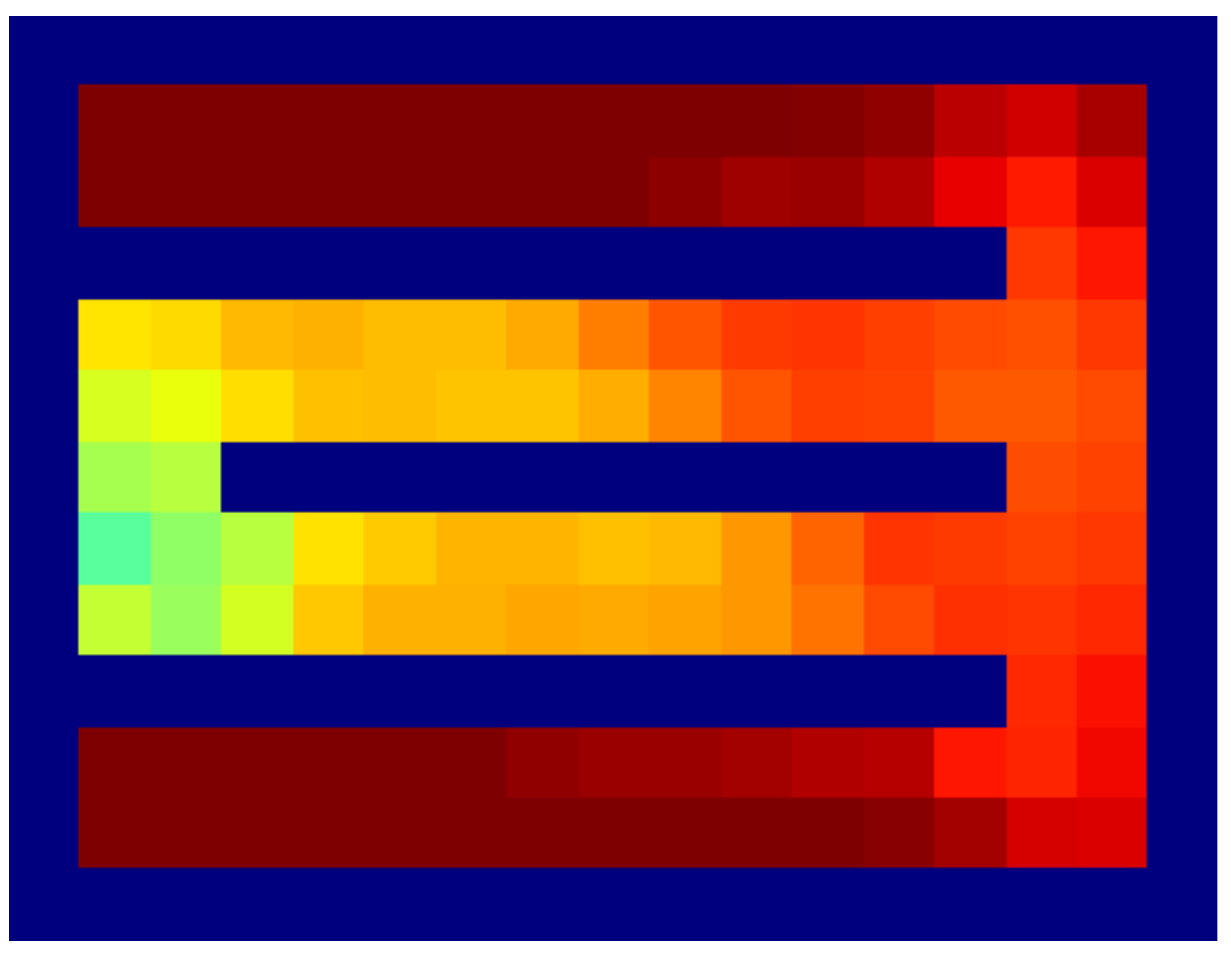}
\hspace{0.4em}
\includegraphics[width=0.18\linewidth]{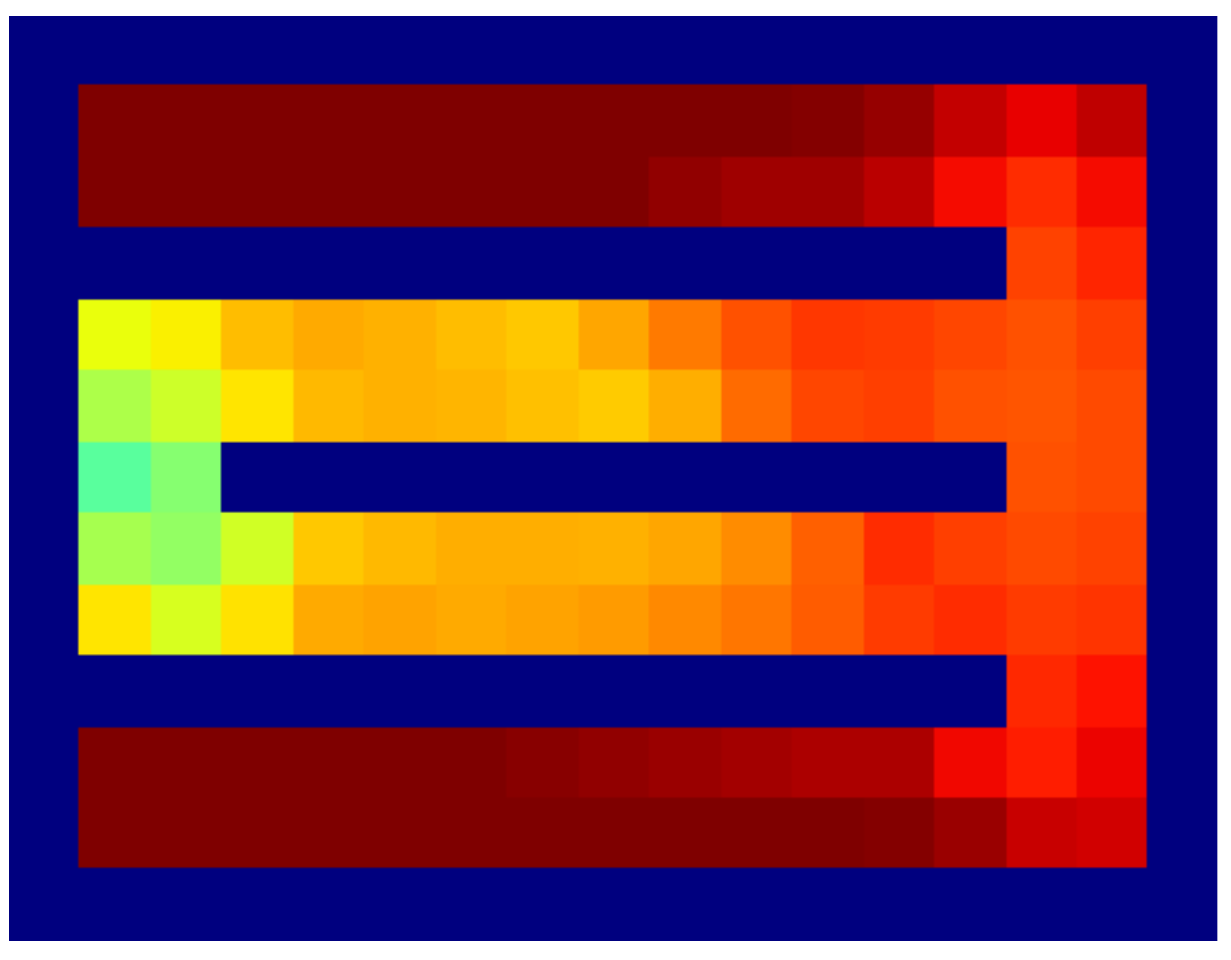}
\hspace{0.4em}
\includegraphics[width=0.18\linewidth]{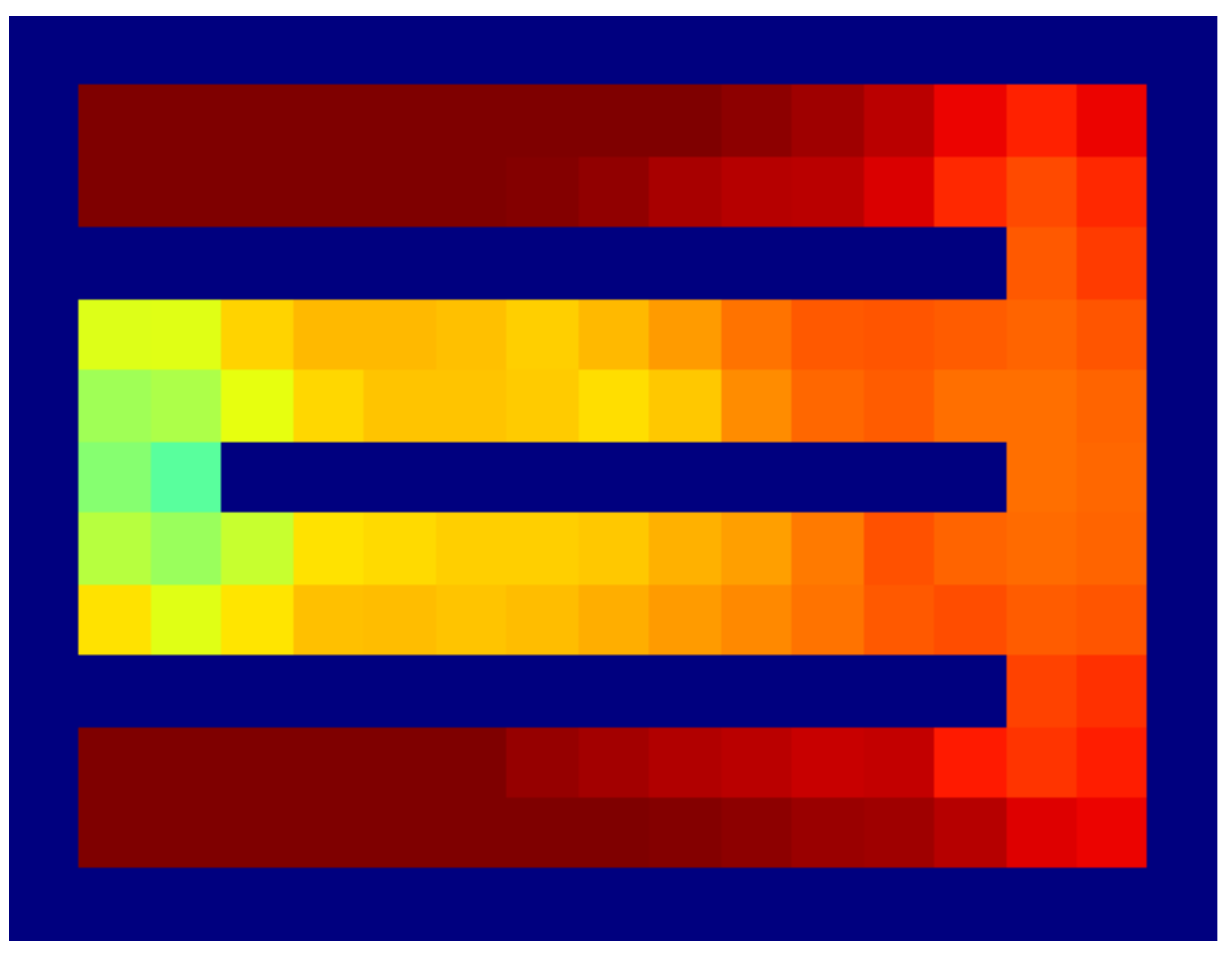}\\

\includegraphics[width=0.18\linewidth]{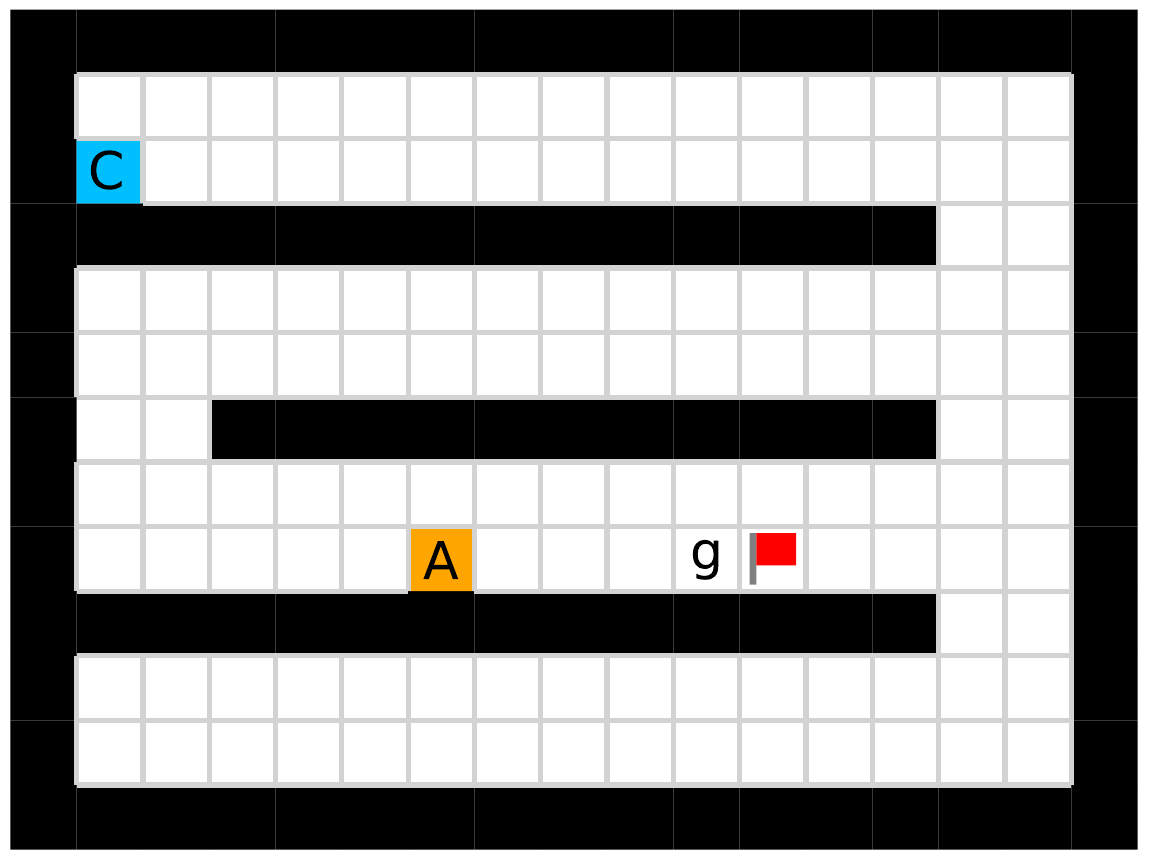}
\hspace{0.4em}
\vspace{0.4em}
\includegraphics[width=0.18\linewidth]{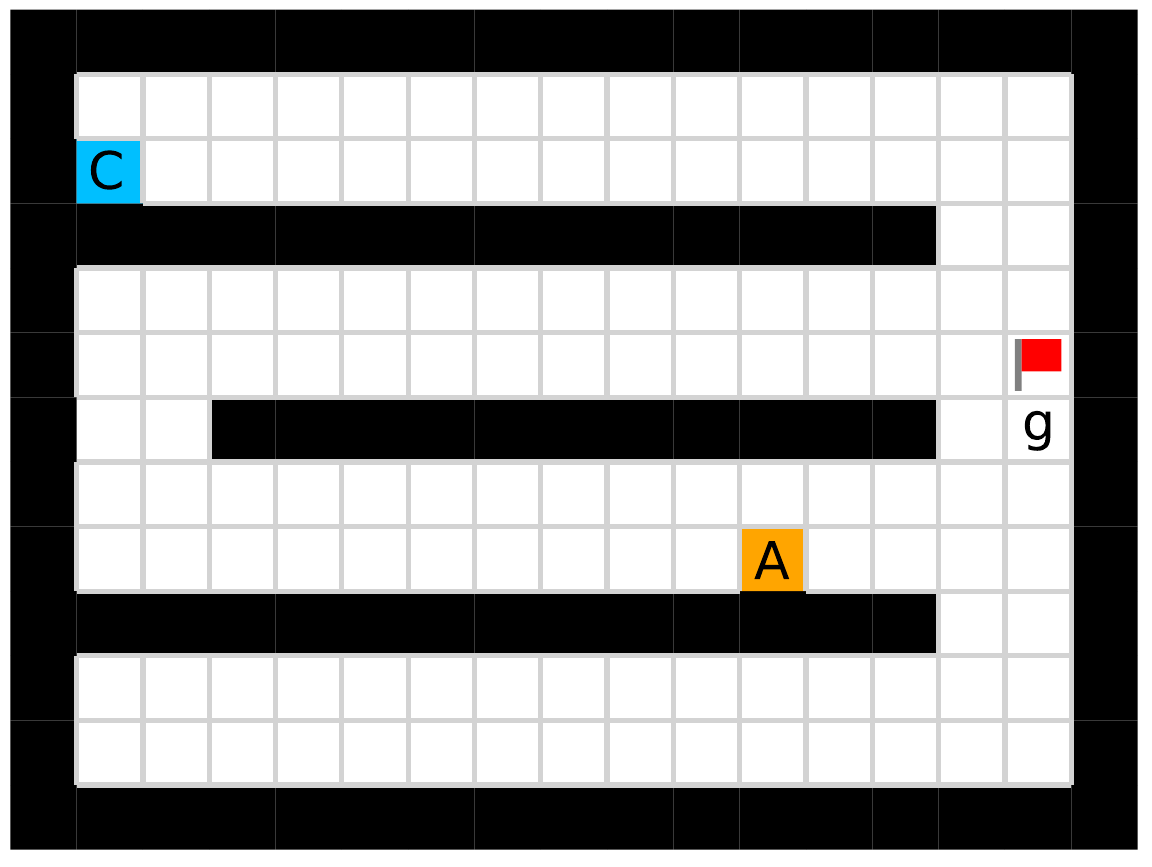}
\hspace{0.4em}
\includegraphics[width=0.18\linewidth]{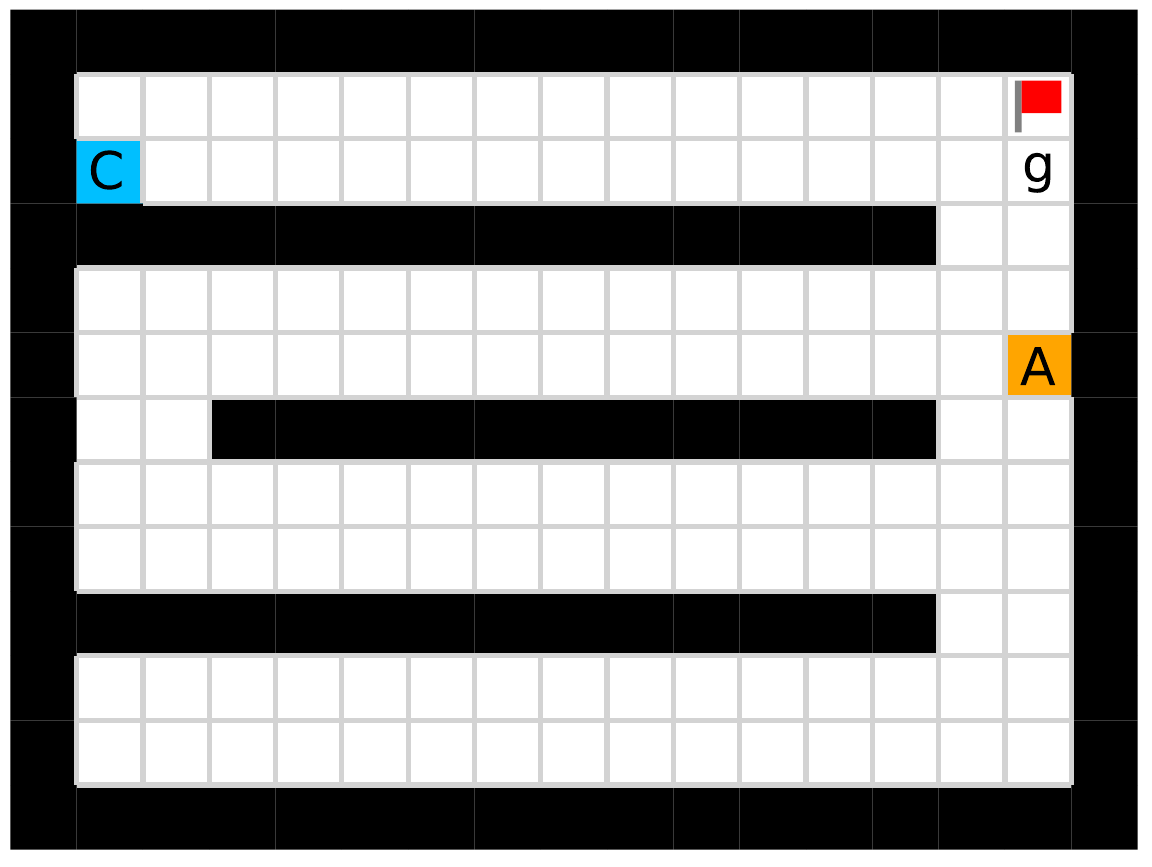}
\hspace{0.4em}
\includegraphics[width=0.18\linewidth]{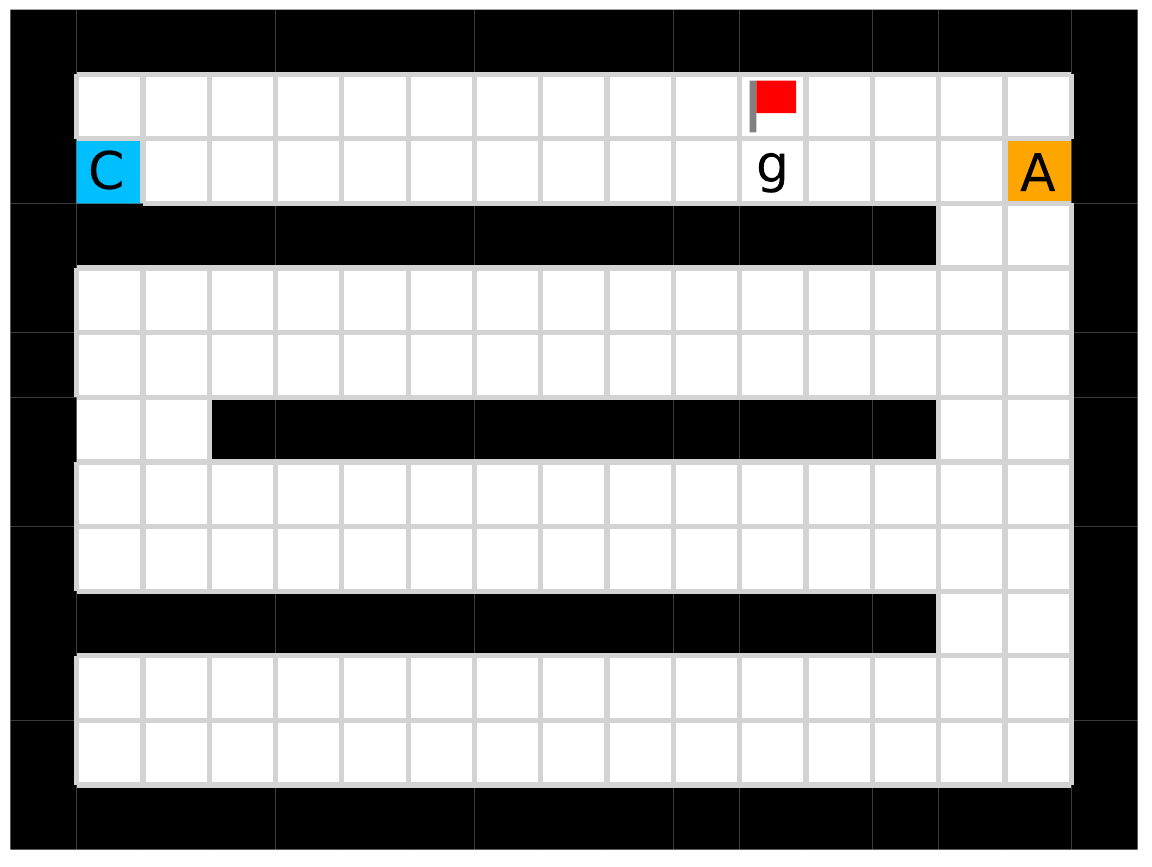}
\hspace{0.4em}
\includegraphics[width=0.18\linewidth]{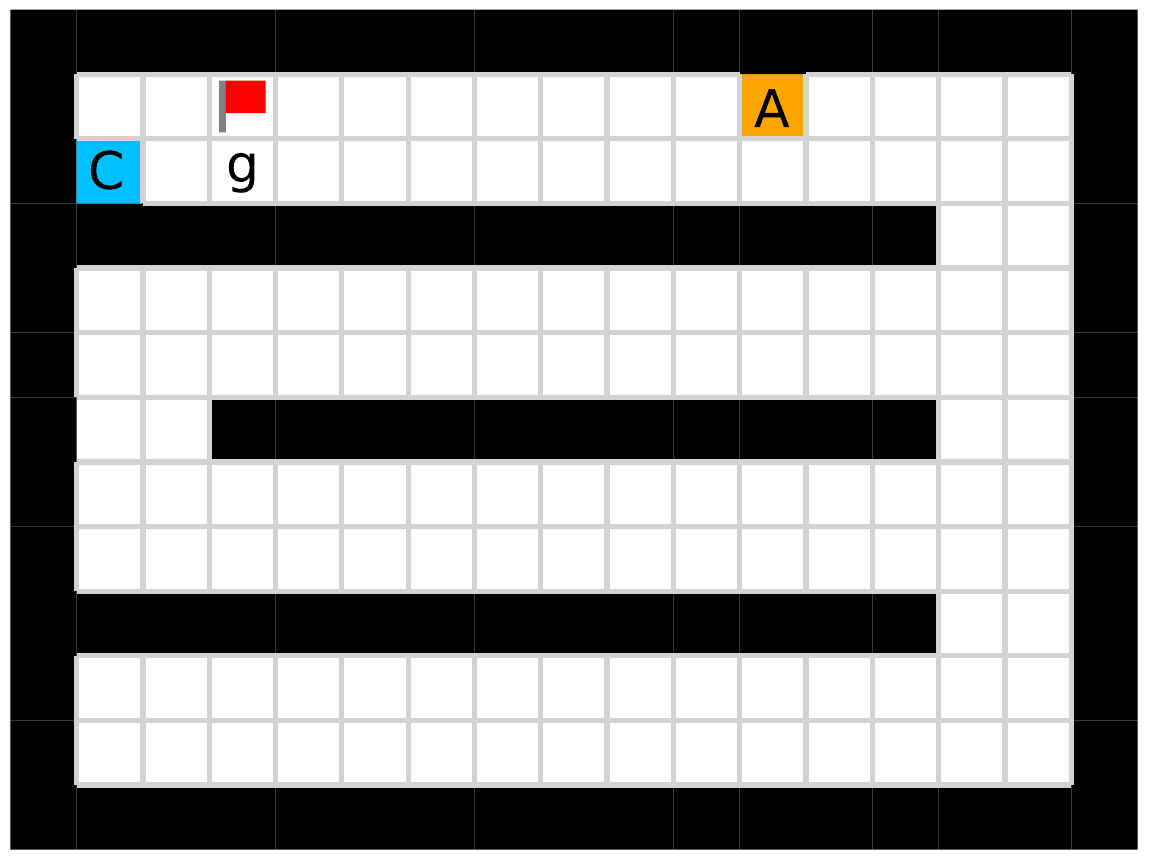}\\

\includegraphics[width=0.18\linewidth]{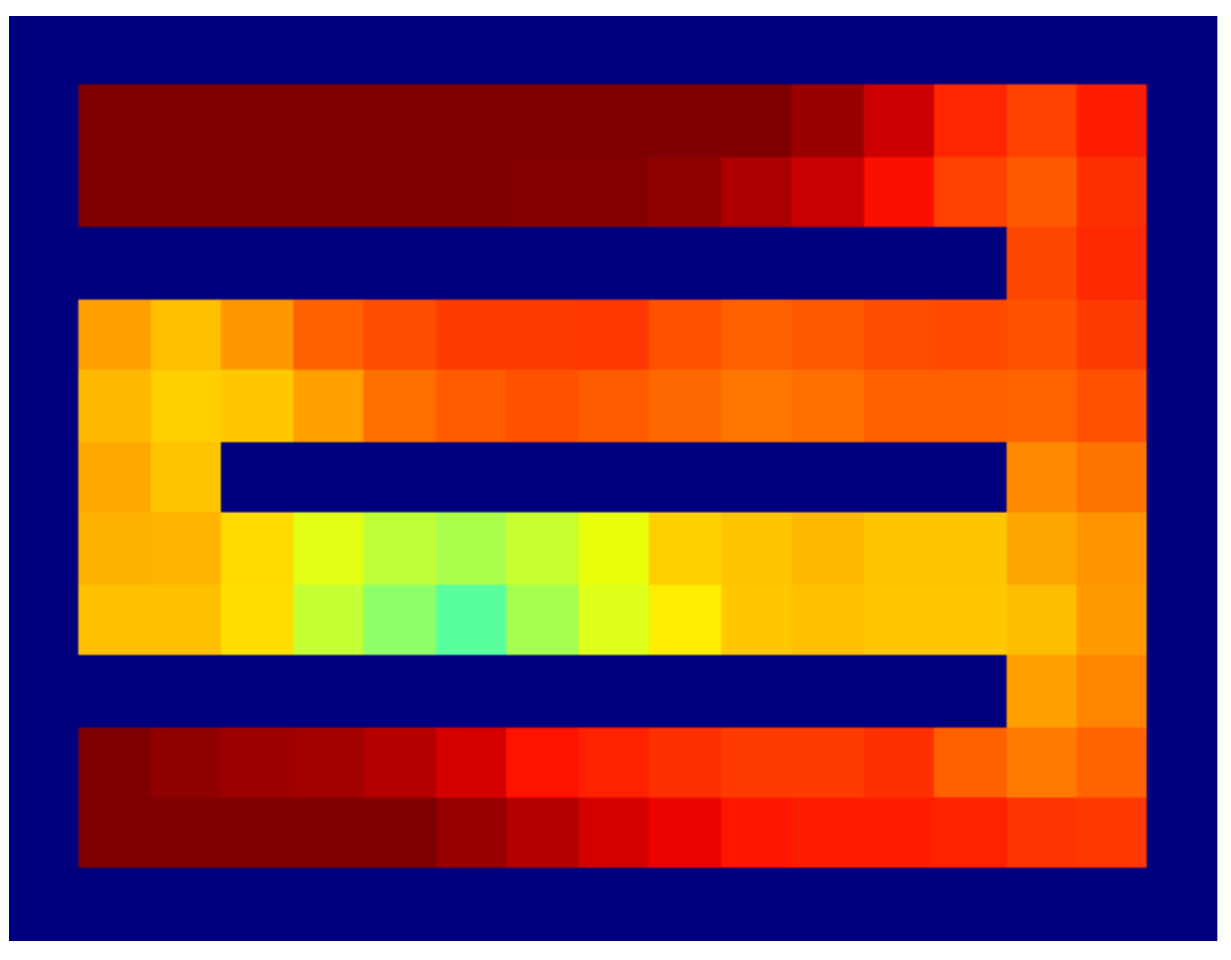}
\hspace{0.4em}
\vspace{1.7em}
\includegraphics[width=0.18\linewidth]{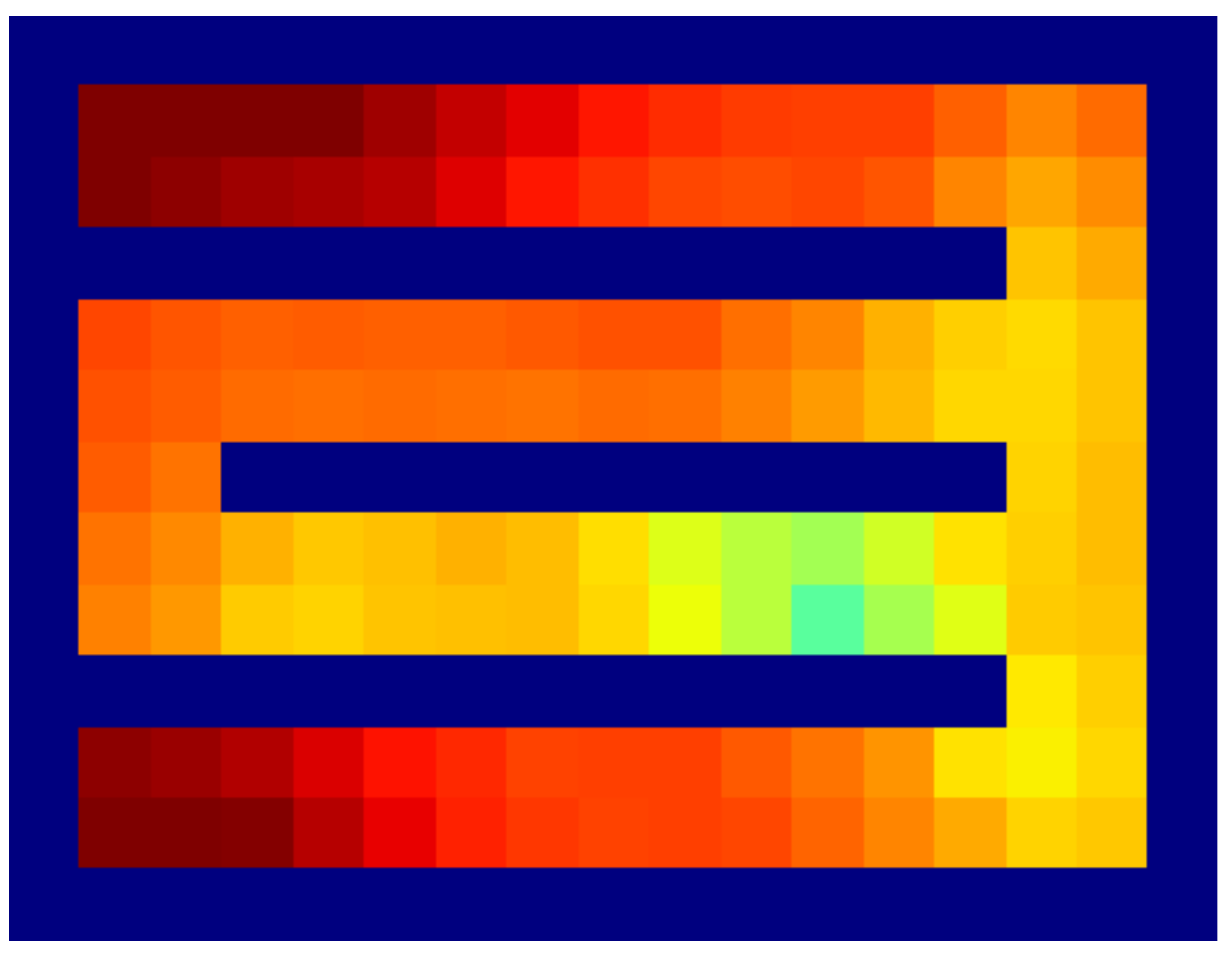}
\hspace{0.4em}
\includegraphics[width=0.18\linewidth]{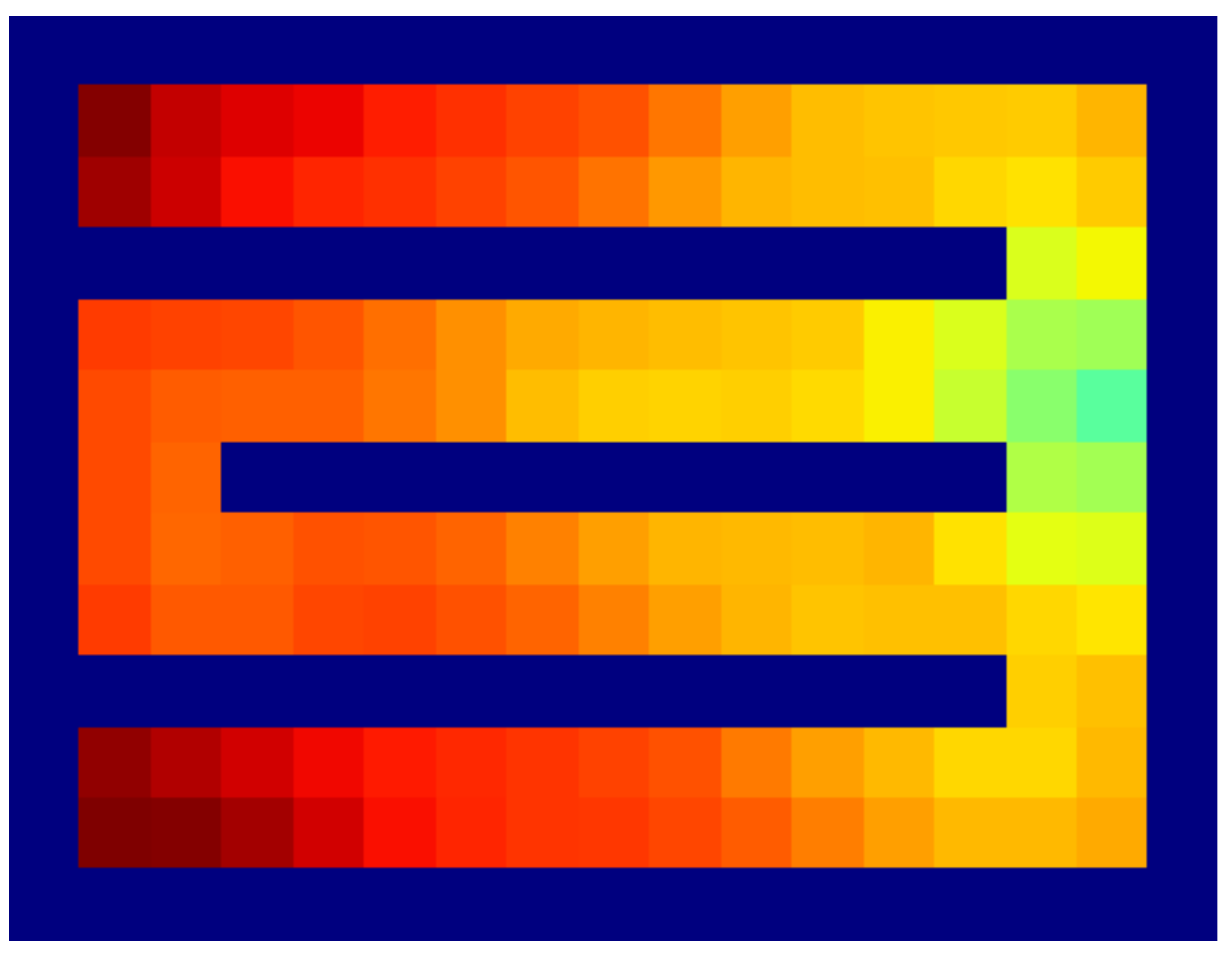}
\hspace{0.4em}
\includegraphics[width=0.18\linewidth]{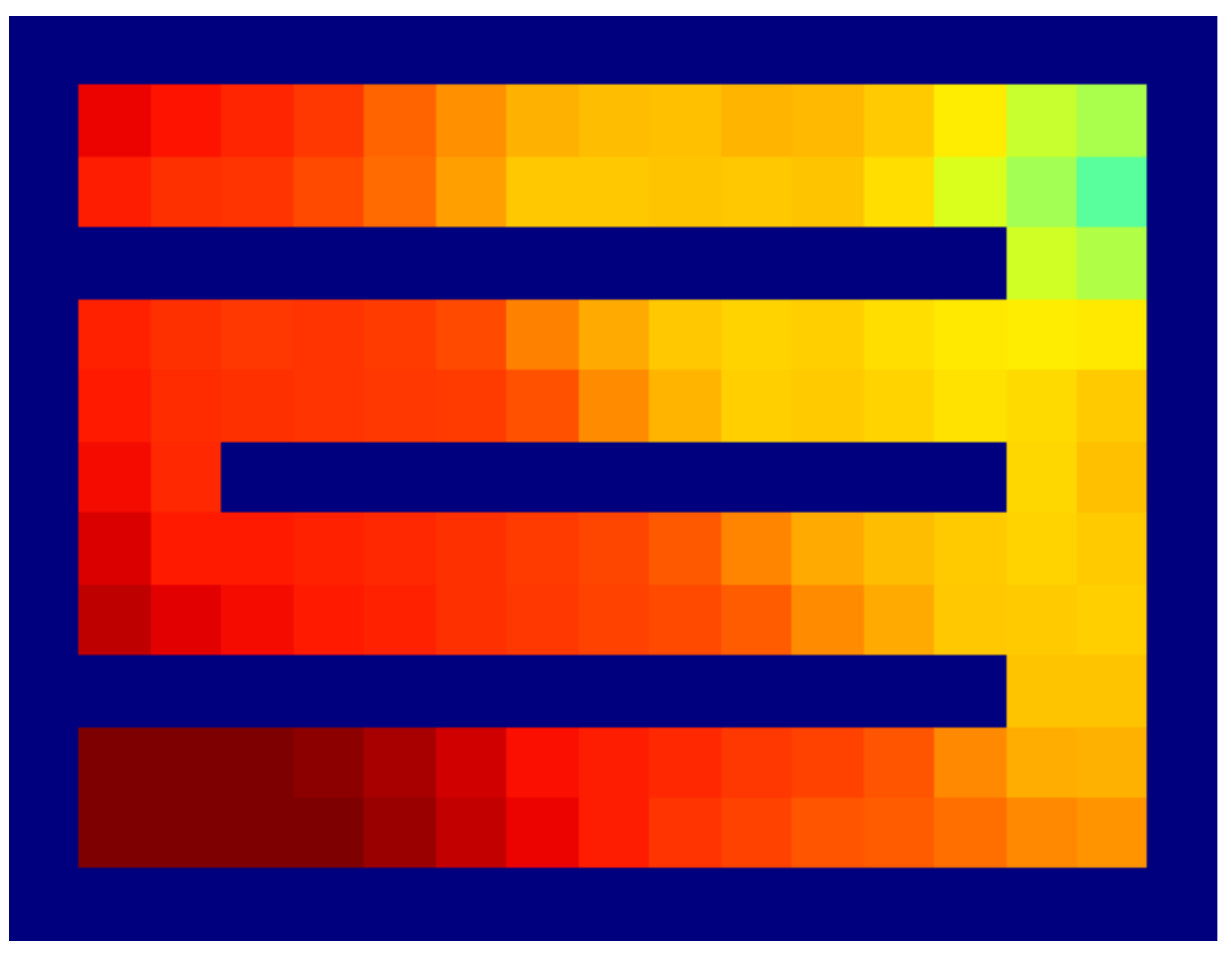}
\hspace{0.4em}
\includegraphics[width=0.18\linewidth]{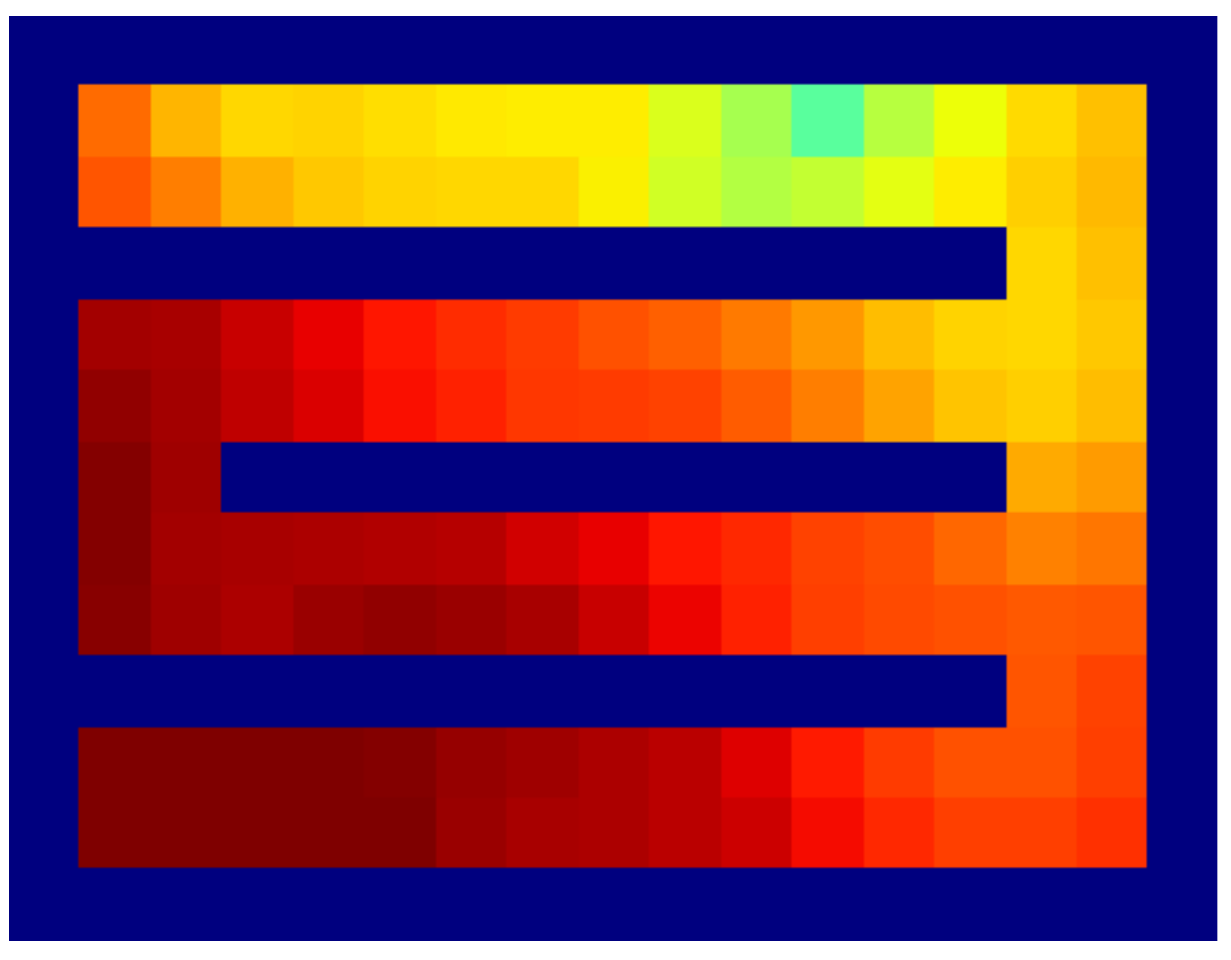}\\

\includegraphics[width=0.18\linewidth]{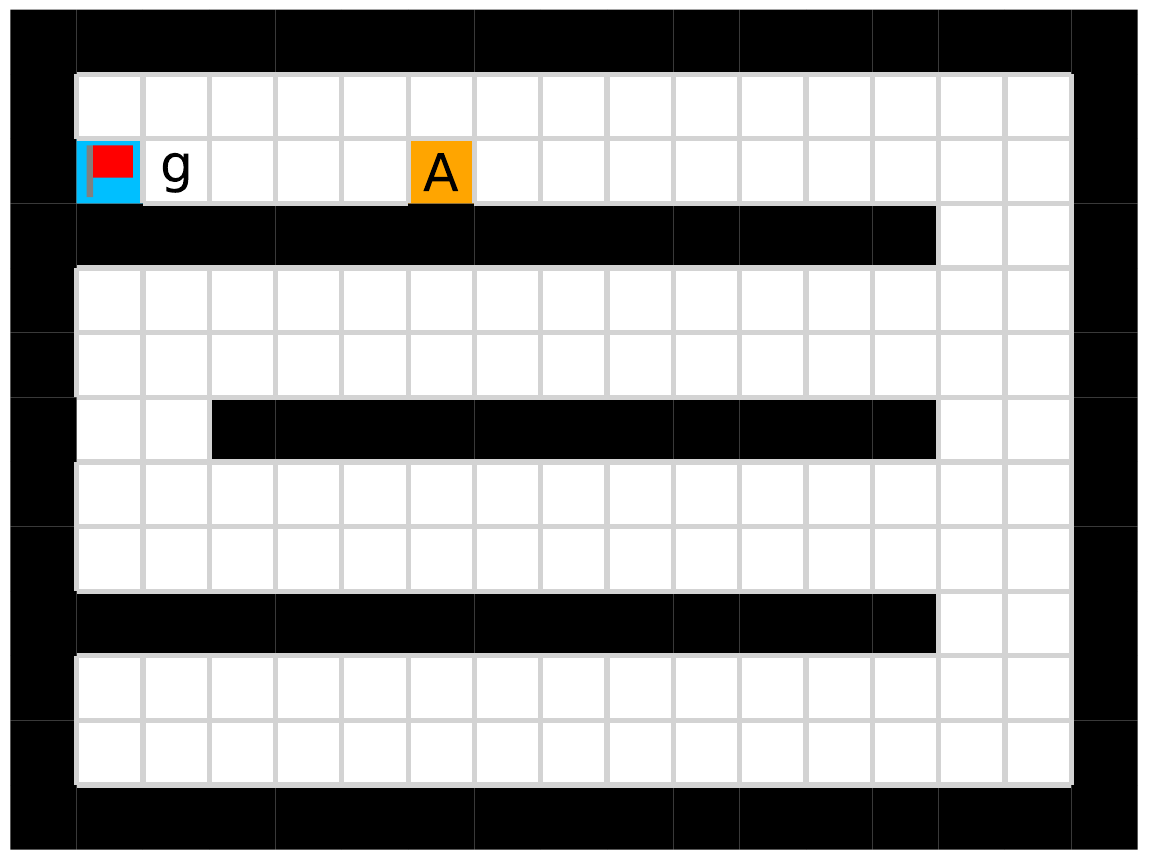}
\hspace{0.4em}
\vspace{0.4em}
\includegraphics[width=0.18\linewidth]{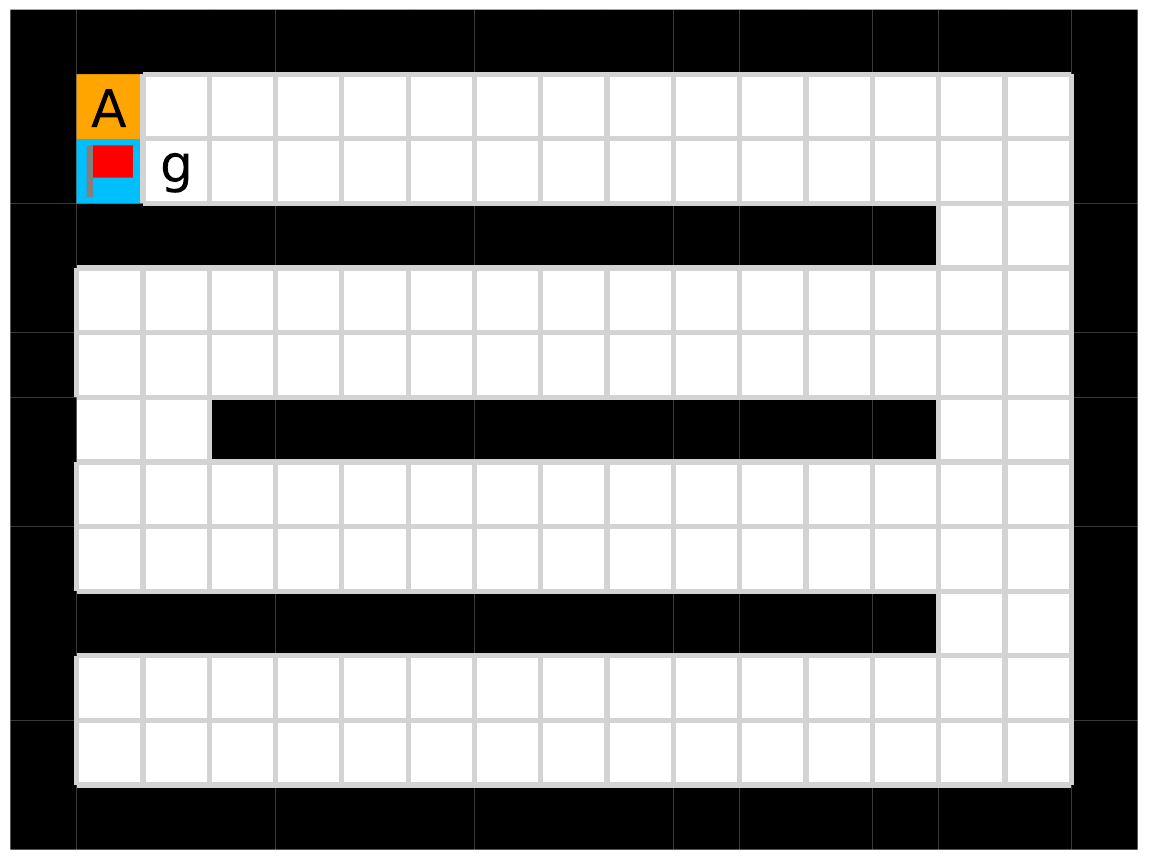} \\

\includegraphics[width=0.18\linewidth]{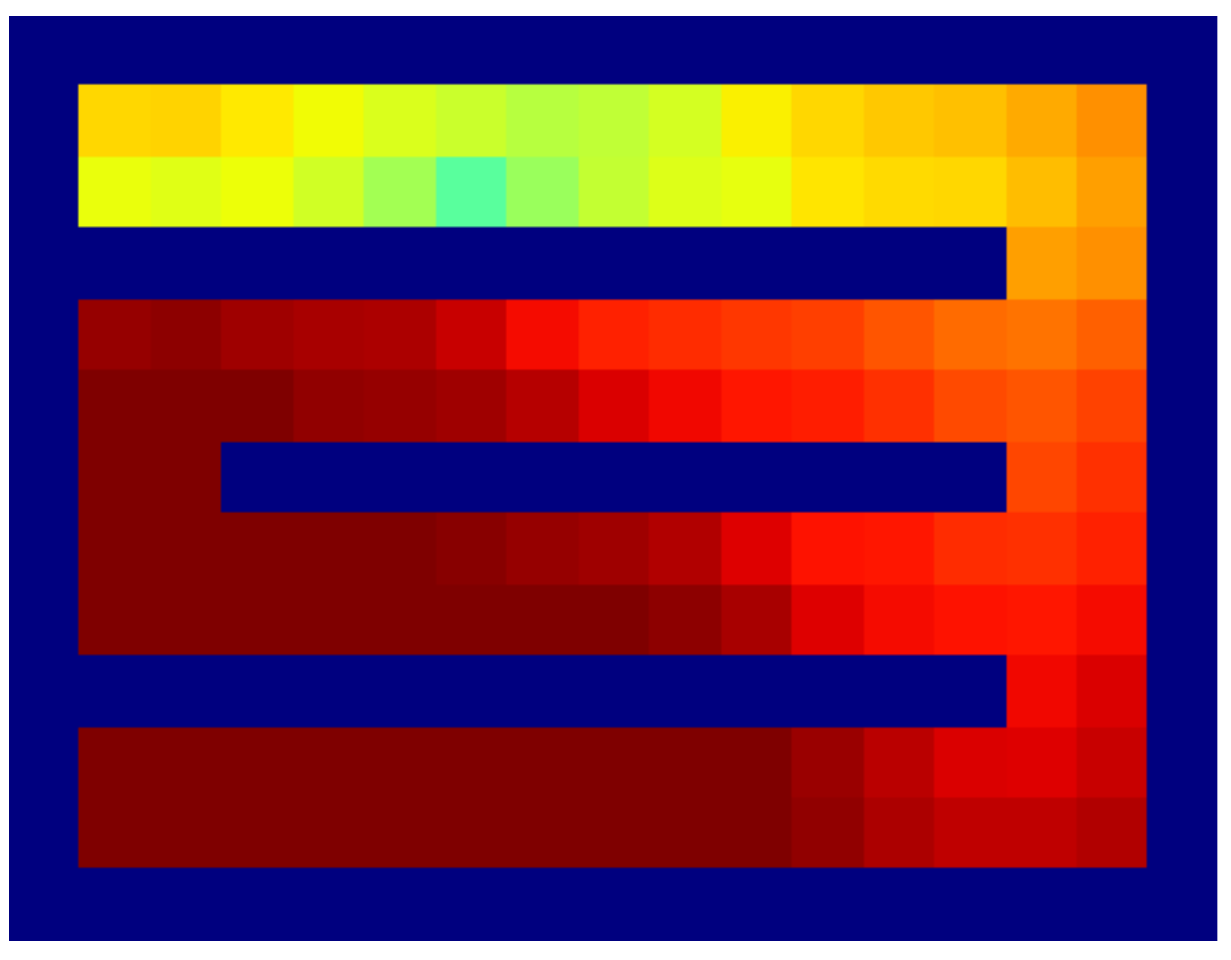}
\hspace{0.4em}
\vspace{0.4em}
\includegraphics[width=0.18\linewidth]{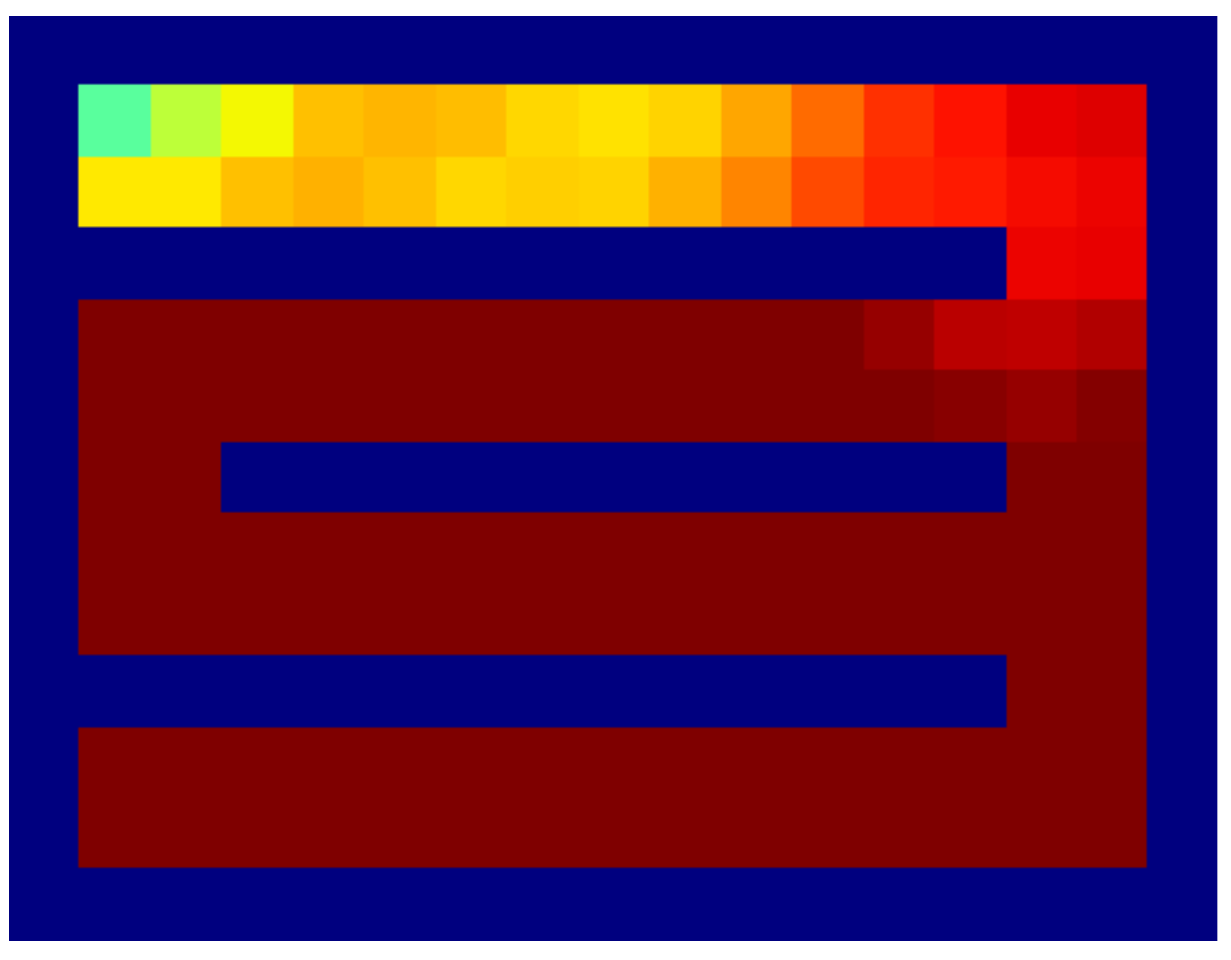}\\

\caption{Additional subgoal and adjacency heatmap visualizations of the Key-Chest task, based on a single evaluation run. The agent (A), key (K), chest (C) and subgoal (g) at different time steps in one episode are plotted. Colder colors in the adjacency heatmaps represent smaller shortest transition distances.}
\label{fig:visualization_supp_keychest}
\end{figure}

We provide additional subgoal and adjacency heatmap visualizations of the Maze and Key-Chest tasks respectively in Figure~\ref{fig:visualization_supp_maze} and Figure~\ref{fig:visualization_supp_keychest}.

\begin{table}
    \centering
    \caption{Hyper-parameters used in discrete control tasks. ``K-C'' in the table refers to ``Key-Chest''.}
    \begin{tabular}{lcc}
        \toprule
        \textbf{Hyper-parameters} & \textbf{Values} & \textbf{Ranges} \\
        \midrule
        \midrule
        High-level TD3 & & \\
        \midrule
        Actor learning rate & \multicolumn{1}{c}{0.0001}& \\
        Critic learning rate & \multicolumn{1}{c}{0.001} & \\
        Replay buffer size & 10000 / 20000 for Maze / K-C & \{10000, 20000\} \\
        Batch size & \multicolumn{1}{c}{64} & \\
        Soft update rate & \multicolumn{1}{c}{0.001} & \\
        Policy update frequency & \multicolumn{1}{c}{2} & \{1, 2\} \\
        $\gamma$ &\multicolumn{1}{c}{0.99} & \\
        High-level action frequency $k$ & \multicolumn{1}{c}{10} & \\
        Reward scaling & \multicolumn{1}{c}{1.0} & \\
        Exploration strategy & Gaussian ($\sigma=3.0 / 5.0$ for Maze / K-C) & \{3.0, 5.0\} \\
        Adjacency loss coefficient $\eta$ & \multicolumn{1}{c}{20} & \{1, 5, 10, 20\} \\
        \midrule
        \midrule
        Low-level A2C & & \\
        \midrule
        Actor learning rate & \multicolumn{1}{c}{0.0001} & \\
        Critic learning rate & \multicolumn{1}{c}{0.0001} & \\
        Entropy weight & \multicolumn{1}{c}{0.01} & \\
        $\gamma$ & \multicolumn{1}{c}{0.99} & \\
        Reward scaling & \multicolumn{1}{c}{1.0} & \\
        \bottomrule
    \end{tabular}
    \label{tab:param_disc}
\end{table}

\begin{table}
    \centering
    \caption{hyper-parameters used in continuous control tasks.}
    \begin{tabular}{lcc}
        \toprule
        \textbf{Hyper-parameters} & \textbf{Values} & \textbf{Ranges} \\
        \midrule
        \midrule
        High-level TD3 & & \\
        \midrule
        Actor learning rate & 0.0001 & \\
        Critic learning rate & 0.001 & \\
        Replay buffer size & 200000 & \\
        Batch size & 128 & \\
        Soft update rate & 0.005 & \\
        Policy update frequency & 1 & \\
        $\gamma$ & 0.99 & \\
        High-level action frequency $k$ & 10 & \\
        Reward scaling & 0.1 / 1.0 for Ant Maze / others & \{0.1, 1.0\} \\
        Exploration strategy & Gaussian ($\sigma=1.0$) & \{1.0, 2.0\} \\
        Adjacency loss coefficient $\eta$ & 20 & \{1, 5, 10, 20\} \\
        \midrule
        \midrule
        Low-level TD3 & & \\
        \midrule
        Actor learning rate & 0.0001 & \\
        Critic learning rate & 0.001 & \\
        Replay buffer size & 200000 & \\
        Batch size & 128 & \\
        Soft update rate & 0.005 & \\
        Policy update frequency & 1 & \\
        $\gamma$ & 0.95 & \\
        Reward scaling & 1.0 & \\
        Exploration strategy & Gaussian ($\sigma=1.0$) \\
        \bottomrule
    \end{tabular}
    \label{tab:param_cont}
\end{table}

\begin{table}
    \centering
    \caption{Hyper-parameters used in adjacency network training.}
    \begin{tabular}{lcc}
        \toprule
        \textbf{Hyper-parameters} & \textbf{Values} & \textbf{Ranges} \\
        \midrule
        \midrule
        Adjacency Network & & \\
        \midrule
        Learning rate & \multicolumn{1}{c}{0.0002} & \\
        Batch size & \multicolumn{1}{c}{64} & \\
        $\epsilon_k$ & \multicolumn{1}{c}{1.0} & \\
        $\delta$ & \multicolumn{1}{c}{0.2} & \\
        Steps for pre-training & 50000 & \\
        Pre-training epochs & 50 & \\
        Online training frequency (steps) & 50000 & \\
        Online training epochs & 25 & \\
        \bottomrule
    \end{tabular}
    \label{tab:param_adj}
\end{table}

\end{document}